\documentclass[11pt]{article}
\ifdefined\XeTeXversion\else
  \DeclareUnicodeCharacter{2192}{\ensuremath{\rightarrow}}
\fi
\usepackage[letterpaper,margin=1in]{geometry}
\usepackage{newtx}
\usepackage{microtype}
\usepackage{booktabs}
\usepackage{array}
\usepackage{graphicx}
\usepackage{listings}
\usepackage{longtable}
\usepackage{rotating}
\usepackage{seqsplit}
\usepackage{calc}
\usepackage[round]{natbib}
\usepackage{caption}
\usepackage[colorlinks=true,linkcolor=black,citecolor=blue,urlcolor=blue]{hyperref}

\title{\bfseries LLM Judges Verify Presence, Not Absence: Omission Blindness in AI Clinical Notes and What Recovers It}
\author{Sebastian Fox\thanks{Lead and corresponding author: \texttt{seb@composo.ai}, ORCID \mbox{\href{https://orcid.org/0000-0001-9839-6952}{0000-0001-9839-6952}}.} \and Luke Markham \and Ryan Lail \and Michael Karotsieris}
\date{}

\providecommand{\tightlist}{\setlength{\itemsep}{0pt}\setlength{\parskip}{0pt}}

\begin{document}
\maketitle

\begin{abstract}
Ambient AI scribes draft clinical notes at scale, and published audits find omission - information the encounter established that the note fails to record - their dominant error class. The standard check is an LLM judge, a second model that reads the transcript and the note and flags problems, and we measure whether these judges can detect omissions. Public corpora cannot supply the answer key (clinician reference notes are materially discrepant with their own transcripts), so we release a benchmark of 500 single-error note pairs built from transcript-derived, audited fact sheets: 298 in which a named fact's primary statement is certainly removed, graded by severity and by how much of it survives elsewhere in the note, against 202 added-or-altered controls. Across eight judge designs, paired discrimination - the flawed note scored below its clean twin, 0.5 a coin flip - reads 0.79-0.94 on added or altered content against 0.50-0.63 on omissions, and on single notes no design flags notes with omissions reliably more often than perfect ones. Wording changes, voting, and prompt optimisation with GEPA move the operating point without creating usable detection. What recovers detection is restructuring the task: list the facts the transcript establishes, then verify each one against the note as a closed presence check. Two methods arrive at that restructuring independently - a per-fact pipeline, and a GEPA-evolved prompt carrying out the same procedure inside one call - and they trade off. The pipeline's flags name the missing fact and its severity at 2.7\% false alarms. The single call detects more notes (36.9\% against 24.6\%, p=0.002 with consultations as the unit, on the evaluation set) at 6.2\% false alarms and roughly a tenth of the measured cost per note at one note per consultation (a third once the per-consultation stages are amortised). Human validation runs through the study. A physician author validated 70 items in a sitting that covered six stages, three of them structurally blinded, and on the ten notes where the pipeline and the best monolithic judge reach opposite conclusions sided with the pipeline on all ten (p=0.002). The severity rubric has since been graded blind by an independent clinician, not an author and with no involvement in the study, and across two sittings on the companion census's findings the two clinicians agree with it to within a grade. On real vendor notes from the companion census no benchmark threshold transfers, but the single call re-calibrated still detects more than the best of the eight at half its false-alarm rate, and on the census notes carrying a verified omission three quarters of the notes the pipeline flags name the very fact the census's panel verified. Omissions whose fact survives as a restatement elsewhere in the note defeat both routes. We release the benchmark, prompts, and all judgements.

\end{abstract}

\section{Introduction}

\begin{figure}[!htbp]
\centering
\includegraphics[width=0.85\linewidth]{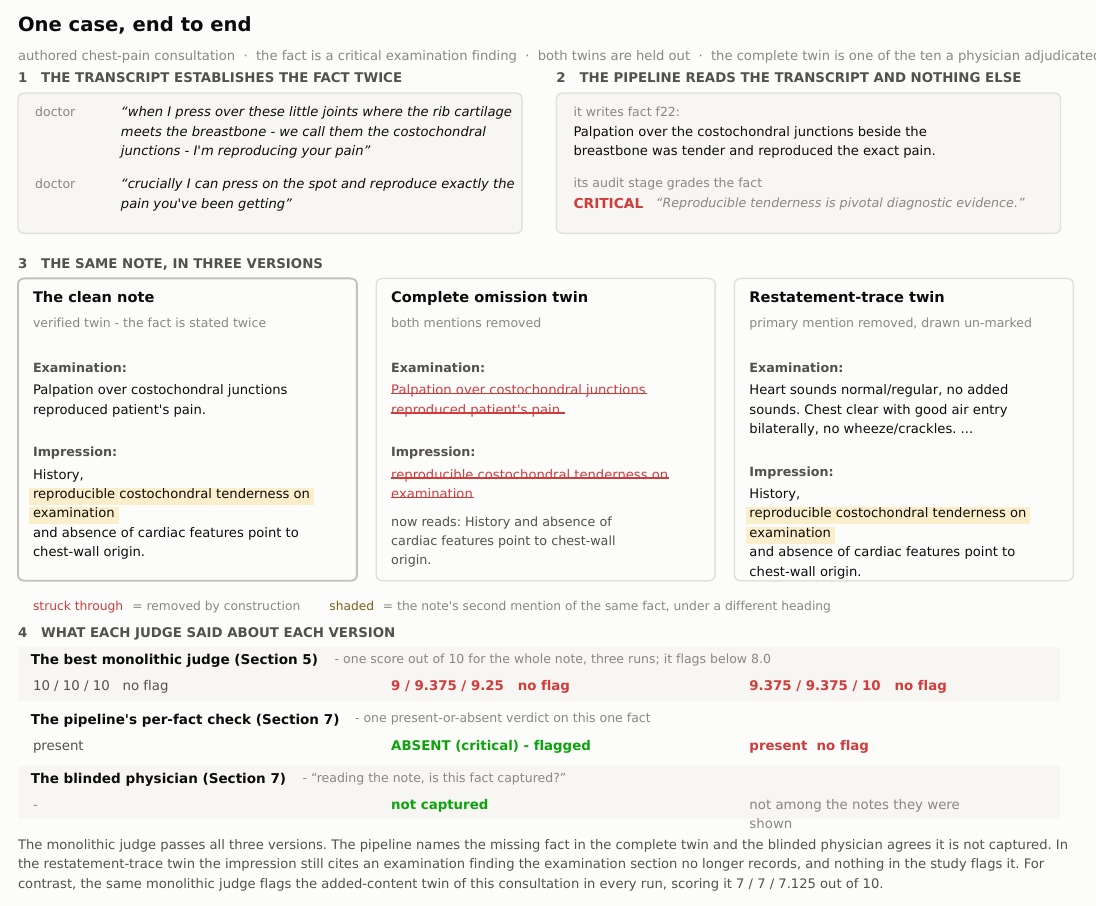}
\caption{\textbf{One case end to end:} a critical examination finding the transcript
establishes twice, the note records twice, and two twins remove differently. The
monolithic judge passes all three versions; the pipeline's per-fact check names the missing
fact in the complete twin and the blinded physician agrees it is not captured. The
restatement-trace twin is drawn un-marked, as a reader sees it: the examination reads
normally, the impression still cites the finding it no longer records, and nothing in
the study flags it. Each note column is abridged to the passages the fact lives in, and
the twins differ from the clean note by deletion only. The case is an authored
consultation, so the figure excerpts only our own transcript and notes and no vendor text
appears anywhere.}
\label{fig:worked}
\end{figure}

A telephone consultation contains no physical examination. Two of the three commercial ambient scribes we audited in a companion census nonetheless produced notes for telephone consultations that document one: history rewritten as performed examination, in the part of the note a clinician signs and a later reader treats as observed fact. That census (``One note in three: a verified census of three deployed AI scribes, and the instrument that counted it'', released concurrently) put three deployed products through the same 142 consultations. Of its 565 notes (282, 141 and 142, since one product writes two note styles), 31.3\% {[}27.0, 35.6{]} carry at least one verified failure, and the largest single group is allergy and medication information that is missing or wrongly recorded. The census also reports that the rate falls to 24.8\% {[}20.8, 29.0{]} once the two failure classes a patient record would have prefilled, invented identity details and dates, are set aside (its Appendix A.7). Published human audits find omission the dominant error class outright, at 54\% to 86.3\% of all errors across three audits \citep{biro2025accuracy,anderson2025evaluating,kernberg2024using}. This paper takes up the question the census leaves open: whether the automated quality layer can see these failures at all. Figure~\ref{fig:worked} takes one missing finding through this paper's own benchmark, from the transcript to each judge's verdict.

That quality layer is the LLM judge: a second model, prompted to read the transcript and the note and flag problems. Judges are what makes evaluation affordable at the scale scribes operate at, and what vendors, buyers and regulators are increasingly asked to trust. The paper's argument runs in four steps. We observe that across an eight-design ablation and, with the single exception Section 5 names, every reference judge we ran, judges verify what a note contains and sit near chance on what it omits, the absence-blindness recently documented for language models in general \citep{fu2025absencebench} arriving intact at the evaluation layer healthcare actually runs. We hypothesise that the open question ``is anything missing?'' gives the model nothing to check, because an omission leaves no text to point at. That predicts two things: remedies that change only how the judge is asked (wording, criteria, answer format, votes, automated prompt optimisation) should move its operating point without creating usable detection, and a fix that gives the judge something concrete to point at should recover it.

The test bears both halves out. Every conventional remedy we tried moved the operating point without creating usable detection. What recovers detection - partially, since no method tested comes close to solving the class - is converting the absence question into presence checks: first list the facts the transcript establishes, then verify each one against the note. Two methods here make that conversion and arrived at it separately, a pipeline we designed around it and a prompt the optimiser GEPA evolved that instructs the same procedure inside one call. Two implementations converging independently on the same restructuring is evidence that the recovery comes from the restructuring rather than from any particular prompt or pipeline design.

The error class published audits find dominant in scribe output - a class our own stricter instrument likely undercounts as well - is precisely the class LLM judges sit near chance on.

\textbf{Contributions.} (i) A released benchmark where absence is certain and graded, by severity and by how much of the removed fact survives elsewhere in the note. (ii) The presence/absence asymmetry, isolated by ablation rather than inferred from one judge design. (iii) A null across five remedy families, each moving the judge's operating point without creating usable single-note detection, with the size of each shift measured; and a localisation of the mechanism, in which a list of concrete facts to check accounts for about a third of the recovery and closed per-fact verdicts for the rest. (iv) Two working detectors implementing that conversion, compared head to head: a fact-enumeration pipeline whose every flag names the missing fact and its severity, and a single GEPA-evolved call at roughly a tenth of the measured cost per note at one note per consultation (a third amortised). With them we give a per-fact decision rule, a blinded physician adjudication, and the case that defeats both methods, named as an open problem.

\section{Related work}

Four literatures meet here. \textbf{Absence detection in language models.} AbsenceBench \citep{fu2025absencebench} established the capability gap outside medicine: given the original and edited documents side by side, models catch insertions at 86.2 to 99.5\% F1 depending on domain and drop by an average of 56.9\% on omissions. It names an architectural cause - transformer attention cannot easily attend to gaps, because an absence supplies no keys - and a repair that works in that setting, inserting placeholder markers where content was removed. Ours is the setting their limitations name as untested and expected to be harder: semantic rather than surface absence, no side-by-side original at deployment, the medical domain, judges rather than diff engines. Their repair is unavailable here, because an evaluator does not know where the hole is. Concurrent work finds the same asymmetry in the reviewing role on non-clinical answer keys, where a reviewer can interrogate only candidates it can name \citep{chen2026judging}; the full comparison is Appendix I.

\textbf{LLM-as-judge reliability in clinical text.} MEDEC \citep{ben-abacha-etal-2025-medec} measures judges on physician-labelled commission errors, the mirror class to omission. Concurrent work \citep{delucia2026same} independently finds clinical completeness judges at chance-to-slightly-above AUC, with rubric and prompting gains that reflect threshold shifts rather than separation - what this paper's Sections 5 and 6 find, reached independently on a different task and without a commission condition. Three further angles on judge reliability bear on this, including the one result pointing the other way; Appendix I takes each in turn \citep{dahlberg2026measuring,hussain2026blending,bergman2026judges,vachhani2026beyond}.

\textbf{Scribe audits and clinical note evaluation.} The audits establish the error classes and omission's dominance \citep{biro2025accuracy,anderson2025evaluating,kernberg2024using,taylor2026quality}. The companion census reads those audits' divergent counts against the instruments that produced them, and a concurrent five-country paired study measures the instrument's own share of a published rate \citep{bergman2026quality}. In a separate planted-error study \citep{biro2025opportunities}, reviewing physicians caught omissions at roughly the same rate as objective errors, so the asymmetry measured here appears characteristic of machine judges rather than of review as such (Section 5.5).

\textbf{Summarisation factuality} is the literature our fix belongs to: atomic-fact decomposition \citep{min-etal-2023-factscore}, QA- and entailment-based metrics \citep{scialom-etal-2021-questeval,laban-etal-2022-summac}, per-fact coverage in the lineage of RAGAS, a widely used open-source evaluation library \citep{es-etal-2024-ragas}. MED-OMIT \citep{schumacher2023medomit} is the fix's direct ancestor - fact enumeration for omission measurement, importance-weighted, on the same public corpus as one of our strata - and CARE \citep{bedi2026care} engineers calibrated flags for hallucination and omission together, arriving by design at the conclusion Section 7 reaches by measurement: presence and absence need separate detectors.

\textbf{The gap.} None of these lines of work simultaneously tests semantic omissions in free clinical notes, with no answer key available to the evaluator, with severity and surviving trace graded, and with benchmark construction reported separately from judge performance. That conjunction is what this paper supplies.

Testing whether a judge can find an omission needs notes in which a named fact is certainly absent, and no public corpus we examined supplies one.

\section{The benchmark: absence made certain and graded}

We set out to build notes with a known hole in them, graded by how much the hole matters and by how much of the fact survives elsewhere. Figure~\ref{fig:construction} lays out the whole build, every stage with what it produced and every check with what it rejected, so this section states only what the instrument guarantees. The protocol, the ledgers and the rubric are Appendix A. One scope decision sits above the rest: the benchmark measures whether facts the transcript establishes survive into the note, a stricter and more measurable standard than clinical documentation adequacy, since a concise note can legitimately leave transcript facts out (the size of our fact sheets against the fact sheets human authors wrote by hand, below, is the measure of that gap).

\subsection{The public corpora's reference notes are not an answer key}

The obvious way to build the benchmark is to borrow the answer key: public corpora pair consultations with clinician-written reference notes, and scoring a generated note against the reference is the standard recipe \citep{papadopoulos-korfiatis-etal-2022-primock57,yim2023acibench}. Treating a reference note as truth assumes it is true, so we audited every usable reference note in our corpus's two public strata against its own transcript, instructed not to penalise valid paraphrase, with one repair cycle (53 of PriMock57's 57 consultations and 45 of ACI-Bench's 48 survive the corpus checks; Appendix A records the drops). All 53 PriMock notes (100\%, 95\% interval {[}93.2, 100{]}) and all 45 ACI-Bench notes (100\%, {[}92.1, 100{]}) carry at least one material discrepancy with their own transcript, at means of 10.7 and 7.1 material discrepancies per note. Counting every severity, missing facts dominate (56\%, or 574 of PriMock's 1,030 discrepancies of all grades), then outright errors, then hardened uncertainty, where ``i think you've got a little contusion'' becomes a definite diagnosis. The caveats on that 100\% are in Appendix A. What matters is the density and its severity split, with 345 of the 1,030 graded critical. Judging a scribe against these references imports 7 to 11 material errors per note into the answer key itself.

\subsection{Building the key from the transcript, and grading what matters}

\begin{figure}[!htbp]
\centering
\includegraphics[width=\linewidth]{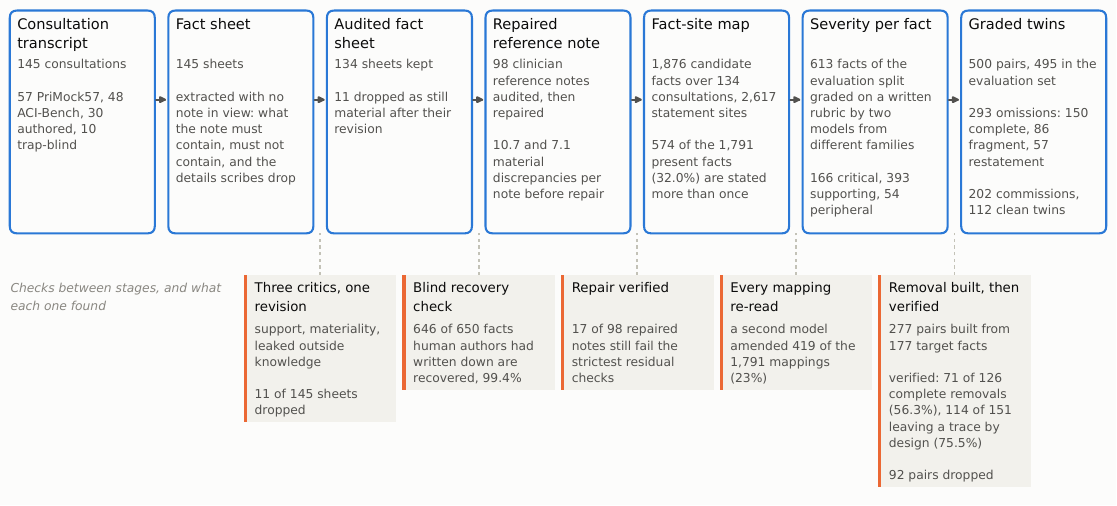}
\caption{\textbf{The answer key is built from the transcript, not borrowed from the
reference note.} Each box is a construction stage and carries what it produced; each chip
below is the check that stands between two stages, and what that check found or
rejected. The trap-blind stratum in the corpus band is authored consultations written with no
knowledge of the benchmark's trap schema, as a representativeness guard. Severity
is graded per fact by two models from different families on a written rubric; they split
on 93 of the 613 facts, and the lower grade is taken. The two checks that carry findings
of their own are blind recovery, where extraction recovers 646 of 650 facts that human
authors had written down for the same consultations, and removal verification, where only
71 of 126 attempts to remove every trace of a fact from a real note survive a
cross-family panel against 114 of 151 for removals that leave a trace by design.}
\label{fig:construction}
\end{figure}

If reference notes cannot be truth, the transcript has to be. For each consultation we build a fact sheet by blind extraction from the transcript, audit it with three model critics, and use it to repair and verify the reference note, which becomes the clean note every pair is built from (Figure~\ref{fig:construction}). The repaired note is complete against its fact sheet, not against everything the transcript contains: 17 of the 98 repaired notes still fail the strictest residual check and are kept and logged (Appendix A). Two facts about the instrument matter here. Its accuracy was measured blind against consultations we had written ourselves, recovering 646 of 650 authored facts (99.4\%, 95\% interval {[}98.8, 99.9{]}, computed by resampling whole consultations) on the 27 of 30 blind re-extractions that passed the critic panel. And it writes longer sheets than the human authors, so a consolidation pass classes each item core or contextual and builds pairs from core items only - per-stratum means of 24.8 to 32.6 against the authors' 24.1, a completeness bar above what a clinician would write to.

Grading omissions by importance needs importance to be stable, and natively it is not: two frontier models grading the same details agree at a Cohen's kappa of 0.177 {[}0.04, 0.31{]}, barely above coincidence. A written rubric fixes most of it. Under it the two cross-family graders whose grades the benchmark ships reach kappa 0.662 {[}0.59, 0.73{]} on the corpus's 683 graded traps, an interval that does not overlap the unanchored run's (that run is a same-family pair of successive generations on 171 matched traps, so the contrast moves the grader pair and the item set as well as the rubric; the same family's rubric-anchored grades against its own unanchored run read kappa 0.39, Appendix A), and where they still disagree the benchmark takes the lower grade. (The 683 traps are the population that stability run graded, spanning authored and extracted sheets; the 613 facts in Figure~\ref{fig:construction} are the evaluation split of the per-fact grading the pairs carry.) Both rubric runs grade more severely than either model does natively, which we disclose wherever a result is conditioned on severity. The rubric is Appendix A, a physician author's blinded re-grade of it Appendix E. The same written rubric has since been applied blind by an \textbf{independent clinician} to a fresh sample of findings in the companion census, with 9 of 12 grades exact and 3 one grade apart (Appendix A summarises both census sittings; the census's own appendix carries the protocol and the item-level grades). Independent, here and wherever the word appears in this paper, means not an author of this study and with no involvement in it at any stage. Across the two census sittings that graded against it, this clinician's twelve and a physician author's regrade of twenty verified findings, 25 of 32 grades are exact and no disagreement anywhere exceeds a single grade, so the rubric reproduces to within a grade under two clinicians who graded disjoint material. What that does not establish is a direction for the rubric itself: the two raters lean opposite ways, so no claim here rests on the rubric's grades leaning one way. (The one-grade lean Section 7.5 reports is a different measurement, of the pipeline's own audit-stage grades against a single rater.)

With the sheets graded, removing a fact should be mechanical, and it is not, because real clinical notes repeat themselves. Mapping every place each candidate fact is stated in the clean note shows 574 of the 1,791 present (32.0\%) stated in more than one place. The redundancy reaches the facts that matter clinically: among the moments each sheet marks as ones a scribe could plausibly get wrong, the entries the severity grades attach to, 51.6\% of critical and 61.2\% of supporting entries are stated more than once. Verification is where the finding is. A cross-family panel verified 71 of 126 complete-removal attempts (56.3\%), so 44\% failed - mostly because the fact was not truly absent after all, or the note no longer read naturally - against 114 of 151 (75.5\%) for partial removals that leave a trace by design, with every edit logged and the 92 rejected pairs excluded, none of them patched (Appendix A). The consequence for the field is direct: a benchmark built only from complete omissions tests an unrepresentative slice of the phenomenon, because in real notes the complete case is the minority that survives construction.

\subsection{What the released set holds}

\begin{table}[!htbp]
\centering
\small
\setlength{\tabcolsep}{4pt}
\begin{tabular}{@{}lrrrr@{}}
\toprule
Surviving trace & Critical & Supporting & Peripheral & All\\
\midrule
Complete & 85 (20) & 36 (22) & 29 (29) & 150 (71)\\
Fragment trace & 44 (29) & 34 (31) & 8 (8) & 86 (68)\\
Restatement trace & 22 (11) & 33 (33) & \textit{2 (2)}$^{\ddagger}$ & 57 (46)\\
\midrule
All & 151 (60) & 103 (86) & 39 (39) & 293 (185)\\
\bottomrule
\end{tabular}
\caption{\textbf{The omission half by surviving trace and severity.} How much of the removed fact survives elsewhere in the note, against how much the loss matters. Severity is graded per fact on a written rubric by two models from different families, taking the lower grade where they disagree. Bracketed figures are the newly built cohort alone, which is the internally uniform one. Counts are the 293 omission pairs of the evaluation set. $^{\ddagger}$ A restatement trace needs a fact stated twice in earnest, and peripheral facts are both the rarest and the least redundant in the corpus, so the restatement-trace peripheral cell holds only 2 pairs; its count is printed in italic and no rate is conditioned on it.}
\label{tab:trace}
\end{table}

The set is 500 pairs, 495 in the evaluation set, over 117 consultations (the five pairs outside the evaluation set sit on consultations the exploratory optimiser's development pool used and are excluded for every judge, leaving 112 evaluation consultations), with 45 matched couples, 90 pairs, in which the same fact is removed at two different trace levels. Throughout, an \textbf{omission} pair removes a fact from the note and a \textbf{commission} pair adds or alters one. The evaluation set's omission half is 293 pairs - 150 complete removals, 86 where a fragment of the fact survives, 57 where a full restatement of it survives - against 202 commissions (95 additions, 107 alterations) and 112 clean twins, each twin being the same note with the fact intact. We name the partial levels by what survives: a \textbf{fragment trace} leaves a fragment of the removed statement or its immediate neighbour, and a \textbf{restatement trace} leaves a second, independently stated mention of the same fact, which almost always sits under a different heading (Section 7). The released data labels the three levels \texttt{complete}, \texttt{partial-weak} (fragment) and \texttt{partial-strong} (restatement), and Table~\ref{tab:trace} is the trace-by-severity layout.

Provenance is a field on every pair, and Appendix A tabulates the four sources the set pools. One consequence belongs here: all 202 commission controls are carried over from an earlier build while most omission pairs are newly built, so the headline asymmetry compares across construction cohorts. The gap dwarfs any cohort difference we can measure: the commission side is anchored externally on physician-curated errors (Section 5), and a robustness slice restricted to the uniformly built omission pairs \emph{widens} the asymmetry, from 0.305 to 0.329 on the best design (Appendix A).

The holdout has two layers. The exploratory optimiser's development pool is 22 consultations carved out of the corpus entirely, and the confirmatory campaign of Section 6 then partitioned the evaluation set by consultation, learning and accepting on 65 of them; the 151-pair subset over the remaining 47, the basis of every held-out claim here, was committed before that search began and never seen by any optimiser.

\section{The judges under test, and how detection is scored}

\begin{figure}[!htbp]
\centering
\includegraphics[width=\linewidth]{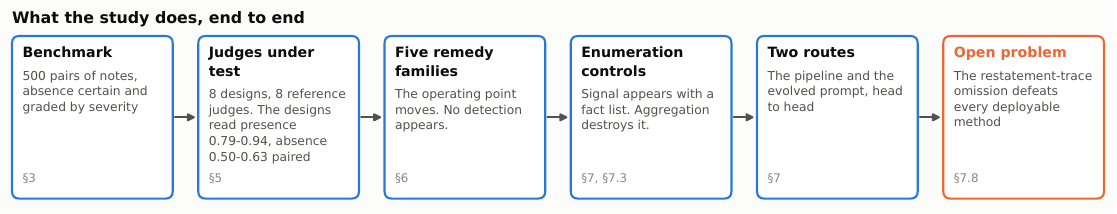}
\caption{\textbf{The experimental programme in one line.} The benchmark of Section~3
feeds an eight-design ablation and eight reference judges (Section~5, the asymmetry);
five remedy families fail to restore single-note detection (Section~6); two enumeration
controls locate the mechanism (Section~7, Section~7.3); two methods implement the
conversion of absence into presence checks and are compared head to head (Section~7);
and one case defeats every deployable method (Section~7.8). Counts, where a box carries
them, are the pairs or designs involved.}
\label{fig:map}
\end{figure}

\subsection{Two ways to keep score, and only one of them exists in production}

\begin{table}[!htbp]
\centering
\footnotesize
\setlength{\tabcolsep}{3pt}
\begin{tabular}{@{}>{\raggedright\arraybackslash}p{1.02in}>{\raggedright\arraybackslash}p{1.55in}>{\raggedright\arraybackslash}p{1.42in}>{\raggedright\arraybackslash}p{2.30in}@{}}
\toprule
Set & What it holds & Its role & The claims measured on it\\
\midrule
Exploratory pass & 38 pairs (32 of them omissions) against 32 clean notes, judged while the pipeline was still being designed & development readings only; ten of its consultations recur in the confirmation subset, with extraction caches reused & no headline claim: the pipeline's first paired read (0.828) and the per-fact rule's back-check (25.0\% at 6.2\%), quoted only as replication checks\\
\addlinespace
Held-out confirmation subset & 151 pairs over 47 consultations: 131 omissions, 20 commissions, 47 clean twins & committed before any optimisation began; never seen by any optimiser & the pipeline tiers' paired scores (three replicates), the per-fact rule's 20.6\% at 2.1\%, the reasoning-budget ladder, the optimiser's held-out tests, the second-family transfer. Marked $^{\dagger}$ in Table~\ref{tab:practitioner} and named per line in the figures\\
\addlinespace
Evaluation set & 495 pairs over 112 consultations: 293 omissions, 202 commissions, 112 clean twins (five pairs on the exploratory optimiser's consultations excluded) & the full released benchmark; the eight designs' and the reference judges' flag rules fixed before the run, the two methods' rules established post hoc & the asymmetry across the eight designs (0.79--0.94 against 0.50--0.63), the reference judges, the remedy families, both methods' deployment rules (24.6\% at 2.7\%; 36.9\% at 6.2\%) and their head-to-head, and the second-family replication of two of the eight designs (Section 5.3)\\
\addlinespace
Real vendor notes & 261 of the companion census's 565 notes: 87 carrying a panel-verified omission, 174 with no verified finding & nothing tuned on it; each judge runs once at its benchmark rule & whether the benchmark's operating points transfer to production notes (Section 7.7, Section 8, Appendix K)\\
\bottomrule
\end{tabular}
\caption{\textbf{Which numbers live on which set.} Every judge number measured on this paper's own benchmark
comes from exactly one of these four sets, and rates measured on different sets are
never compared or ranked against each other, with one deliberate exception, flagged
in the sentence that makes it: Appendix K's transfer measurement, where the movement of an
operating point between two sets is itself the result.
The first three are built from the benchmark
of Section~3; the fourth is real scribe output with the companion census's verified
findings as ground truth.}
\label{tab:sets}
\end{table}

Figure~\ref{fig:map} maps the experimental programme this section begins. Everything that follows is scored two ways, and much of the paper rests on the distinction. \textbf{Paired discrimination} asks whether the judge scores the flawed note strictly below its own verified-clean twin. It is the share of pairs the judge orders correctly, plus half of any ties (a paired win rate, in the same family of measure as AUC), where 0.500 is a coin flip. It is a capability ceiling and nothing more, because production never has the twin. \textbf{Single-note detection} is the deployment measure: one note arrives, the judge flags it or does not, and that flag rate means nothing except beside the same judge's false-alarm rate on clean notes. Each design carries a \textbf{flag rule}, the fixed threshold or verdict that turns its output into a flag, and ``at its own rule'' below means measured under that rule. A judge's \textbf{operating point} is those two rates together, and every remedy below is judged by whether it slides the point along a curve or lifts the curve. We call detection \textbf{usable} when it clears the same judge's false-alarm rate by more than chance at a false-alarm rate a deployment could carry, which throughout this paper means 10\% or below. Both rates are reported for every design that can express them, and never combined into one number. Every judge number measured on this paper's own benchmark comes from one of four sets of notes (an exploratory pass, a held-out confirmation subset, the full evaluation set, and the 261 real vendor notes taken from the companion census), and Table~\ref{tab:sets} maps which claims live on each. No rate is compared across two of them except where the sentence says so, at Appendix K's transfer measurement. Two things are measured elsewhere and say so where they appear: Section 5.4's external commission anchor, on physician-labelled texts, and the census figures quoted from the companion paper, which rest on that paper's own larger denominators rather than on the 261 notes here.

\subsection{Eighteen judge configurations, and the ablation underneath them}

\begin{table}[!htbp]
\centering
\small
\setlength{\tabcolsep}{5pt}
\begin{tabular}{@{}p{0.21\linewidth}p{0.37\linewidth}cp{0.16\linewidth}p{0.13\linewidth}@{}}
\toprule
Group & Design & n & Answer returned & Scored on\\
\midrule
The ablation & every combination of three choices: check faithfulness alone or faithfulness and completeness $\times$ answer yes/no or score 0--10 $\times$ ask once or eight times & 8 & a verdict or a score & evaluation set, 3 replicates\\
\addlinespace
Reference judges & the deployed faithfulness judge, in three wordings differing only in how they treat omissions & 3 & a verdict & evaluation set\\
 & G-Eval: scores several named dimensions, each as one open question & 1 & scores & evaluation set\\
 & checklist judge: generates a per-consultation checklist ($\sim$25 items) and answers each & 1 & item answers, read as a score & evaluation set\\
 & the engineered omission judge: the strongest single-prompt judge we could write by hand, with eight few-shots & 1 & a score + free text & evaluation set\\
 & the RAGAS-style recipe: per-fact coverage in the RAGAS lineage, our own deliberately naive implementation, which extracts the transcript's facts once ($\sim$62) and checks each against the note & 1 & per-fact verdicts, read as a score & evaluation set\\
 & the optimiser's exploratory winner at the standard settings (its re-run with a reasoning budget is Section 7's evolved prompt) & 1 & a score & evaluation set\\
\addlinespace
Enumerate-then-check pipeline & two-stage: extract facts from the transcript alone, then one closed present-or-absent check per fact & 1 & per-fact verdicts & held-out subset\\
 & three-stage: adds a critic audit, severity grades, a full/partial/absent verdict and the decision rule & 1 & graded per-fact verdicts & held-out subset\\
\midrule
 & & 18 & &\\
\bottomrule
\end{tabular}
\caption{\textbf{The eighteen judge configurations.} All single-call judges run with no reasoning effort inside a 1,024-token answer cap; the multi-call recipes carry per-call budgets (Section 4). ``Evaluation set'' is the 495 pairs and 112 clean twins; ``held-out subset'' the 151 pairs and 47 twins over 47 consultations that no optimiser ever saw. The eight reference judges are the second group, and the figures plot the same eight. Section 7 additionally re-runs a subset of these prompts with a reasoning budget; those re-runs are the same prompts, not new designs.}
\label{tab:judges}
\end{table}

Three groups of judges share one pinned GPT-family judge model for every judging call, at matched settings (the pipeline's extraction, audit and second-look stages run on their own pins). Table~\ref{tab:judges} lays out the full roster, scores follow in Table~\ref{tab:grid} and Table~\ref{tab:practitioner}, and every model in the study's judging and construction path is pinned in Appendix D. Those settings include how much the judge may deliberate: every ablation design and every reference judge runs with no reasoning effort inside a 1,024-token answer cap, chosen to reflect the cost and latency budgets an evaluation actually runs to in the field, inside a deployed application, while the multi-call recipes have per-call budgets instead. Section 7 measures what raising that budget does. Throughout, we call a judge \textbf{monolithic} when it reads the whole note with one prompt and returns one holistic judgement, as against the enumerating designs, which list the transcript's facts first and check each one. Two things the roster cannot show belong here: the reference judges include the faithfulness judge running in our own production evaluation - \textbf{the deployed faithfulness judge} below - in three wordings differing only in how they treat omissions (Appendix H), and \textbf{the RAGAS-style recipe} below, a per-fact coverage recipe in the RAGAS lineage, is our own deliberately naive implementation, its prompts disjoint from the pipeline's. One more reference judge earns its handle here: the strongest single-prompt judge we could write by hand, with eight worked examples, is \textbf{the engineered judge} below.

The ablation crosses the three design choices underneath any deployed judge. The first is \textbf{what the judge is told to check}: that everything in the note is supported by the transcript, or that plus whether anything important is missing, paired so that criterion scope itself is the variable. The second is \textbf{how it answers}, a yes/no verdict or a 0-to-10 score. The third is \textbf{how many times it is asked}, once or as an eight-call ensemble combined by a winsorised mean, the extreme scores replaced by the nearest kept values before averaging. That gives \textbf{eight designs} differing one decision at a time, deliberately more minimal than anything someone would ship. Each ran three times: 14,568 judgements over 65,556 model calls, zero recorded parse failures, and the same verdict on the same note across runs 93.2\% of the time.

\subsection{Analysis conventions}

Repeat runs collapse to a per-note majority before any significance test. Bootstrap resampling is at the consultation, not the note, because pairs from one consultation share a transcript, a clean twin and a fact list; for the same reason, comparisons between two methods' flags are collapsed to the consultation before testing. Paired flag comparisons use an exact McNemar test with a Holm correction across each family of comparisons, and the one unpaired comparison of two small rates uses Boschloo's exact unconditional test; proportions carry Wilson 95\% intervals and agreement coefficients cluster-bootstrap ones, so a bare bracketed interval anywhere below is one of those two at the 95\% level. The results sections quote the p-value and leave the test's name here. Where we claim something made no difference, we state how large a difference we could have detected rather than reporting a p-value above 0.05.

\section{The finding: judges verify presence, not absence}

\subsection{Presence is detected strongly, absence at close to chance}

\begin{table*}[!htbp]
\centering
\footnotesize
\setlength{\tabcolsep}{2.1pt}
\begin{tabular}{@{}>{\raggedright\arraybackslash}p{1.62in}ccccccc@{}}
\toprule
& \multicolumn{2}{c}{paired discrimination} & \multicolumn{3}{c}{at its own flag rule} & & \\
\cmidrule(lr){2-3}\cmidrule(lr){4-6}
Design & commissions & omissions & detection & false alarms & $z$ & AUC, omissions & \begin{tabular}[c]{@{}c@{}}best detection at\\$\leq$10\% false alarms\end{tabular}\\
\midrule
Faithfulness only, yes/no, one sample (F-bin-k1) & 0.875 & 0.500 & 9.6\% & 8.9\% & 0.34 & 0.503 & -- \\
Faithfulness only, yes/no, eight samples (F-bin-k8) & 0.904 & 0.509 & 9.6\% & 7.4\% & 1.15 & 0.512 & -- \\
Faithfulness and completeness, yes/no, one sample (FC-bin-k1) & 0.792 & 0.539 & 40.3\% & 31.8\% & 2.71 & 0.542 & -- \\
Faithfulness and completeness, yes/no, eight samples (FC-bin-k8) & 0.869 & 0.591 & 42.3\% & 32.1\% & 3.25 & 0.562 & -- \\
Faithfulness only, score, one sample (F-score-k1) & 0.911 & 0.549 & 1.0\% & 1.5\% & -0.68 & 0.549 & 1.0\% @ 1.5\% \\
Faithfulness only, score, eight samples (F-score-k8) & 0.944 & 0.582 & 2.4\% & 2.1\% & 0.32 & 0.549 & 2.4\% @ 2.1\% \\
Faithfulness and completeness, score, one sample (FC-score-k1) & 0.895 & 0.585 & 4.7\% & 5.7\% & -0.71 & 0.573 & 4.7\% @ 5.7\% \\
Faithfulness and completeness, score, eight samples (FC-score-k8) & 0.939 & 0.634 & 8.3\% & 6.5\% & 1.02 & 0.575 & 8.3\% @ 6.5\% \\
\bottomrule
\end{tabular}
\caption{\textbf{The eight judge designs, on the 495-pair evaluation set.}
Each design crosses what the judge is told to check (faithfulness only, F; faithfulness and
completeness, FC), how it answers (a yes/no verdict, bin; a 0-to-10 score, score) and how many
times it is asked (once, k1; eight times as an ensemble combined by a winsorised mean, the extreme scores replaced by the nearest kept values before averaging, k8). Paired
discrimination is the tie-adjusted probability that an errored note scores below its own
verified-clean twin, where 0.500 is a coin flip; it needs the twin, so it is a ceiling rather
than a deployment measure. Detection and false alarms are single-note flag rates at each
design's own rule, fixed before the run (a FAIL verdict for the single-sample yes/no designs, a
winsorised mean of the eight encoded verdicts at or below 5.0 for the eight-sample yes/no
designs, a score of 7 or below for the single-sample scored designs, and a
winsorised mean below 8.0 for the eight-sample scored ones), over 879 omission records and 336 clean records: three replicates of 293 omission pairs
and 112 clean twins, with 202 commission pairs carrying the commission column. $z$ is the
two-proportion statistic for a design's detection against its own false alarms, pooled
over the three replicates; pooling counts three passes over the same notes as independent
observations and inflates $z$, so the per-note-majority reading in Section 5 (1.2 to 1.7
standard errors for the two yes/no completeness designs) is the honest strength of that gap. Where the flag
line was drawn is not what limits these designs. The best-separating cut the data itself
suggests - the threshold that maximises Youden's J (detection minus false alarms) on the mean-over-runs aggregate - lands at 8.17 (F-score-k1), 8.40 (F-score-k8), 8.50 (FC-score-k1) and 8.19 (FC-score-k8), every
one of them above the 7 and 8.0 fixed before the run rather than below; and threshold-free, the
area under the curve stays between 0.503 and 0.575 across all eight. A dash means the
quantity is not swept: a single yes/no verdict leaves no score to sweep, and the eight-verdict
aggregate is a vote fraction rather than a score. The AUC printed for a single-sample yes/no design is the
area its single operating point defines; for the eight-sample yes/no designs it is computed over
the eight-verdict vote fraction.}
\label{tab:grid}
\end{table*}

\begin{figure}[!htbp]
\centering
\includegraphics[width=\linewidth]{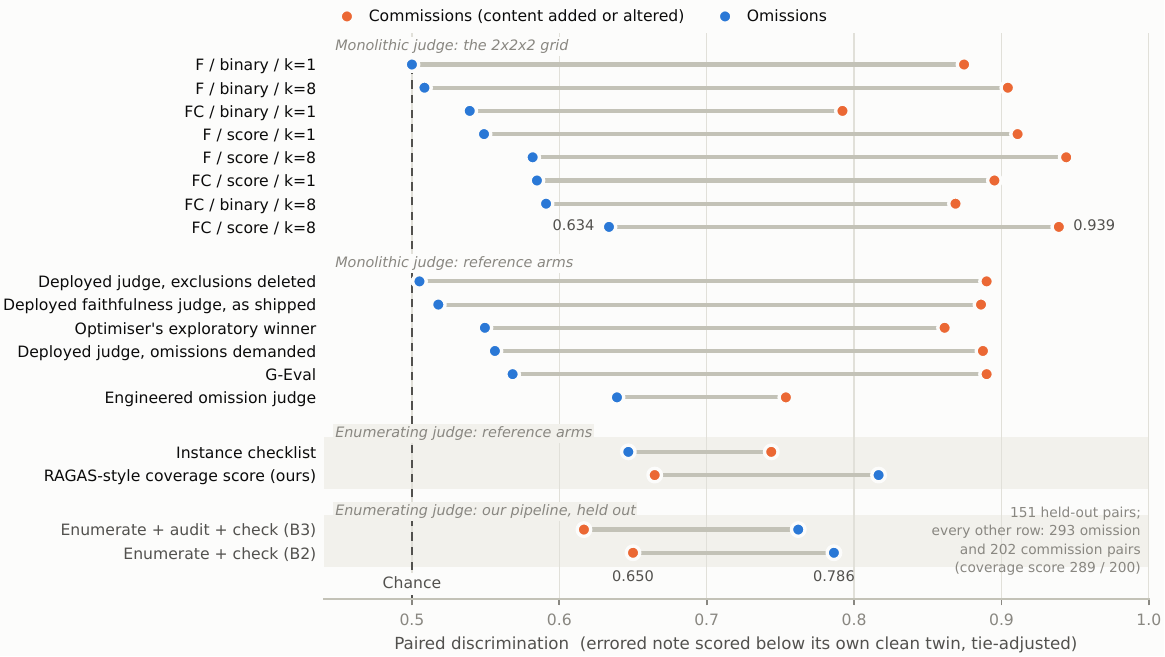}
\caption{\textbf{Paired discrimination on omissions against commissions, per judge
design.} The measure is the tie-adjusted probability that the errored note scores
strictly below its own clean twin, plus half the tie mass; chance is 0.500. The sets:
the monolithic and reference-judge rows are the 495 evaluation pairs over 112
consultations (293 omissions, 202 commissions; the eight designs at 3 replicates,
reference judges at 2), and the two pipeline rows are the 151 held-out confirmation pairs
over 47 consultations (131 omissions, 20 commissions), means over 3 replicates. The
coverage-score row is the one exception to those counts: its paired values rest on 289
omission and 200 commission pairs per replicate, because parse failures cost that arm 14
of its 1,214 records. Design codes read as criterion / answer format / sample count for
the designs Section~4 describes in words; the released data spells them
\texttt{F-bin-k1} to \texttt{FC-score-k8}.}
\label{fig:asymmetry}
\end{figure}

Across all eight ablation designs, judges separate commissions from their clean twins and do not separate omissions. On added or altered content the eight run at 0.792 to 0.944 paired, six of them at 0.87 or above (Table~\ref{tab:grid}, Figure~\ref{fig:asymmetry}). On omissions the same eight read 0.500 to 0.634. The faithfulness-only judge answering yes or no is at exactly a coin flip with the clean twin in hand, and even with all three design choices at their best values (faithfulness plus completeness, scored, eight samples - the best monolithic judge throughout this paper) the score reaches only 0.634 (0.601 to 0.655 across the three replicates). Inside that best design, what signal remains concentrates where it helps least. By how much of the fact survives: complete removals 0.690, fragment traces 0.607, and restatement traces 0.526, a coin flip - the restatement case is a deliberately built minority, taken up as a named open case in Section 7. By severity: critical 0.683, supporting 0.586, peripheral 0.568. The severity gradient is real, modest, and present only in the best design; the trace gradient is the larger effect, because a fact stated twice and half-removed defeats the judge outright whatever its severity. The matched-fact contrast in flag rate between complete and partial removals of the same fact is p=0.022 uncorrected and does not survive the family-wise correction Section 4 sets (24 matched contrasts, three per design: complete against each partial trace level and against the two pooled; adjusted p=0.54), so it is a pattern the data suggest rather than an effect they establish.

\subsection{On a single note, nothing usable survives from the eight designs}

Those are ceilings measured with the twin available for comparison, so the deployment question is separate. No design among the eight separates notes with omissions from perfect ones at a false-alarm rate a deployment could carry. Among the scored designs the largest gap between detection and false alarms is 1.02 standard errors, which is noise (two of the eight even run marginally negative, flagging clean notes slightly more often than errored ones); the two yes/no completeness designs show larger gaps, flagging 40--42\% of omission notes against 32\% of clean ones, but once repeat runs collapse to a per-note majority that gap is 1.2 to 1.7 standard errors, and a judge that flags a third of perfect notes is not a usable detector either way. Nor is that an artefact of where the flag line was drawn. Threshold-free, the area under the detection curve runs 0.503 to 0.575 across the eight, and swept to a 10\% false-alarm ceiling the best scored design reaches 8.3\% detection at 6.5\% false alarms, a gap that does not clear its own noise (Table~\ref{tab:grid}). What is missing is separability, not calibration.

\subsection{The same asymmetry in the reference judges and a second model family}

The reference judges reproduce it, with one exception that Section 7.8 takes up, the RAGAS-style recipe, which inverts it. The deployed faithfulness judge scores 0.518 paired on omissions against 0.886 on commissions, G-Eval 0.568 against 0.890, the checklist judge (a transcript-generated checklist of roughly 25 items, answered item by item) 0.647 against 0.744, the engineered judge 0.639 against 0.754, and the first optimisation campaign's winning prompt 0.549 against 0.861. All 21 candidate prompts in that campaign's development pool detect added and altered content better than omissions, by 14.6 to 49.3 percentage points (Appendix B). The asymmetry also survives a change of judge model family: re-running two of the eight designs with a Gemini-family judge, prompts unchanged to the character, commissions sit at 0.951 and 0.961 against omissions at 0.551 and 0.683, with the restatement-trace case at chance in every replicate. One difference between the families is genuine. The second family's completeness-scored design clears its own noise on single notes in all three replicates (19.1 to 20.5\% detection at 0.9 to 3.6\% false alarms), which no design in the primary family did. So single-note detection is judge-dependent and the single-note claims above are scoped to the family measured; what replicates is the asymmetry and its conditioning. One caveat: the second family's endpoint refuses a no-reasoning setting, so family and reasoning budget move together in this check (Section 8). Only restructured designs break the pattern (Section 7).

\subsection{The commission arm holds on physician-labelled data}

One arm of the asymmetry can be checked against independent labels, written by physicians with no connection to this study. MEDEC is physician-labelled clinical text, each item either error-free or carrying exactly one error with its corrected version supplied; its five error types are all substitutions, so it contains no omissions to detect, and it has no source document, so the prompt's transcript slot carries a no-source placeholder. Re-running the best of the eight on a 300-text stratified sample frozen before judging (156 errored texts and 144 error-free, each errored text paired with its own corrected version) gives paired 0.827 and AUC 0.811, with 51.3\% of the 156 errored texts detected at a 10\% false-alarm rate on the 144 error-free ones. The matched contrast with the same design on our omissions is the AUC pair, 0.811 against 0.575 (8.3\% detection at 6.5\% false alarms), so the commission side is externally anchored. What MEDEC anchors is the judge's ability to spot a clinically wrong statement with no source to check it against - the closest external check available for the commission class, and not the same task as transcript verification. The same labels test the widest remedy of Section 6 in advance. Two prompts identical except that one adds '' AND complete'' and the omission clause produce real-looking activity, with detection rising from 87.8\% to 96.8\% (p=1.2e-4) and false alarms from 52.8\% to 72.9\% (p=4.9e-6) over the 156 and 144 texts. Underneath, nothing separates better: AUC 0.800 against 0.811, and detection at a matched 10\% false-alarm rate identical to three decimal places. The omission arm has no such external anchor, for a reason we give in Section 8.

\subsection{Human reviewers do not share the asymmetry}

The comparator the literature supplies points the other way. In a planted-error study of AI-drafted patient messages, reviewing physicians caught omissions at roughly the same rate as objective errors, about a quarter to a third of each \citep{biro2025opportunities}. We did not run a human baseline of our own, and that study is a different task on a different corpus, so the comparison is indicative only (Appendix I collects every published human figure we could find). Read with that caution, the asymmetry appears to be characteristic of the machine-judge setting rather than an unavoidable property of review: human reviewers miss omissions too, but not selectively.

\section{Attempted fixes: five remedies move the operating point, none restores usable detection}

We tested five families of remedy - what the judge checks, how it answers, how many votes it gets, how the prompt is worded, and letting an optimiser rewrite the prompt. Every one shifted the judge's operating point along the same weak curve without producing usable single-note separation; what follows reports the size of each shift, states for the searched remedy the largest held-out gain the data cannot exclude, and does not say that no other remedy would work. All five are measured at the compute settings Section 4 records, and the optimiser searched inside that same budget. Two further levers sit outside them: the judge model family, which Section 5 showed shifts absolute levels while leaving the asymmetry in place, and the reasoning budget, which Section 7 reports.

\subsection{Scope, format and voting move the point, not the curve}

\begin{table*}[!htbp]
\centering
\footnotesize 
\setlength{\tabcolsep}{4pt}
\renewcommand{\arraystretch}{1.0} 
\begin{tabular}{@{}p{0.21\linewidth}p{0.19\linewidth}p{0.35\linewidth}p{0.18\linewidth}@{}}
\toprule
Lever & Paired delta on omissions & What it did to the operating point & Measured on\\
\midrule
What the judge is told to check: faithfulness only, to faithfulness and completeness & +0.036 to +0.082 (mean +0.052) \newline \emph{four per-design deltas, holding format and sample count fixed} & Detection rises in all four comparisons, from 1.0-9.6\% before to 4.7-42.3\% after, and false alarms rise with it, 1.5-8.9\% to 5.7-32.1\%; of the eight designs only the two yes/no completeness ones have detection that clears their own false alarms on pooled judgements (z 2.71 and 3.25; 1.24 and 1.65 once repeat runs collapse to a per-note majority, the paper's unit of inference), and they do it at 31.8\% and 32.1\% false alarms. & the eight designs, 495-pair evaluation set, 3 replicates (879 omission records, 336 clean records) \\
\addlinespace
How it answers: a yes/no verdict, to a 0-to-10 score & +0.043 to +0.073 (mean +0.053) \newline \emph{four per-design deltas, holding criterion and sample count fixed} & Both rates fall together, detection from 9.6-42.3\% to 1.0-8.3\% and false alarms from 7.4-32.1\% to 1.5-6.5\%, so the scored designs are quieter rather than sharper: their detection never clears their own false alarms by more than 1.02 standard errors. & the eight designs, 495-pair evaluation set, 3 replicates (879 omission records, 336 clean records) \\
\addlinespace
How many times it is asked: once, to eight times combined by a winsorised mean & +0.009 to +0.052 (mean +0.036) \newline \emph{four per-design deltas, holding criterion and format fixed} & The operating point barely moves: the largest change is the completeness-scored pair, detection 4.7\% to 8.3\% at false alarms 5.7\% to 6.5\%, and no scored design's detection separates from its own false alarms at either sample count. & the eight designs, 495-pair evaluation set, 3 replicates (879 omission records, 336 clean records) \\
\addlinespace
How the prompt words the omission instruction: the deployed faithfulness judge in three wordings, as shipped, with its omissions-are-not-errors exclusion deleted, and with an affirmative instruction to report omissions added & +0.051 (one comparison, not a span) \newline \emph{the affirmative instruction against the same prompt with the exclusion merely deleted; paired 0.518 as shipped, 0.505 deleted, 0.556 affirmative} & The affirmative instruction takes detection from 13.1\% to 25.6\% (77 and 150 of 586 omission judgements, against 16.4\% as shipped) while false alarms hold at 27 of 224 clean judgements (12.1\%) for all three wordings, the same rate though not the same notes; deleting the exclusion on its own takes paired discrimination from 0.518 to 0.505 and detection from 16.4\% to 13.1\%. & three wordings of the deployed faithfulness judge, 495-pair evaluation set, 2 replicates (586 omission records, 224 clean records) \\
\addlinespace
Letting an optimiser rewrite the prompt: reflective prompt evolution, aimed at the per-fact checker and at the best monolithic judge & +0.052 and +0.073 on the acceptance sets, -0.023 and +0.004 held out \newline \emph{two search legs, each measured on the 20-consultation acceptance set that chose the winner and then once on the 47-consultation held-out set} & Held out, the per-fact checker's winner gains 0.8 points of detection (22 against 21 of 131 omissions) at the same 8.5\% false alarms, and the monolithic winner loses 4.6 points (11 against 17 of 131) at the same 6.4\%. The earlier campaign's winner carried a false-alarm rate of 9.1\% (2 of 22 clean development notes) to 21.9\% (49 of 224) on data it had never seen, at 23.0\% detection (135 of 586). & acceptance set 48 omission pairs and 20 clean twins; held-out set 131 omission pairs and 47 clean twins, one replicate; the earlier campaign's winner on the 495-pair evaluation set, 2 replicates \\
\bottomrule
\end{tabular}
\caption{\textbf{What each lever bought, and what it did to the judge's operating point.}
Paired discrimination is the tie-adjusted probability that an errored note scores below its own
verified-clean twin, so 0.500 is a coin flip and a lever worth $+0.04$ buys four correctly
ordered pairs in every hundred, only where a clean twin exists to compare against. The first
three levers are quoted as the span of four per-design deltas, each holding the other two
choices fixed, because a single headline delta would hide how far the spans overlap: criterion
and format are indistinguishable on the mean ($+0.052$ against $+0.053$) and ensembling is smallest ($+0.036$).
Every paired figure here is tie-adjusted, because a finer aggregate splits ties and would
otherwise inflate a strict win rate. The completeness-scored pair shows why. Asking eight times rather than once takes the
strict win rate on omissions from 27\% to 43\%, but over the same 293 pairs the tie rate falls from 63\% to 41\%, and once half the tie mass is counted the paired figure moves 0.585 to 0.634. That is a lever worth $+0.049$, not sixteen points. Every paired figure in this paper is tie-adjusted for that reason.}
\label{tab:levers}
\end{table*}

\begin{figure}[!htbp]
\centering
\includegraphics[width=\linewidth]{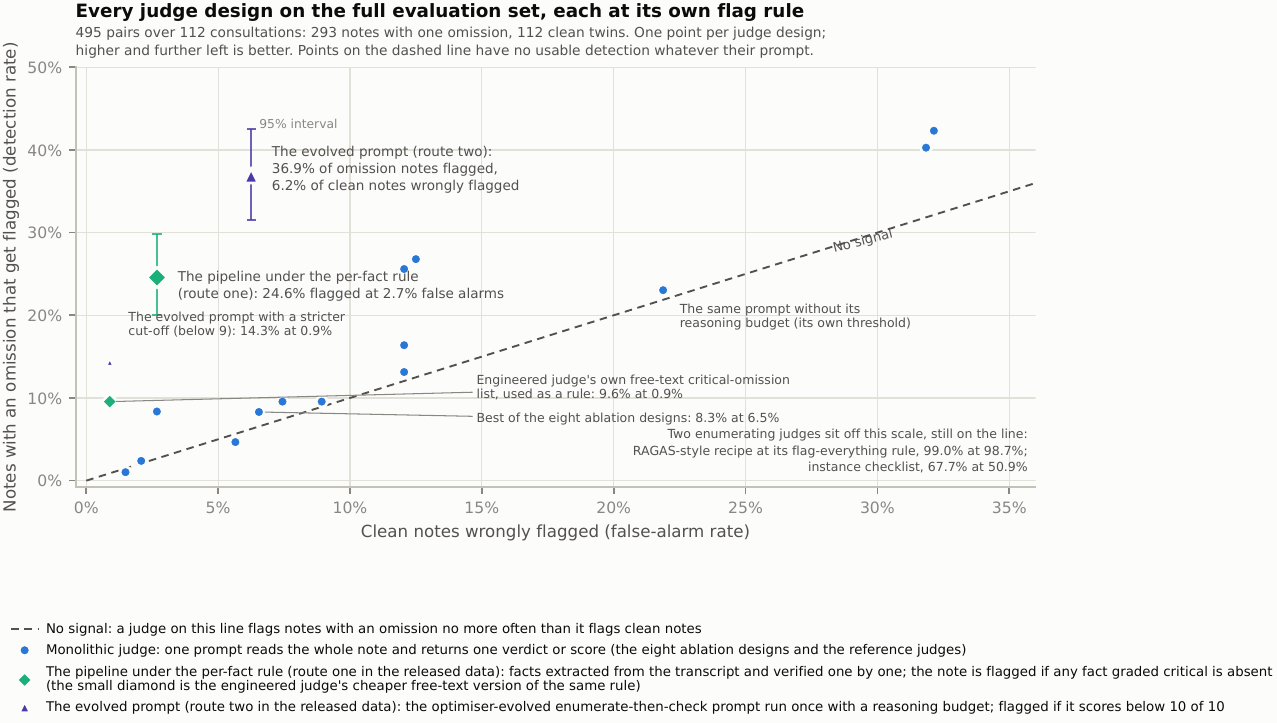}
\caption{\textbf{Every judge design at its own flag rule: the levers slide a point along the
no-signal line, and only the two methods leave it.} Across, how often a judge wrongly flags a
clean note; up, how often it flags a note that really carries an omission. A useful detector
sits high and to the left; the dashed diagonal is no signal, a judge that flags both kinds of
note equally often. Every design is on the 495-pair evaluation set (293 notes with one
omission, 112 clean twins; the eight designs pooled over three replicates, 879 and 336
records, the reference judges over two, 586 and 224, and the two method points note-level),
each at its own flag rule -
fixed before the run for the eight designs and the reference judges, established post hoc for
the two methods' rules (Sections~7.5 and~7.6). The eight designs and the deployed-judge wordings
sit on or beside the line, and the levers of Section~6 only slide a point along it. Two points
sit well clear of it, each with a 95\% whisker on its detection rate: the pipeline under the
per-fact rule (route one in the released data), at 24.6\% detection on 2.7\% false alarms, and
the evolved prompt (route two), at 36.9\% on 6.2\%. Three further points are labelled for
reference: the evolved prompt with a stricter cut-off, the same prompt run without its
reasoning budget, and the engineered judge's own free-text critical-omission list read as a
rule. Two enumerating judges flag almost every note and so fall outside the plotted
range; their rates are printed inside the figure. Every unlabelled point's values are in
Table~\ref{tab:grid}, Table~\ref{tab:levers} or Table~\ref{tab:practitioner}; how the
pipeline's flag is made from its verdicts is Figure~\ref{fig:verdictrule}.}
\label{fig:relocation}
\end{figure}

The three choices of Section 4 are the first three levers, and their effects are modest and additive (Table~\ref{tab:levers}, Figure~\ref{fig:relocation}). Criterion scope has the widest span of per-design deltas, +0.036 to +0.082 paired, output format is indistinguishable from it on the mean, and asking eight times instead of once is the smallest and least reliable lever (Table~\ref{tab:levers} has each span). Together the three take paired discrimination on omissions from 0.500 to 0.634 while the deployment measure stays at noise throughout.

\subsection{One wording change helps, and it is an instruction rather than a deletion}

A natural experiment answers whether wording can do better. The deployed faithfulness judge of Section 4 (Appendix H) includes the line ``omissions are NOT errors''. Three versions differing only in how they treat omissions - as shipped, with the exclusion deleted, and with an affirmative instruction to report omissions added - score paired 0.518, 0.505 and 0.556, with single-note detection 16.4\%, 13.1\% and 25.6\%. Deleting the exclusion does not help, and if anything hurts. The whole effect belongs to the one affirmative instruction, worth +0.051 paired over the version with the exclusion deleted, the same order as the widest lever above. All three flag 27 of 224 clean-note judgements, the same rate though not the same notes, so this is the one place in the study where a wording change moved detection at no measured cost in false alarms - though at a 12.1\% false-alarm rate none of the three wordings is a usable detector. It is still a 0.556, closer to a coin flip than to the 0.89 the same judge reaches on content that is present.

\subsection{Does prompt optimisation help?}

If hand-wording is the limit, the next step is to let an optimiser search. We used reflective prompt evolution (GEPA) \citep{agrawal2026gepa}, which runs a judge prompt over training examples, reads natural-language traces of what it got wrong, rewrites the prompt, and keeps a rewrite when it scores better. We ran three campaigns: an exploratory first pass, whose winner is the prompt Section 5 scores at 0.549; a second, aimed at the best of the eight designs, which accepted one mutation that scored below its own seed and produced nothing usable; and the confirmatory campaign we rest on. That campaign split its data by consultation, 45 to learn from, 20 as the acceptance set on which candidates were kept or rejected, and 47 held out for a final test committed before the search began and touched once, at the end (Appendix B carries all three in full).

No gain survived the held-out test. Aimed at the pipeline's per-fact checker, where a gain looked most plausible, the winner gained 5.2 points of paired discrimination on the acceptance set, then came back 2.3 points worse than the unoptimised baseline held out, on an interval, {[}-7.4, +2.7{]}, that excludes held-out gains larger than about 2.7 points. Aimed at the best monolithic judge, the winner cleared a decision rule fixed before the run and the held-out confirmation returned +0.004 paired, a fifteenth of the roughly ±0.06 of noise a single baseline run carries on these pairs. The detail that matters most for anyone who uses an optimiser: the first target's acceptance-set gain was not significant even on its own acceptance set, at 8 wins against 3, p=0.23. The held-out test did not overturn a real effect; it declined to confirm a difference never distinguishable from noise on the very sample used to accept it. Here a 20-consultation validator handed us an improvement a 47-consultation held-out set did not see.

One setting underlies all three campaigns and bounds how far their null reaches. Candidate prompts were executed at the judge settings of Section 4, and only the model writing the rewrites was given room to think. That matters because the exploratory campaign's winning prompt has as its central instruction to enumerate the transcript's facts first and then check them one by one, a procedure that takes room to carry out, and no candidate in any of the three searches ever had that room. Re-run unchanged on the held-out subset with a reasoning budget, that prompt moves from 0.546 on that subset to 0.669 paired on omissions and acquires a usable flag rule. Section 7 takes it up as one of the study's two working detectors, the evolved prompt. So the nulls are consistent with a search confined to a budget in which the strategy it kept proposing could not run, as much as with any limit of monolithic judging. A decision about how the evaluation was run, invisible in everything the search reported, determined what it appeared to find. Two bounds stand: the wording null holds at the settings the searches scored under, and restoring the budget to an already-selected prompt is not the same experiment as searching at that budget, which we did not run.

\subsection{Why: an omission leaves nothing to point at}

The pattern has a mechanism behind it, and it is not our inference alone. AbsenceBench identified it for language models in general - transformer attention cannot easily attend to gaps, because an absence supplies no keys to attend to - and confirmed it by inserting placeholder markers that make each gap visible in the input, recovering much of the loss \citep{fu2025absencebench}. An omission in a note is the same problem at the evaluation layer: nothing for an evidence-seeking judge to cite, where a commission leaves the offending text in hand. Every remedy above changes how hard the judge looks or how it reports; none creates the thing it would look at, and the placeholder repair is unavailable to an evaluator, because knowing where the gap is would already be the answer. Consistent with this, in the census's verification panel the survival of candidate findings tracks how objectively checkable a claim is (wrong dose 32.2\%, fabrication 24.6\%, omission 6.1\%). The optimisation campaigns make the same point from the other side: the search, asked to fix the judge and free to write anything, wrote enumeration into its winning prompt - the same conversion of absence into presence checks, found rather than designed.

\section{What works: convert the absence question into presence checks}

This section reports what recovers detection, with its size stated first: the best method here catches a little over a third of notes with an omission at a usable false-alarm rate, an improvement rather than a solution. One measurement in the remedy sweep points somewhere different. Standard evaluation tooling already contains a per-fact recipe: extract the source document's facts once, check each one against the text under evaluation, flag anything not fully covered. The RAGAS-style recipe of Section 4, our own deliberately naive implementation in that lineage, is unusable as shipped, checking roughly 62 unaudited facts per consultation that no real note carries and so flagging 98.7\% of clean notes. The score underneath is another matter: swept to a calibrated threshold it reaches 20.9\% detection at 10.0\% false alarms against 8.3\% at 6.5\% for the best of the eight designs, and its coverage component alone reads 0.817 paired on omissions, past every monolithic judge on the same pairs. The omission signal appears as soon as the judge has a list of concrete facts to check, and the aggregation into a note-level flag destroys it, so the problem has two separable bottlenecks: producing a good enough fact list, and preserving the per-fact evidence all the way to the decision.

\begin{figure}[!htbp]
\centering
\includegraphics[width=\linewidth]{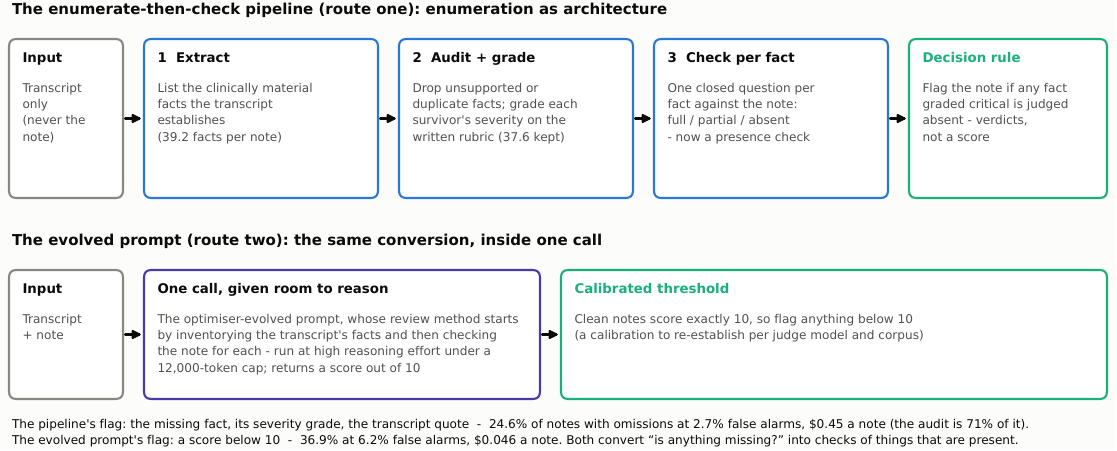}
\caption{\textbf{The two methods, as run.} The enumerate-then-check pipeline (route one in
the released data) converts the absence question into presence checks in the architecture: an extraction call that never sees the note, an
audit that drops weak facts and grades the rest, one closed check per fact, and a
decision rule over the verdicts. The lane draws the three-stage tier, the only one whose
audit grades severity and so the only one the decision rule can be stated on; the
two-stage tier stops after the per-fact check and carries the higher paired score. The
three-stage tier also runs a second look over the facts the check called absent, which
the lane does not draw (Appendix C). The evolved prompt (route two) makes the same conversion
inside one call: a GEPA-evolved prompt whose review method starts by inventorying the transcript's facts
and then checking the note for each, executed at high reasoning effort and read through a
clean-note calibration. Operating points are the evaluation-set figures of
Table~\ref{tab:practitioner} and prices those of Table~\ref{tab:cost}; the fact counts
inside the pipeline's lane are the held-out confirmation run's, 39.2 facts extracted per
note and 37.6 left after the audit.}
\label{fig:routes}
\end{figure}

Two methods convert the absence question into presence checks, and arrived at it independently (Figure~\ref{fig:routes} draws both; the released data labels them route one and route two). \textbf{The enumerate-then-check pipeline}, the pipeline for short, builds it into the architecture, with an extraction call that never sees the note, a severity-graded audit, and a closed present-or-absent verdict per fact. \textbf{The evolved prompt} does it inside one call. It is the exploratory campaign's winning prompt from Section 6, which becomes a working detector once given the budget to execute its own enumerate-then-check instruction.

\subsection{The pipeline: do at judging time what construction did}

Construction in Section 3 faced the judge's problem and never asked ``is anything missing?''. It enumerated the facts the transcript establishes and checked each one. Fact enumeration for omission measurement is not new - MED-OMIT built it for medical summaries in 2023 \citep{schumacher2023medomit} - but what has not been shown is that it recovers judge detection where every monolithic remedy fails, and what to do with the verdicts. Doing it at inference time gives a pipeline: extract the clinically material facts from the transcript alone, never from the note, so blindness is structural rather than promised, then verify each fact against the note as a closed lookup. \textbf{The two-stage pipeline} (B2 in the released data and in this paper's tables) stops there, making one keyed present-or-absent call per note. \textbf{The three-stage pipeline} (B3) adds a critic audit that drops unsupported or duplicated facts, grades each survivor's severity against the rubric of Section 3, and runs a quote-verified second look over anything flagged absent. Neither tier fixes a flag threshold in advance. Every operating point below is swept and quoted with its own false-alarm rate (Appendix C).

\subsection{Held out, enumeration works, and a blinded physician agrees}

\begin{table}[!htbp]
\centering
\small
\setlength{\tabcolsep}{4pt}
\begin{tabular}{@{}p{0.30\linewidth}p{0.62\linewidth}@{}}
\toprule
What was judged & Result\\
\midrule
Pipeline against monolithic judge, on the 10 notes
where they disagree & the physician's verdict matched the three-stage pipeline
100\% [72, 100] and the monolithic judge
0\% [0, 28]; $p=0.002$.
Restricted to the 8 notes where the monolithic
judge's flag is attributable to the fact in front of them,
100\% [68, 100] against
0\% [0, 32], $p=0.008$\\
\addlinespace
Per-stage validation, 34 items & inference-time fact extraction 10/10 = 100\% [72, 100]; reference-note repair 8/8 = 100\% [68, 100]; omission-pair rejection 6/6 = 100\% [61, 100]; taxonomy panel cuts 0/10 = 0\% [0, 28]. The taxonomy row is a false-kill rate; two of its ten items are
pack-caused abstentions, so its honest denominator is 8 (0\% [0, 32.4], Appendix E); the first three are endorsements of a stated verdict, not blind draws\\
\addlinespace
Severity grading, 20 facts graded blind & exact agreement
70\% [48, 86], 85\% weighted,
linear-weighted $\kappa=0.63$ (bootstrap 95\%
0.32--0.86); all
6 disagreements are one grade, none two\\
\bottomrule
\end{tabular}
\caption{\textbf{One physician author, 70 items in a
single 58-minute sitting} - the 64 scored items above, plus three realism ratings and
three free-text answers; blinding varies by stage as stated. Wilson 95\%
intervals throughout, because every section holds 6 to 20 items and the interval is the
result. The rater is an author and designed the instruments they are grading: the adjudication,
the severity grading and the taxonomy panel cuts were blinded structurally - the rater was
shown no indication of which judge said what or whether a note had been edited, no machine
severity grade and no sign of which items the two graders split on, and no name for the
bucket a cut finding came from - while the extraction, repair and pair-rejection
rows state a verdict the item has to carry to be answerable, and should be read as a
clinician checking and not objecting. The taxonomy panel is the companion census's
verification panel, and a cut is a candidate finding it refused. The severity row validates the dataset's severity
axis, not the audit stage's own grades that the per-fact rule fires on.}
\label{tab:sitting}
\end{table}

On 151 held-out pairs over 47 consultations (131 omissions and 20 commissions, against 47 clean twins), the two-stage pipeline reaches 0.786 paired discrimination on omissions as a mean of three runs spanning 0.771 to 0.801, against 0.531 to 0.649 for the best monolithic judge's three runs on those same pairs. The pipeline's worst run clears the baseline's best by 0.12, the means differ by +0.19, and replicate one's difference of +0.20 has a 95\% interval of {[}0.14, 0.27{]} with each consultation counted as one observation; the pipeline orders more of its pairs correctly than the monolithic judge on 25 of the 47 consultations and fewer on 7 (p=0.002). The three runs also measure stability. The two-stage tier's span is 0.031 and the three-stage tier's 0.015, against 0.118 for the monolithic judge on the identical pairs, so the arm under test shows a quarter of the run-to-run spread of the baseline it is measured against.

The paired figure replicated from the exploratory pass, 0.828 to 0.786; the single-note figure did not, halving from 31.2\% at 9.4\% false alarms to 16.0\% at 8.5\%, and 16.0\% is the figure we report. On those same pairs the RAGAS-style recipe, sharing no prompt text with the pipeline, reaches 0.801 through its coverage component - matching the top of the pipeline's own three-run range - though it has no usable operating point under its natural rule and no severity grades.

The strongest objection is circularity: the pipeline resembles the benchmark's own construction, both enumerating facts from the transcript, so its win could be alignment between construction and judging rather than a property of the task. The objection has a model-level form too, since the pipeline's extraction stage runs on the pinned model that built the fact sheets and its audit on the model that verified the pairs (Appendix D). Three lines of evidence answer both forms. The RAGAS-style recipe lands inside the pipeline's own run-to-run range with prompts of its own and with its facts extracted by the judge model, not by the model that built the answer key (Appendix D), and concurrent work at another institution reports the same monolithic null and threshold-shift pattern on a different clinical task \citep{delucia2026same}. The extraction instrument, run blind, recovers 99.4\% of facts authored by humans before any pipeline existed, on the sheets its own critic panel kept. And a blinded physician adjudicated real disagreements between the three-stage pipeline and the best monolithic judge (Table~\ref{tab:sitting}).

Ten notes where the two reach opposite conclusions went to a physician author, drawn deterministically by seed, one per consultation, from the confirmation run's 35 disagreement notes and stratified for coverage rather than by expected outcome. The rater was not told which judge said what or whether the note had been edited, and abstention was offered as a real answer. They abstained zero times and sided with the pipeline on 10 of 10 ({[}72.2, 100{]}, p=0.002), and 8--0 on the subset where the monolithic judge's flag is attributable to the fact in front of them. In six of the ten the pipeline had named the fact construction removed, so those six read as the monolithic judge missing a verified absence rather than as an open contest; the other four are the three the monolithic judge flagged and the rule's one clean-note flag. Six of the ten put this paper's claim in front of a clinician on a whole consultation note - a missing clindamycin allergy, a missing positive Murphy's sign - and Figure~\ref{fig:worked} walks one of them end to end. This is a conditional result on selected disagreements and the rater is an author; read it with both facts in view. It was one of three structurally blinded stages in a 70-item sitting (Appendix E).

\subsection{What creates the signal: the list is a third of it, the closed verdicts the rest}

Two controls locate the mechanism, separating two ideas that sound alike. G-Eval \emph{decomposes the judgement}, splitting scoring into abstract criteria but still asking each as an open question over the whole note, and it performs like the monolithic judges, 0.568 on omissions against 0.890 on commissions. The checklist judge goes halfway to \emph{enumerating the facts}, generating a roughly 25-item checklist from the transcript and answering each item, and on the evaluation set it falls between at 0.647, against the monolithic best of 0.634 and the \textasciitilde62-fact RAGAS-style recipe's coverage component at 0.817. What creates the signal is listing the concrete facts to check rather than splitting the judgement into criteria, and a fuller list helps only up to a point: the \textasciitilde25-item checklist reads 0.647 against the \textasciitilde62-fact recipe's 0.817 on the same evaluation set, while on identical held-out pairs the recipe's \textasciitilde62 facts land inside the run-to-run range of the two-stage pipeline's \textasciitilde39 (Section 7.2), so beyond a reasonably complete list its length stops mattering.

The control that separates the two: give the same monolithic judge the pipeline's own audited fact list in its prompt, but keep the single open score. On these held-out pairs, handing the judge the list moves paired omissions from 0.570 to 0.634 in one run, against a control whose own three runs span 0.508 to 0.603 - a gain that clears that control and still stops inside the best monolithic judge's own three-run spread on these pairs, 0.531 to 0.649. (That 0.634 is a held-out figure and shares its digits, by coincidence, with the evaluation-set best of Section 5.) A list of the same kind read through \textbf{a closed per-fact output structure} (a present-or-absent verdict keyed to each fact, rather than one score over the note) reaches 0.786; Section 7.4 gives the two list sizes. Of the gap between a plain monolithic judge and the pipeline, having the list accounts for roughly a third and the closed verdicts for the rest. The fact block is a real prompt change, so part of that gain could be the wording of the block itself; the criterion and the answer format were held identical between the two prompts, which limits how much wording could account for without excluding it. Even holding the list, the judge compresses all 198 held-out notes onto four distinct score values: a list does not stop the compression, it only changes what is compressed. Where AbsenceBench's placeholder repair fixes the input representation, here the input is the smaller part of the fix. The restructuring survives the change of judge family, with the rule transferring at almost the same operating point and one tier reversal, in full in Appendix G.

\subsection{Deliberation is a real lever, and it is not the mechanism}

\begin{table}[!htbp]
\centering
\footnotesize
\setlength{\tabcolsep}{4pt}
\begin{tabular}{@{}>{\raggedright\arraybackslash}p{1.95in}lccl@{}}
\toprule
Prompt (unchanged within each block) & reasoning budget & runs & paired, omissions & change against its own control \\
\midrule
Faithfulness and completeness, score, one call & none, 1{,}024 & 3 & 0.570 & reference \\
 & medium, 12{,}000 & 1 & 0.641 & +0.071 [+0.009, +0.134] \\
 & high, 12{,}000 & 3 & 0.670 & +0.101 [+0.048, +0.153] \\
\addlinespace
The same, with the pipeline's audited fact list in the prompt & none, 1{,}024 & 1 & 0.634 & reference \\
 & high, 12{,}000 & 1 & 0.699 & +0.065 [$-$0.016, +0.145] \\
\addlinespace
The engineered omission judge & none, 1{,}024 & 2 & 0.635 & reference \\
 & high, 12{,}000 & 1 & 0.649 & +0.013 [$-$0.047, +0.071] \\
\addlinespace
The optimiser's exploratory winner (the evolved prompt, once budgeted) & none, 1{,}024 & 2 & 0.546 & reference \\
 & high, 12{,}000 & 3 & 0.669 & +0.123 [+0.059, +0.189] \\
\midrule
Two-stage pipeline, closed per-fact verdicts & one call per fact, none & 3 & 0.786 & not a monolithic judge \\
\bottomrule
\end{tabular}
\caption{\textbf{What a reasoning budget buys each prompt, on the 151-pair held-out
subset.} One set of pairs throughout: 131 omission pairs and 20 commission pairs over 47
consultations, with 47 clean twins. Within each block the prompt is identical, character for character, between
the control row and the rows beneath it, and the judge model, temperature, seed, notes and
flag rule are held, so the reasoning setting and the token cap are the only quantities that
move. Paired discrimination is the tie-adjusted probability that a note carrying an omission
scores below its own verified-clean twin, where 0.500 is a coin flip; a value here is the
mean over the runs in the ``runs'' column. Intervals are 95\% bootstrap intervals on the change,
resampling whole consultations, 10{,}000 resamples. Four rows rest on a single run and should
be read against their control's own run-to-run spread, which on this subset is 0.508 to 0.603
for the completeness-scored single-call design. Reading: the budget is worth about 0.10 of
paired discrimination to a plain criterion and to the optimiser's winner, nothing to the
engineered omission judge, and the strongest monolithic judge it can build, with the fact list and the
budget together, still sits about 0.09 below the pipeline reading a list of the same kind
through closed per-fact verdicts - the audited 37.6-fact list in the prompt against the
two-stage tier's own 39.2 unaudited facts (0.786 as a mean over three runs spanning 0.771
to 0.801).}
\label{tab:compute}
\end{table}

Every control so far holds the judge's compute fixed, leaving one alternative reading: the pipeline spends a call per fact where a monolithic design spends one short answer, so the difference could be thinking time rather than task structure. We tested it on the same 151 held-out pairs by giving the monolithic prompts a reasoning budget and changing nothing else (Table~\ref{tab:compute}).

Compute is a real lever, and what it is worth depends on the prompt. Raising the plain completeness-scored judge from no reasoning effort to high lifts paired discrimination on omissions from 0.570 to 0.670, +0.101 {[}+0.048, +0.153{]} and monotonic in dose; it gives the engineered judge nothing at all, +0.013 {[}-0.047, +0.071{]}, and the evolved prompt the most, +0.123 {[}+0.059, +0.189{]}.

The strongest monolithic judge we can assemble with both advantages at once - the pipeline's own audited fact list in the prompt and a high reasoning budget - reaches 0.699, still about 0.09 below the two-stage pipeline's 0.786 on the same pairs, from a list of the same kind: the audited 37.6-fact list against the two-stage tier's 39.2 unaudited facts. It costs about three times as much per note as the pipeline's checking stage. So the budget on its own closes roughly half the distance between the plain completeness-scored judge and the pipeline (0.570 to 0.670, against the pipeline's 0.786) and then stops, and what it is worth varies by an order of magnitude across prompts, so it is a real lever rather than a substitute for restructuring the task.

Two things do not improve when compute is added, and one of them gets worse. Paired discrimination on added and altered content rises with the budget for every prompt given one, on the 20 commission pairs in this subset (enough to read as a direction, not a size), so the presence-versus-absence asymmetry widens with compute instead of closing. And the omission whose restatement survives elsewhere stays at chance throughout, 0.495 to 0.546 paired across the four prompts Table~\ref{tab:compute} reports.

\subsection{Act on the verdicts, not a score}

\begin{figure}[!htbp]
\centering
\includegraphics[width=\linewidth]{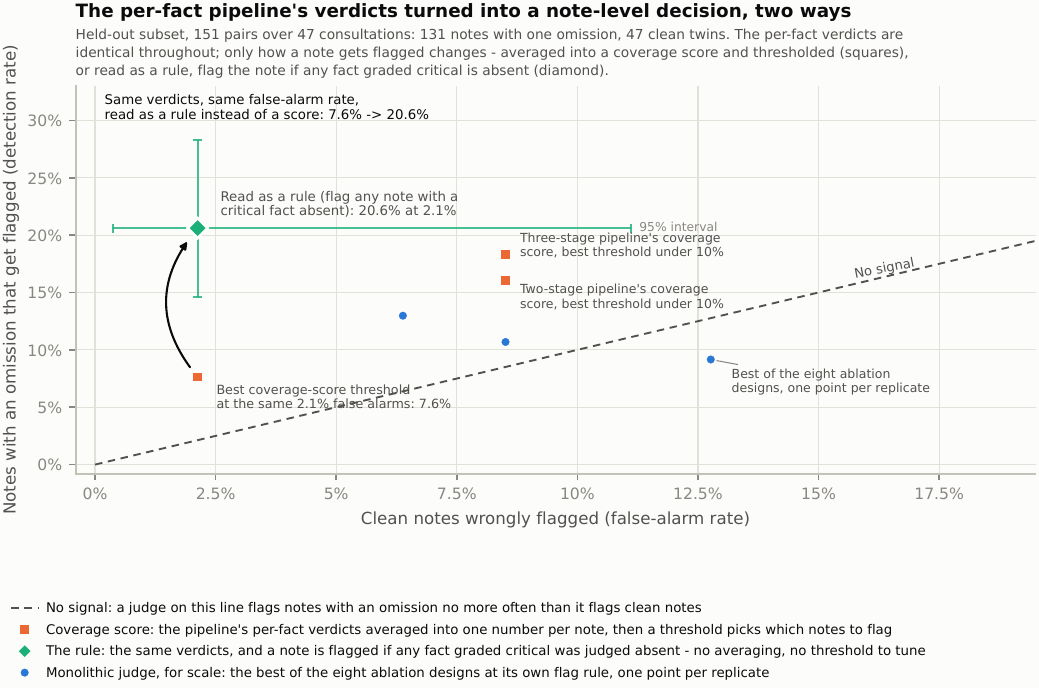}
\caption{\textbf{The same per-fact verdicts, turned into a note-level decision two ways.} A
pipeline that returns one verdict per fact still has to decide, per note, whether to flag it.
On the 151-pair held-out subset (131 notes with one omission, 47 clean twins) the three-stage
pipeline's verdicts are read two ways with nothing else changed. Averaged into a coverage score
and thresholded at the rule's own 2.1\% false-alarm rate, they detect 7.6\% of the notes with
an omission; the best threshold inside a 10\% false-alarm budget reaches 18.3\% at 8.5\% for
the three-stage tier and 16.0\% at 8.5\% for the two-stage tier. Read as the rule - flag the
note if any fact graded critical is judged absent - the identical verdicts detect 20.6\% at
2.1\% (the plotted rule point is replicate 1 of three; across the three the detection reads
20.6, 22.9 and 21.4\%, with the false-alarm rate identical in every one). The best of the
eight ablation designs is drawn for scale at its own rule, one point per replicate. The
identical verdicts, read as a rule rather than averaged into a score, give 2.7 times the
detection at the same false-alarm rate.}
\label{fig:verdictrule}
\end{figure}

The three-stage pipeline produces a graded verdict per fact and still has to make a note-level decision, and the conventional move, averaging the verdicts into a coverage score and thresholding it, is the weak way to use them. Clean notes average 0.953 coverage and notes with omissions 0.929, so every threshold cuts inside a 2.4-point sliver, because one missing fact among about 38 barely moves an average. The usable object is \textbf{the per-fact rule}, a rule over the individual verdicts: \emph{flag the note if any fact graded critical is judged absent}.

Held out, the rule catches 20.6\% of notes containing an omission (27 of 131, {[}13.8, 28.2{]}; replicates two and three read 22.9\% and 21.4\%, and a strict majority of the three lands at 21.4\% {[}15.2, 29.2{]}, so the figure we report is the lowest of the three) and 32.8\% of those carrying a critical-severity omission (19 of 58; 37.9\% and 34.5\% in the other two runs), at a false-alarm rate of 2.1\% (1 of 47, {[}0.4, 11.1{]}) - the same single clean note in all three runs, so the rule's quietness belongs to the rule and not to one seed. The separation from its own noise is significant with the consultation as the unit of inference (p=2e-06) and holds on the uniformly built omission subset alone (16.7\%, 15 of 90, p=0.008).

At that same noise level the best threshold on these \emph{same verdicts'} aggregate score manages 7.6\% (Figure~\ref{fig:verdictrule}). The rule delivers 2.7 times the detection from identical evidence, and nothing was added to the judge, only the way its output is read. It was formulated after the confirmation run by re-analysing verdicts already recorded, then checked against the earlier exploratory pass, whose verdicts predate it, and held at 25.0\% at 6.2\%. Its one clean-note flag stands as a false alarm throughout, on a fact the physician author re-graded supporting rather than critical (Appendix E).

The rule also scales. Over the full 495-pair evaluation set - whose prompts are hand-written and were never optimised on anything, though 47 of its 112 consultations are the held-out subset the rule was formulated on - it catches 24.6\% of notes containing an omission (72 of 293, {[}20.0, 29.8{]}) at a 2.7\% false-alarm rate (3 of 112 clean twins, {[}0.9, 7.6{]}), and 39.7\% of the omissions graded critical (60 of 151); the 162 omission pairs the rule was not formulated on read 27.8\% on their own, so the pooled rate is not carried by the formulation subset. Table~\ref{tab:practitioner} sets out the per-design comparison behind that margin. Below 5\% false alarms exactly one aggregate threshold in the primary family separates its detection from its own false-alarm rate by more than chance, the engineered judge's 8.4\% at the same 2.7\% (z = 2.88), and the rule detects 2.9 times as much at an identical rate.

The rule needs one ingredient the cheaper pipeline cannot supply: a severity grade attached to every fact \emph{before any note is judged}, which the two-stage tier never produces, so the rule cannot be stated on its output at all. The three-stage tier's audit grades every fact against the rubric of Section 3, and that turns out to be what the stage is for: the audit was originally built around its full/partial/absent verdict, a diagnostic that did not replicate out of sample (Appendix C), and its severity grades proved to be the output that matters.

The choice between tiers is then costed rather than ranked. The two-stage tier has the higher paired score - 0.786 against 0.762 as three-run means, and in every individual run - at one fifth of the cost. The three-stage tier wins the deployment measure where the cheaper tier reads zero: swept as a coverage score inside the same 10\% false-alarm budget, it catches 12.1\% of restatement-trace omissions held out against the two-stage tier's zero, and 31.0\% of critical omissions against 20.7\%. Those are the tiers' swept scores. The per-fact rule itself catches no restatement traces at all (Section 7.8). Only the three-stage tier can express that rule.

The physician author of Section 7.2 re-graded ten of the facts the rule fired on, confirming critical on 1 of the 10 in the blind pass ({[}1.8, 40.4{]}) and on 6 of 10 ({[}31.3, 83.2{]}) after unblinded reconciliation - half the gap being context the fast blind pass had not weighed - and grading all ten clinically material. That does not move the rule's rates, measured as they are against constructed absence, but it does mean a flag's ``critical'' sat about one grade above this study's one clinician rater's own on routine content (Appendix E); Section 8 reads that lean as a property of the rater rather than of the rubric. The rule's quietness rests on the same thing. Its low false-alarm rate is a property of the audit stage's critical threshold, and a threshold re-set from clinician grades would move it: set at the rater's own critical line it would fire on fewer facts, and set at clinically material, which all ten were, on more. Re-scoring the stored verdicts under either is the cheapest next experiment in the study (Section 9).

\subsection{The evolved prompt: the same conversion inside one call}

The second method needs no pipeline machinery. The evolved prompt is the exploratory campaign's GEPA-evolved prompt of Section 6 (\texttt{gepa-04} in the release, and route two in the released data and the figures), whose central instruction is to enumerate the transcript's facts first and then check each one, re-run unchanged at high reasoning effort under a 12,000-token cap, three times with different seeds. Without the budget the same prompt has no comparable operating point, reaching 10.3\% detection at 9.6\% false alarms held out, swept to a 10\% ceiling, so ``the evolved prompt'' below always means this budgeted configuration. With room to execute its own instruction it does something no monolithic design in the primary family did at its fixed rule: it scores clean notes at the top of the scale. (The second family's plain completeness-scored judge does the same, and Appendix G reports the operating point that gives it.) On the 47 held-out clean twins the three runs score 45, 46 and 45 of them at exactly 10 out of 10, which turns ``flag anything below 10'' into a rule a practitioner can state. Under it the strict majority of three catches 32.1\% {[}24.7, 40.5{]} of the 131 held-out notes containing an omission at a 2.1\% false-alarm rate.

The full evaluation set is a fair test for this prompt too, since its development pool was carved out of the corpus entirely, and at 2.4x the sample the rule still works, at a noisier point. The majority-of-three rule catches 36.9\% of the 293 notes containing an omission ({[}31.5, 42.5{]}) at a 6.2\% false-alarm rate (7 of 112 clean twins, {[}3.1, 12.3{]}), and 58.9\% of the critical-severity omissions (89 of 151), against the per-fact rule's 39.7\% on the same notes. By trace level the flags follow the study's gradient: 53.3\% of complete removals (80 of 150), 27.9\% of fragment traces (24 of 86), and 7.0\% of restatements (4 of 57) - the case that defeats every deployable method here (Section 7.8).

Two properties of the method are measured rather than assumed. First, the threshold is a calibration and not a design guarantee. It was read off the first replicate's score distribution and then held unchanged across the other two seeds and the full evaluation set. It works only because clean notes score exactly 10, and that softens on the full set - 96 to 98\% of the held-out twins sat at 10 against 90 to 94\% of the evaluation-set twins, which is why the false alarms read 2.1\% and 6.2\% - so a deployment would have to re-establish it on its own clean notes. On the companion census's real vendor notes the calibration fails outright, with 55.8\% of panel-cleared notes scoring 10, and a usable point exists only after the threshold is re-established there (Appendix K). Second, the calibration transfers across judge families: on the second family the same prompt scores 43 to 45 of the 47 held-out twins at 10 and reads 34.4\% detection at 6.4\% false alarms. What the method does not buy is discrimination or explanation. Its paired score on omissions is 0.669 held out, well short of the pipeline's 0.786 on the same pairs, and 0.670 on the full evaluation set by the convention Table~\ref{tab:practitioner} uses throughout, and its flag is a number below 10 with nothing attached.

\subsection{The head-to-head, what each price buys, and the same methods on real notes}

\begin{table*}[!htbp]
\centering
\footnotesize
\setlength{\tabcolsep}{2.1pt}
\begin{tabular}{@{}>{\raggedright\arraybackslash}p{2.40in}ccccc@{}}
\toprule
& \multicolumn{2}{c}{Paired discrimination} & \multicolumn{2}{c}{At its own flag rule} &\\
\cmidrule(lr){2-3}\cmidrule(lr){4-5}
Approach & Omissions & Commissions & Detection & False alarms & \begin{tabular}[c]{@{}c@{}}Best detection at\\$\leq$10\% false alarms\end{tabular}\\
\midrule
\multicolumn{6}{@{}l}{\itshape One call at standard settings, asked an open question} \\[1pt]
Best of the eight grid designs (FC / score / k=8) & 0.634 & 0.939 & 8.3\% & 6.5\% & 8.3\% \\
Cheapest of the eight (F / score / k=1) & 0.549 & 0.911 & 1.0\% & 1.5\% & 1.0\% \\
Best yes/no design (FC / binary / k=8) & 0.591 & 0.869 & 42.3\% & 32.1\% & -- \\
Engineered omission judge (hand-written, 8 few-shots) & 0.639 & 0.754 & 8.4\% & 2.7\% & 8.4\% \\
Deployed faithfulness judge, as shipped & 0.518 & 0.886 & 16.4\% & 12.1\% & -- \\
Optimiser's exploratory winner, standard settings & 0.549 & 0.861 & 23.0\% & 21.9\% & 7.7\% \\
G-Eval (decomposes into dimensions, never enumerates) & 0.568 & 0.890 & 26.8\% & 12.5\% & -- \\
\midrule
\multicolumn{6}{@{}l}{\itshape Enumerate the transcript, read the result as a score} \\[1pt]
Instance checklist (in the style of \citealp{zhou-etal-2025-feedback}) & 0.647 & 0.744 & 67.7\% & 50.9\% & 16.0\% \\
RAGAS-style recipe (ours, deliberately naive) & 0.809 & 0.712 & 99.0\% & 98.7\% & 20.9\% \\
RAGAS-style coverage score alone & 0.817 & 0.665 & 99.0\% & 98.7\% & 20.9\% \\
Enumerate + check (pipeline B2)$^{\dagger}$ & 0.786 & 0.650 & -- & -- & 16.0\% \\
Enumerate + audit + check (pipeline B3)$^{\dagger}$ & 0.762 & 0.617 & -- & -- & 18.3\% \\
\midrule
\multicolumn{6}{@{}l}{\itshape The deployable operating points, each at its own rule} \\[1pt]
The pipeline under the per-fact rule (route one): flag if any critical fact is absent & 0.795 & 0.661 & 24.6\% & 2.7\% & 24.6\% \\
The evolved prompt (route two): the optimiser's winner with a reasoning budget, one call & 0.670 & 0.978 & 36.9\% & 6.2\% & 36.9\% \\
The evolved prompt on the second judge family (gemini-3.1-pro, same prompt)$^{\dagger}$ & 0.659 & 0.975 & 34.4\% & 6.4\% & 34.4\% \\
Free-text rival: the engineered judge's own critical-omission list & 0.639 & 0.754 & 9.6\% & 0.9\% & -- \\
\bottomrule
\end{tabular}
\caption{\textbf{What each approach delivers.}
Paired discrimination is the tie-adjusted probability that an errored note scores below its
own verified-clean twin; 0.500 is chance and it is a ceiling, since production has no clean
twin. Detection and false alarms are absolute, on single notes, at each design's own flag rule -
fixed before the run for the eight designs and the reference judges, established post hoc
for the two methods' rules (Sections 7.5 and 7.6). Unmarked rows are the evaluation set: 293 omission pairs, 202
commission pairs and 112 clean twins - the eight designs at three replicates, the reference judges at
two, and the last block's rows at the rules stated there (the per-fact rule one
replicate, the evolved prompt a majority over three). $^{\dagger}$ marks the held-out
confirmation subset: 131 omission pairs, 20 commission pairs and 47 clean twins; those rows
are not directly comparable to the others. The two coverage-recipe rows depart from
those counts in their paired columns only: those rest on 289 omission and 200 commission
pairs, because parse failures cost that arm 14 of its 1,214 records; their absolute columns
keep the full 586 omission and 224 clean records, counting an unparseable record as no flag.
The coverage score alone carries no flag rule of its own, so its detection, false-alarm and
swept cells repeat the recipe's in the row above it. The two pipeline-tier rows print paired means
over three replicates (Section 7.2 gives the spreads); their sweep column is replicate 1.
One denominator holds per set of items, and rows
measured on different sets are never ranked against each other. The two methods of Section 7 close the table at
their own deployment rules: the per-fact rule and the evolved prompt's flag-below-10 rule
both already sit inside the 10\% false-alarm budget, so the sweep column repeats their own
detection. The pipeline row's flag rates are the per-fact rule's, while its two paired columns
are the three-stage tier's own coverage score on the same evaluation-set pairs, one
replicate, since the rule returns no score to rank pairs by. Neither dominates: the evolved prompt buys the higher-detection point at 2.3 times the
rule's false-alarm rate and a tenth of its price, resting on a per-deployment
calibration; the pipeline buys the quiet point, and each of its flags names the missing
fact with an audit-stage severity grade - a grade that this study's one clinician rater read about one grade above their own on routine content (Section~8 reads that lean as a property of the rater rather than of the rubric). A dash means the quantity is not expressible for that
design - a binary or 5-point aggregate admits no threshold sweep, the free-text rival is a
rule over a list with no score to sweep, and the two pipeline tiers write no flag, so
they are swept rather than read at a rule. What each row costs and what its
flag contains are in Table~\ref{tab:cost}, over the same rows in the same order.}
\label{tab:practitioner}
\end{table*}

\begin{table*}[!htbp]
\centering
\footnotesize
\setlength{\tabcolsep}{2.1pt}
\begin{tabular}{@{}>{\raggedright\arraybackslash}p{2.30in}r>{\raggedright\arraybackslash}p{1.05in}>{\raggedright\arraybackslash}p{1.65in}@{}}
\toprule
Approach & \$/note & What a flag carries & Hidden cost\\
\midrule
\multicolumn{4}{@{}l}{\itshape One call at standard settings, asked an open question} \\[1pt]
Best of the eight grid designs (FC / score / k=8) & \$0.036 & A score & -- \\
Cheapest of the eight (F / score / k=1) & \$0.004 & A score & -- \\
Best yes/no design (FC / binary / k=8) & \$0.035 & A verdict & -- \\
Engineered omission judge (hand-written, 8 few-shots) & \$0.005 & A score + free text & -- \\
Deployed faithfulness judge, as shipped & \$0.005 & A verdict & -- \\
Optimiser's exploratory winner, standard settings & \$0.008 & A score & -- \\
G-Eval (decomposes into dimensions, never enumerates) & \$0.008 & A score & -- \\
\midrule
\multicolumn{4}{@{}l}{\itshape Enumerate the transcript, read the result as a score} \\[1pt]
Instance checklist (in the style of \citealp{zhou-etal-2025-feedback})$^{\ddagger}$ & \$0.006 & A score & One cached artefact per consultation, not ledgered \\
RAGAS-style recipe (ours, deliberately naive)$^{\ddagger}$ & \$0.030 & A score & One cached artefact per consultation, not ledgered \\
RAGAS-style coverage score alone$^{\ddagger}$ & \$0.030 & A score & One cached artefact per consultation, not ledgered \\
Enumerate + check (pipeline B2) & \$0.094 & A coverage score & -- \\
Enumerate + audit + check (pipeline B3) & \$0.45 & A coverage score & -- \\
\midrule
\multicolumn{4}{@{}l}{\itshape The deployable operating points, each at its own rule} \\[1pt]
The pipeline under the per-fact rule (route one): flag if any critical fact is absent & \$0.45 & The named fact + severity & -- \\
The evolved prompt (route two): the optimiser's winner with a reasoning budget, one call & \$0.046 & A score below 10 & A clean-note calibration, re-established per deployment \\
The evolved prompt on the second judge family (gemini-3.1-pro, same prompt) & \$0.028 & A score below 10 & A clean-note calibration, re-established per deployment \\
Free-text rival: the engineered judge's own critical-omission list & \$0.005 & The fact, free text & -- \\
\bottomrule
\end{tabular}
\caption{\textbf{What each approach costs, and what its flag contains.}
The rows and their order are those of Table~\ref{tab:practitioner}, which carries the same
approaches' discrimination and operating points. Cost is measured from the run
receipts in a production framing of one note per consultation, so per-consultation stages
are not amortised across a benchmark's twins; amortised across this benchmark's twins the
two pipeline tiers read \$0.028 and \$0.140 a note (the two ratios differ because the tiers
spend different shares of their price on per-consultation stages). The difference between the
two tiers' prices exceeds the audit stage's own 71\% share of \$0.45 because the three-stage
tier also runs its closed check at dearer settings and a second look over flagged facts.
What a flag carries is what the design
hands the reader when it fires: a bare score or verdict, or the named missing fact, with or
without a severity grade and a supporting quote. Hidden cost is what the row asks for beyond
its ledgered price. $^{\ddagger}$ marks designs that also generate
one cached artefact per consultation whose cost is not separately ledgered, making those
figures lower bounds; the two rows that read a threshold against a clean-note calibration
need a labelled clean set per deployment to place that threshold, which no price here
includes. A dash means the row asks for nothing beyond its price.}
\label{tab:cost}
\end{table*}

Over the same 495 evaluation pairs, the comparison the held-out subset could not resolve becomes decidable. Collapsed to consultations as Section 4's convention requires, the evolved prompt fires on 25 consultations where the per-fact rule does not, against 7 the other way (both fire on 52 and neither on 28), and that contrast gives p=0.002, so at its own operating point the evolved prompt detects significantly more notes with omissions than the per-fact rule. (Held out the same contrast read 10 to 3, p=0.09: the same direction, too few consultations to resolve it.)

That test reads the evolved prompt as a majority of three runs against the rule's one run, and removing that estimator mismatch strengthens the result. A majority fires only when two of three runs fire, so it is the more conservative estimator, and read one run at a time the evolved prompt's three draws give p=2.2e-05, 3.1e-04 and 1.2e-04 on the same consultations. What the equalised reading costs is the operating point (single draws detect 38.6 to 41.0\% at 6.2 to 9.8\% false alarms, against the majority's 36.9\% at 6.2\%), so the deployed-rule comparison above is the one a deployment would live at, and it is the weaker form of the contrast.

Neither operating point dominates. The evolved prompt buys its detection at 6.2\% false alarms against the rule's 2.7\%, and it cannot be tuned down to meet the rule, because most of its detection sits on notes scored exactly 9, so the next quieter threshold collapses to 14.3\% detection at 0.9\% false alarms, below the rule. The tuning fails in the other direction too. Widening the rule's trigger to supporting-severity facts lifts detection to 59.0\% but fires on 35.7\% of clean notes, because legitimate notes routinely leave merely-supporting facts unstated, so each method has exactly one usable operating point, and they are different ones. The flags are different objects too: a score below 10 with nothing attached, against the missing fact by name with its severity grade and the transcript quote behind it - the difference between ``re-read this note against its transcript'' and ``check whether the note records the clindamycin allergy''. The dearer method, the pipeline, buys precision and flag content rather than volume, and it also stays ahead on the capability measure on both sets of pairs. The two catch substantially different notes: together they fire on 84 of the 112 consultations against 77 for the evolved prompt alone, which makes running both the obvious next experiment.

The results form a ladder: performance in Table~\ref{tab:practitioner}, prices and what each flag contains in Table~\ref{tab:cost}, at the pinned credit rates with measurement dates in Appendix D. Detection at a usable false-alarm rate peaks at the evolved prompt, \$0.046 a note, a tenth of the pipeline's price at one note per consultation and a third once the pipeline's per-consultation stages are amortised across several notes. The money above it buys a quieter flag that names the missing fact: the full three-stage pipeline with the rule is \$0.45 a note, of which the severity audit is 71\% and the extraction 19\%. Two untested changes target those two lines: folding severity grading into the extraction call, which would remove the audit's share, and a cheaper extraction model (Section 8). The engineered judge's own free-text critical-omission list is the cheap end of the named-fact family (Table~\ref{tab:practitioner}'s last row). At half a cent it stays quiet, 9.6\% detection at 0.9\% false alarms against the rule's 24.6\% at 2.7\%, and hands over free text where the rule hands a keyed verdict with an audited severity grade.

On a stronger judge family the ordering compresses from below: the second family's monolithic completeness-scored judge clears its own noise in one call, and the rule's edge there is +5.6 points of detection (22.1\% against that judge's 16.5\%, both at a matched 2.1\% false-alarm rate; Appendix G) plus what its flag contains. What does not change across families is the paired gap, the two methods' complementary strengths, and the fact that nothing here comes close to solving the class.

Those prices and thresholds all come from constructed pairs, and one run measures what survives contact with real scribe output. The best of the eight designs and both methods, each at its own published flag rule, were pointed at 261 notes from the companion census: all 87 carrying a panel-verified omission, and 174 of the notes its panel cleared (Appendix K). Every operating point moves. Detection roughly doubles on both methods (the pipeline 51.7\%, the evolved prompt 85.1\%) and false alarms rise 8.3-fold and 7.1-fold (to 22.4\% and 44.2\%), because a real note is less complete than a constructed pair everywhere rather than at one planted site. The pipeline checks the same number of audited facts per note on both corpora and judges twice as many of them absent on the real notes, 2.0 per note against 1.1.

No threshold set on constructed clean twins survives the move, and the evolved prompt's flag-below-10 calibration fails exactly as Section 7.6 predicts. Once thresholds are re-established on the real notes, the evolved prompt still detects more than the monolithic judge at half its false-alarm rate, 32.2\% at 5.2\% against the monolithic judge's 29.9\% at 9.8\% at its own rule; the monolithic judge was not re-thresholded, but no threshold of its own reaches that detection inside a 5\% false-alarm band (Appendix K). The per-fact rule was not re-thresholded either, so on real notes the pipeline has no demonstrated operating point inside the 10\% bar this paper uses, and re-setting its audit stage's critical threshold there is the experiment that would supply one. And the pipeline's flags keep their content there: of the 45 notes it flagged among the 87 carrying a verified omission, 34 (75.6\% {[}61.3, 85.8{]}) name the very fact the census's panel verified, a figure no injected benchmark can supply, because agreement there is true by construction. That match is scored by lexical containment, so read it as an indicative screen rather than an adjudicated precision. The full design, rates and caveats are Appendix K.

\subsection{The cost on commissions, and the open problem}

\begin{figure}[!htbp]
\centering
\includegraphics[width=0.95\linewidth]{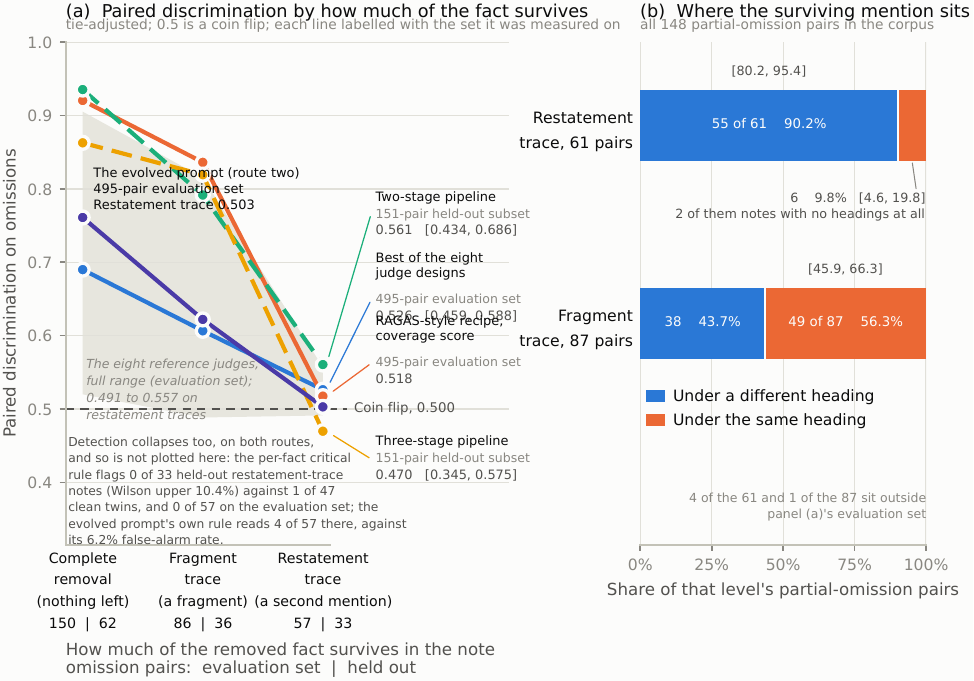}
\caption{\textbf{Every method we measured in the primary judge family collapses on the
omission whose restatement survives elsewhere in the note, and the supply of those cases
says why.} (a) Paired discrimination
on omissions - how often a note carrying an omission scores below its own verified-clean
twin, tie-adjusted, where 0.5 is a coin flip - split by how much of the removed fact
survives: nothing (complete removal), a fragment of it (fragment trace), or an
explicit or closely paraphrased second mention (restatement trace). Each line names the set it was measured on, and
the two sets are read separately rather than ranked against each other: the best of the
eight judge designs, the range across the eight reference judges, the RAGAS-style
recipe (our implementation, read as its coverage score rather than at its own
flag-everything rule), and the evolved prompt (route two in the released data, the
optimiser's winner at reasoning effort high, three replicates averaged) are measured on
the 495-pair evaluation set, whose 293
omission pairs split 150 / 86 / 57 across the three levels; the two-stage and three-stage
pipelines are measured on the 151-pair held-out subset, whose 131 omission pairs split
62 / 36 / 33. The band spans all eight reference judges, the coverage recipe among them
at the aggregate score it ships, so that recipe appears twice: once inside the band and
once as its own line at the coverage component alone. The coverage recipe is also the one
line with denominators of its own,
148 / 84 / 57, because parse failures cost it 14 of its 1,214 records. The interval
printed beside each restatement-trace value is a 95\% bootstrap interval over
consultations, computed over 35 consultations for the evaluation-set line that carries one
and 20 for the held-out ones, and every one of them contains 0.5; the coverage recipe's and
the evolved prompt's restatement-trace values have no such interval in the released
artefacts and are quoted as point estimates.
Detection collapses with discrimination, and is annotated rather than plotted because it
is a rate at a rule rather than a discrimination score: the per-fact rule flags
0 of 33 held-out restatement-trace notes against 1 of 47 clean twins, 0 of 57 on the full
evaluation set against 3 of 112, and the evolved prompt's own rule reads 4 of 57 there
against its 6.2\% false-alarm rate. (b) Where the surviving mention sits, over all 148
partial-omission pairs in the corpus (61 restatement-trace, of which 57 are in the
evaluation set, and 87 fragment-trace, of which 86 are): in 55 of 61 restatement-trace
pairs,
90.2\% [80.2, 95.4], the survivor sits under a different heading from the instance that
was removed, while 49 of 87 fragment-trace pairs, 56.3\% [45.9, 66.3], keep it under the
same
heading; two of the six same-heading restatement traces are notes with no headings at all.
Reading: once a restatement trace survives, every method plotted here sits at or below its
own noise floor. The one exception the study holds is not plotted - the second judge
family's checker discriminates the class at 0.636 paired while the rule still catches none
(Section~7.8). The supply explains why - a restatement trace is almost always
a different-heading survivor, recoverable only by reading the note end to end.}
\label{fig:trace}
\end{figure}

The recovery this section reports has two costs. The first is a cost on commissions, and it belongs to the pipeline. On the evaluation set the enumerating designs sit at the bottom of the study on paired commissions - the checklist judge at 0.744, the RAGAS-style recipe at 0.712, and the three-stage pipeline's coverage score at 0.661 (0.617 on the held-out subset) - below every one of the eight designs while occupying the top of the omission ordering, and the inversion is at least as wide on the second judge family. That is mechanically what a coverage score should do, since a fabrication does not reduce coverage. One case bounds the mechanism: the engineered judge is a single monolithic call rather than an enumerating design and still scores 0.754 on paired commissions, in the same band, so part of what looks like a cost of enumeration may be attention - its prompt and all eight of its worked examples are about omissions. The evolved prompt does not pay this price, holding paired commissions at 0.978 across 202 commission pairs while detecting omissions, but its own cost is the per-deployment calibration. Production evaluation built on a coverage-scored pipeline therefore needs a faithfulness judge running alongside.

The second cost is shared: even the best method catches barely more than a third of notes containing an omission at a usable false-alarm rate, an improvement rather than a solution, and the notes they miss are concentrated in one class. That case defeats every deployable method in this study. Restatement-trace omissions - the primary statement gone while an explicit or closely paraphrased restatement survives elsewhere - are 57 of the 293 omissions here, a minority class we built enough of to measure, and the kind in which the least information is actually lost; pooling it into the headline numbers makes them conservative. Within the class every method we measured in the primary judge family sits at or below its own noise floor - the eight designs, the reference judges, both pipeline tiers, and the evolved prompt that is the study's best detector everywhere else. Figure~\ref{fig:trace} plots each with the set it was measured on and a 95\% interval where the released artefacts carry one (two of the five lines do not), and the per-fact rule flags 0 of 57. The second-family checker splits the problem in two: it discriminates restatement traces well above chance (0.636 paired) while the rule still catches none, so signal exists at the fact level and the decision layer cannot act on it.

The obvious clinical objection is that a note still saying the thing somewhere is not deficient, so a judge that stays quiet is right rather than blind. Splitting the same stored verdicts by where the survivor sits answers it: in 55 of 61 restatement pairs corpus-wide (57 of the 61 sit in the evaluation set; 90.2\% {[}80.2, 95.4{]}) the surviving restatement sits under a different heading from the removed instance, and of the six same-heading pairs two are notes with no headings at all. The contrast with fragment traces is the mechanism: a fragment belongs to the same sentence or its neighbour and stays put (56.3\% same-section), while a fact stated twice in full is almost always stated once in the history and once in the reasoning. So what survives is recoverable only by reading the note end to end - the misplaced-text failure the census counts at 7.4\% of its verified findings. At this point the problem stops looking like omission detection and starts looking like structural consistency: the fact survives in the note, but not where the note claims to record it.

Two qualifications belong beside that. In 37 of 61 pairs the survivor sits somewhere in the assessment-and-plan block, so on the typical pair a reader who works through the whole note can still find the fact. And the split shows no detection gradient, which at these counts it could not have: every contrast is non-significant, and the same-section group holds five evaluated pairs (five of its six sit in the evaluation set), on which zero detections are compatible with rates up to about 43\%. The natural attack is structural rather than a better judge: score coverage per section instead of per note, which turns the cross-section survivor from an undetectable absence into a detectable misplacement, and the released site map supplies the section labels. The redundancy that made absence hard to construct in Section 3 is what makes the last omissions undetectable here, measured four ways: 32.0\% of the facts a note records stated more than once, 44\% of complete removals failing verification, a quarter of the census's 40 sampled omission refusals (candidate findings its panel declined to verify) turning on a restatement elsewhere (a share that moves with the sample - a 15-refusal pilot of the same audit read it at 0\%), and 90.2\% of surviving restatements sitting under another heading. We state it as the field's open problem rather than ours alone, and release the graded benchmark as the instrument for attacking it.

\section{Limitations}

\textbf{No external anchor for the omission half.} The principal unresolved validity question is whether the transcript-derived omission labels correspond to independent clinician judgements of documentation failure. The commission arm has an external anchor in MEDEC; the omission arm has none, because the only external physician-labelled benchmark whose task matches ours is drawn entirely from a corpus that is one of our own strata. Until such an anchor exists, the omission results are internal to the instrument this paper builds: constructed, verified and clinician-checked, but not independently labelled. Establishing that correspondence is the first thing future work has to do.

\textbf{One judge model family for the headline results.} Every headline judgement in Sections 5 to 7 comes from one judge model family. The second-family checks reproduce the asymmetry, the restatement-trace collapse (with Section 7.8's nuance: the transferred checker discriminates the class while the rule still catches none) and the pipeline's lead (Appendix G); the evolved prompt's calibration transfers too (Section 7.6). Single-note detection proves judge-dependent, so single-note claims are scoped to the family measured. One confound is disclosed: reasoning effort is not matched tier-for-tier across the two families in the Section 5 checks, because the second family's endpoint refuses a no-reasoning setting; the evolved prompt's transfer does not carry that confound, since both families run at high effort.

\textbf{What the compute settings bound.} Sections 5 and 6 are measured with the judge given no reasoning effort inside a 1,024-token cap, and all three prompt searches executed their candidates the same way, so the search explored a compute regime a deployment need not live in. Section 7 measures what a reasoning budget does to those prompts, but only to prompts already selected; a search re-run at a larger budget is a different experiment and we did not run it. One control is missing on the same subset: at matched compute the plain completeness-scored prompt matches the evolved prompt on paired discrimination (0.670 against 0.669), so the evolved prompt's measured advantage is the clean-note calibration and the operating point it admits, and whether a plain prompt given the same budget would score clean notes at 10 and so admit a flag rule of its own was not tested. Those re-runs add one limit of their own: the evolved prompt's clean-note calibration is a behaviour of this corpus and judge, not a property of the design.

\textbf{What the head-to-head can and cannot say.} The Section 7 comparison is at each method's own deployment threshold, not at matched false-alarm rates. Its estimator asymmetry is measured in Section 7.7, and removing it strengthens the contrast. The evaluation-set extension was run without an analysis plan fixed in advance: the tests reported are the ones established on the held-out sample, applied unchanged, with no new test chosen after seeing the data. The pipeline still has one replicate on the evaluation set (only the held-out subset carries three), so the equalisation reads the evolved prompt down rather than replicating the rule up.

\textbf{One clinician on this paper's own items, and the difference between endorsement and blinding.} A physician author validated 94 of this paper's own items across two sittings - 70 in a sitting that covered six stages of the benchmark and pipeline, and 24 more on the audit stage's own severity grades. Stages where the item itself had to state the verdict being audited are \textbf{endorsements} - a clinician checked and did not object; only the panel-cut, severity and judge-disagreement stages were structurally blinded, and only those are measurements. They are an author who designed the instruments they graded, and there was one rater across both sittings with no repeated item, so no within-rater figure exists. Every stage holds 6 to 20 items, so four stages reading 100\% with lower bounds of 61--72\% rule out gross failure and nothing finer: the intervals are the result, the point estimates are not. Severity-conditioned results use the benchmark's grades; the per-fact rule fires on the audit stage's own. One thing has now been checked from outside: the severity rubric. An independent clinician graded against it, blind, in the companion census (Section 3). That is external evidence the rubric reproduces to within a grade, and it is also why the one-grade lean reported in Section 7 is now read as a property of the rater rather than of the rubric - the two clinicians lean opposite ways. It is not a second reading of this paper's own items: nobody outside the author group has adjudicated a judge output here, and that gap is the one this limitation names.

\textbf{Injected, not naturally occurring, errors.} The companion census shows the classes are real, but injected pairs are cleaner than production failures, and we measured the gap on both arms. On commissions, the same judge scores paired 0.827 on MEDEC's physician-curated errors against 0.939 on our constructed ones. On omissions, Section 7.7's run against real vendor notes shows every operating point moving - detection and false alarms both rise steeply at the benchmark's own rules - so the benchmark's operating points do not transfer to production traffic, though once thresholds are re-established there the evolved prompt still detects more than the monolithic judge at half its false-alarm rate (Appendix K). The false-alarm reads there are upper bounds, not estimates: the census's cleared notes are clean only to the strictness of its discovery, and its own standards experiment moves the share of notes carrying something reportable from 27.8\% to 96.5\%. All commission pairs also come from the earlier build, so the construction-cohort robustness slice exists only for omissions, and a build effect acting on both classes at once cannot be excluded by slicing. Each pair also carries exactly one planted error where a real note carries several (the pipeline judges 2.0 facts absent per real note against 1.1 on the benchmark's pairs), so nothing here measures how a judge behaves when failures interact. One further bound runs through the constructed side: the clean twins are verified complete against the audited fact sheet rather than against the transcript, so every clean-note false-alarm rate this paper reports for an enumerating judge is an upper bound on wrongful flags of a complete note, and at the same time an optimistic reading of what a production note would draw, since no production note is complete against any sheet (Section 7.7).

\textbf{What the held-out subset is, and is not, held out from.} Ten of the confirmation's 47 consultations also appear in the exploratory pass with extraction caches reused, so it is a fully held-out test of the judging and only a partial one of the extraction. The pipeline tiers' paired scores and the per-fact rule carry three runs each, with spreads a quarter to an eighth of the monolithic baseline's (Section 7.2); the swept single-note figures and the per-trace splits remain single runs. The subset is held out with respect to optimisation, not to the number of analyses run over it - the budgeted re-runs, fact-list ablation and family transfer all read the same 151 pairs, with no multiplicity correction - and the per-fact rule itself was formulated post hoc, then checked on the earlier exploratory pass; the evolved prompt's flag-below-10 threshold was likewise read off the first replicate's scores and then held unchanged.

\textbf{Human detection, from the literature rather than our own measurement.} We did not run a human baseline. The published comparators run from about 25\% to 92\%, depending on the task and on whether the reader had the source document alongside. Our best configurations detect 24.6 to 36.9\% (at 2.7\% and 6.2\% false alarms respectively), the same range as the one comparator in which physicians reviewed without the source document - they caught about a quarter to a third of planted omissions - while the higher human figures come from easier tasks. None of this settles the question. Appendix I collects every figure we could find, and none of them covers the restatement case.

\textbf{Corpus limits.} English-language UK and US primary and ambulatory care only, and no note carries an EHR structured field, so we cannot speak to structured-field omissions at all. The two cost levers named in Section 7 are identified but not tested, and latency was not measured (Appendix C), so Section 7's price ladder is a cost ladder only.

\section{Conclusion and release}

Judges verify presence, not absence. No remedy we tried creates usable detection in place: wording, criterion scope, voting and prompt search all move the operating point along the same weak curve, and a larger reasoning budget lifts discrimination without closing the asymmetry. What works is converting the absence question into presence checks, and most of the recovery comes from the closed per-fact verdicts and the decision rule over them rather than from the fact list alone. This study's two strongest methods implement the conversion independently: a pipeline that enumerates the transcript's facts and returns a verdict on each, whose every flag names the missing fact, and a single GEPA-evolved call that detects more notes at roughly a tenth of the measured cost per note at one note per consultation (a third amortised). Acting on per-fact verdicts rather than a score is what makes the pipeline usable in deployment; a clean-note calibration, re-established per deployment, is what makes the evolved prompt so. An omission whose restatement survives elsewhere in the note still defeats both methods, and that is an open problem for the field. We release the instrument for attacking it. The four questions this record should be read against, and our own answers to them, are Appendix J.

\textbf{What follows directly.} The cheapest next experiments are visible in the stored verdicts: running both methods together, and re-setting the audit stage's critical threshold from clinician grades before re-scoring the verdicts already recorded. The direct sequel to Section 6 is a prompt search run at a reasoning budget, since the strategy the optimiser kept proposing could not execute inside the budget its candidates were scored under. The restatement case points at per-section coverage scoring, for which the released site map supplies the labels. Beyond those, the omission half still needs the external anchor Section 8 opens with.

\textbf{Author contributions.} S.F. designed the study, built the instruments, ran the experiments and analyses, performed the physician validation of Appendix E, and wrote the paper. L.M. reviewed the manuscript; R.L. and M.K. supported the work at Composo.

\textbf{Funding.} The study was carried out at Composo and received no external funding.

\textbf{Acknowledgements.} We thank Dr Hannah Warren-Miell for grading the severity rubric blind on the companion census's findings (Section 3); Dr Warren-Miell is not an author and had no other involvement in the study.

\textbf{Competing interests.} The authors build evaluation tooling commercially. Every judge evaluated here, including the deployed faithfulness judge from the authors' production evaluation, is fully specified in the release: prompts, model pins and judgements. The pipeline this paper recommends is the authors' own design, and the reference set includes a judge from the authors' production evaluation; both are released in full.

\textbf{Ethics and data.} The corpus contains no real patient data: PriMock57's consultations are between clinicians and actor patients (CC BY 4.0), ACI-Bench's encounters are simulated doctor-patient conversations released for research (CC BY 4.0), and the remaining strata are authored scenarios; no patients were involved and no clinical records were used. The three commercial scribe products whose notes appear in the companion census and in Section 7.7's real-error run are anonymised throughout as Scribe A, B and C. Nothing here is a clinical-safety evaluation of any named product, and no result should be used to select or reject a scribe for clinical use.

\textbf{What we release.} One bundle, shared with the companion census paper and citing the same dataset DOI: the 500 graded pairs with their clean notes, errored twins, removed facts and grades; the fact-site map and both graders' severity verdicts; the transcripts and fact sheets; every verified census finding, products anonymised; every benchmark judge run's judgements with the raw completions (the real-vendor-note run's records are judgements of withheld note text and are withheld with it); the clinician-validation records, as keys, answers and scored results (the item packs that print vendor note text are withheld with it; the two that carry none - the severity items with the rubric verbatim, and the construction-adjudication queue - ship); and the full harness. Appendix F is the release index.

\textbf{What we withhold.} The text of the notes the three scribe products wrote is not released. The products' terms of service differ on republication, and we withhold all three alike rather than release asymmetrically, which would make products identifiable by their absence (Appendix F). Released findings carry the transcript-side evidence and never the note's own words; a replicator with vendor accounts can regenerate equivalent notes with the released harness.

\section*{Data and code availability}

The benchmark and all study data are released as \textbf{OmissionBench} at \url{https://huggingface.co/datasets/ComposoAI/OmissionBench} under CC BY 4.0, with a DataCite DOI minted at publication against the released revision. The harness - capture scripts, every prompt including superseded versions, judge configurations and model pins - is at \url{https://github.com/composo-ai/omission-bench} under the MIT licence, archived to Zenodo at its tagged releases under DOI \texttt{10.5281/zenodo.22160954}. The released data comprise the 500 graded pairs (495 in the evaluation set), each with its clean note, errored twin, injected or removed fact and its severity and trace grades; the fact-site map and the per-fact severity grades with both graders' verdicts; the authored scenario transcripts, and the two public strata's transcripts in derived form under their own upstream CC BY 4.0 licences with attribution; the per-consultation fact sheets; every verified finding of the census with products anonymised, and the transcript-side evidence quoted; every benchmark judge run's judgements with its run manifest and the raw completions alongside (the real-vendor-note run's records excepted, since they judge withheld note text); the four clinician sittings' scored results, per-item keys and answers as returned; and note-level structural metadata. Vendor note text is withheld for licence reasons (Appendix~F): released findings quote the transcript side and never the note's own words, and a replicator with their own product accounts can regenerate equivalent notes with the released harness. The synthesised consultation audio is not redistributed and is regenerable from the released transcripts with the harness's text-to-speech recipe and its voice and model pins.

\bibliography{references}

\clearpage
\appendix

\section*{A guide to the appendices}

The eleven appendices fall into four groups, and most readers will want only one of them.

\textbf{Two that answer a question the paper raises about itself.} Appendix~J asks the four
questions that separate a scribe evaluation that means something from one that does not, and
answers them against this study's own record. Appendix~F is the release index: what ships, under
which licence, and where the artefact behind any given number lives.

\textbf{How the instruments were built and run.} Appendix~A is the ground truth in full - the
construction protocol, the extraction instrument's trajectory, and the severity rubric printed
verbatim - and is where to go with any question about how absence was made certain.
Appendix~C carries the pipeline's diagnostics and the parameters a deployment would set;
Appendix~D the model pins, settings, spend and reproducibility record.

\textbf{The objects the argument is about, printed rather than described.} Appendix~B holds the
three prompt-optimisation campaigns behind Section 6's null, with two prompts
verbatim: the confirmatory campaign's winner, which is careful, clinically sensible and bought
nothing held out, and the exploratory campaign's winner, which is the prompt
Section 7 deploys as the evolved prompt. Appendix~H prints the deployed faithfulness judge
whose wording Section 6 edits.

\textbf{The evidence at full resolution.} Appendix~E is the clinician validation, both sittings,
stage by stage, each with what its sample size does and does not rule out. Appendix~G is the
second judge family. Appendix~K is the run against real vendor notes. Appendix~I collects the
comparator literature and every published human-detection figure we could find.

\section{Ground truth: construction protocol, instrument trajectory, and the severity rubric}

This appendix is the construction record behind Section 3. It gives the fact-sheet schema and the critic panel that audits it, the full trajectory of the extraction instrument on real recordings, the blind-recovery check and the sheet-size asymmetry it exposed, the severity rubric printed in full, the reference-note repair loop, the pair-construction rules and their verification outcomes, the two cohorts of pairs carried over from an earlier build, and the earlier frozen pair set as a snapshot. A replicator should be able to rebuild the benchmark from this appendix, the released prompts and the released artefacts. Model pins and seeds for every stage are in Appendix D, with the transport and hash records in the released harness; Figure~\ref{fig:construction} is the same pipeline as a schematic.

\subsection{The fact sheet and the three critics}

\subsubsection{The schema}

For each consultation we extract a fact sheet from the transcript alone, in one call, with no note of any kind in the prompt. The sheet has exactly three keys.

{
{\small\begin{longtable}[]{@{}
  >{\raggedright\arraybackslash}p{(\linewidth - 4\tabcolsep) * \real{0.3333}}
  >{\raggedright\arraybackslash}p{(\linewidth - 4\tabcolsep) * \real{0.3333}}
  >{\raggedright\arraybackslash}p{(\linewidth - 4\tabcolsep) * \real{0.3333}}@{}}
\toprule\noalign{}
\begin{minipage}[b]{\linewidth}\raggedright
key
\end{minipage} & \begin{minipage}[b]{\linewidth}\raggedright
one item is
\end{minipage} & \begin{minipage}[b]{\linewidth}\raggedright
fields
\end{minipage} \\
\midrule\noalign{}
\endhead
\bottomrule\noalign{}
\endlastfoot
\texttt{must\_\hspace{0pt}contain} & one atomic fact a correct and complete note of this consultation has to include & \texttt{fact} (precisely stated, no compounds), \texttt{evidence} (a verbatim transcript quote of at most 25 words that supports it), \texttt{load\_\hspace{0pt}bearing} (\texttt{high} or \texttt{medium}, how much a competent clinician's reading of the note depends on it) \\
\texttt{must\_\hspace{0pt}not\_\hspace{0pt}contain} & one specific wrong claim a scribe could plausibly make on this consultation & \texttt{assertion} (the erroneous claim), \texttt{why\_\hspace{0pt}wrong} (one line on what the transcript actually establishes) \\
\texttt{salience\_\hspace{0pt}traps} & one moment in this transcript where a scribe could plausibly go wrong & \texttt{trap} (what is tempting to get wrong), \texttt{correct\_\hspace{0pt}handling} (what a correct note does), \texttt{mode} (one of omission, negation, dose value, laterality, attribution, modality hardening, temporal, fabrication, decision status, anchoring), \texttt{importance} (the severity grade of A.4) \\
\end{longtable}\addtocounter{table}{-1}}
}

The extraction prompt directs coverage at the clinically decisive content of a consultation: presenting complaint, key positives and pertinent negatives, examination findings or the explicit absence of an examination on a remote consultation, the working impression with any hedging preserved, the plan with drug names, doses, frequencies and durations as actually said, safety-netting, and material social or risk history. Three rules do the work that makes the sheet usable as an answer key: one fact per item with no compounds; nothing enters \texttt{must\_\hspace{0pt}contain} without a quotable supporting span; and where the transcript is ambiguous the sheet preserves the ambiguity rather than resolving it. The \texttt{evidence} and \texttt{load\_\hspace{0pt}bearing} fields exist so that a support audit can be run mechanically rather than by re-reading the transcript for every item. Both prompt versions are released in the code repository as \texttt{prompts/\hspace{0pt}extraction\_\hspace{0pt}prompt\_\hspace{0pt}v1.txt} and \texttt{prompts/\hspace{0pt}extraction\_\hspace{0pt}prompt\_\hspace{0pt}v2.txt}.

\subsubsection{The three critics}

Every extracted sheet is audited by three separate critic calls before it is used for anything. The transcript is immutable, so the panel asks only whether the sheet is a faithful reading of it.

\begin{itemize}
\tightlist
\item
  \textbf{Support} checks internal consistency between transcript and sheet: every \texttt{must\_\hspace{0pt}contain} item genuinely said, every trap a real moment in the dialogue, every \texttt{must\_\hspace{0pt}not\_\hspace{0pt}contain} item really an error for this consultation, no self-contradiction.
\item
  \textbf{Materiality} judges clinical importance rather than support: it flags trivia in the required facts (padding raises the completeness bar and penalises good notes), clinically important transcript content missing from \texttt{must\_\hspace{0pt}contain} (a working diagnosis, a red-flag safety net, a drug dose, a stated allergy, a decisive negative), and \texttt{load\_\hspace{0pt}bearing:\ high} grades clearly wrong in either direction.
\item
  \textbf{Leakage} guards the one contamination channel a blind extraction has: any item resting on information not in the transcript - outside clinical knowledge asserted as fact, content that could only come from a reference note or from knowing how scribes typically fail, meta language, and paraphrase drift where the \texttt{evidence} quote does not appear in the transcript or does not support the \texttt{fact}.
\end{itemize}

Critic findings carry a severity of \texttt{minor} or \texttt{material}. Any material issue triggers \textbf{one revision cycle}: the sheet is rewritten to fix only the flagged issues, with the transcript required back character-for-character identical, and the revised sheet is re-audited by the support critic. There is no second revision. A sheet whose material issues survive the revision is dropped, not patched, with the one exception described in A.2.

\subsubsection{What the panel kept}

{
{\small\begin{longtable}[]{@{}
  >{\raggedright\arraybackslash}p{(\linewidth - 6\tabcolsep) * \real{0.2500}}
  >{\raggedright\arraybackslash}p{(\linewidth - 6\tabcolsep) * \real{0.2500}}
  >{\raggedright\arraybackslash}p{(\linewidth - 6\tabcolsep) * \real{0.2500}}
  >{\raggedright\arraybackslash}p{(\linewidth - 6\tabcolsep) * \real{0.2500}}@{}}
\toprule\noalign{}
\begin{minipage}[b]{\linewidth}\raggedright
stratum
\end{minipage} & \begin{minipage}[b]{\linewidth}\raggedright
consultations
\end{minipage} & \begin{minipage}[b]{\linewidth}\raggedright
sheets kept
\end{minipage} & \begin{minipage}[b]{\linewidth}\raggedright
dropped
\end{minipage} \\
\midrule\noalign{}
\endhead
\bottomrule\noalign{}
\endlastfoot
PriMock57 (real UK primary-care recordings, clinicians with actor patients) & 57 & 53 & 4 (7.0\%) \\
ACI-Bench (US ambulatory, subsampled) & 48 & 45 & 3 (6\%) \\
Trap-blind authored (a stratum written with no knowledge of the trap schema, as a representativeness guard) & 10 & 9 & 1 (10\%) \\
Authored consultations, re-extracted blind & 30 & 27 & 3 (10\%) \\
\end{longtable}\addtocounter{table}{-1}}
}

That is \textbf{134 audited sheets} (53 + 45 + 9 + 27 blind re-extractions), alongside the 30 authored consultations' own sheets, written when those consultations were written, which are what A.3's blind-recovery check scores against. Counted as consultations rather than as sheets, 137 distinct ones carry a sheet. The 134 consultations that carry a verified clean note (A.6, F.3) are a different set of the same size as these 134 sheets. The ACI-Bench sample of 48 is drawn from 207 encounters with 67 excluded because they had been used to train a prior optimised judge in this codebase; those 67 turn out to be exactly ACI-Bench's official training split, so the sample is drawn purely from data held out from that judge (\texttt{provenance/\hspace{0pt}subsample\_\hspace{0pt}manifest.json} in the code repository).

\subsection{The extraction instrument on real recordings: 33.3\% to 22.8\% to 19.3\% to 7.0\%}

The first version of the extraction prompt worked on authored transcripts and did not work on real ones. On PriMock57 it produced sheets carrying material issues that survived the revision cycle in \textbf{19 of 57 consultations (33.3\%)}, against 6\% on ACI-Bench and 10\% on the two authored strata. The critic reports root-caused it, and the causes are systematic rather than random: PriMock's transcripts are real recordings carrying speech-recognition corruption and conversational filler, and the first prompt mishandled that in four specific ways.

The four causes, each verified in the critic reports with worked cases, were: resolving garbled audio into confident fact; bundling a correct rendering with a genuine error inside one forbidden item; padding the required-fact list with conversational detail no clinician would expect in a note; and writing traps in scribe-meta framing that encodes knowledge of how scribes typically fail rather than deriving the trap from the transcript. None of this was a failure of the critic panel, which caught every case; the worked record ships in the release (\texttt{provenance/\hspace{0pt}primock\_\hspace{0pt}instrument\_\hspace{0pt}trajectory.json} in the code repository).

The four fixes went into a second version of the extraction prompt, derived from the first by addressing exactly those four causes and making no other edits: (1) an uncertainty rule, so that a garbled, ambiguous or contradicted turn can never become a confident fact, and required facts may rest only on clean evidence; (2) an equivalence protection, so that \texttt{must\_\hspace{0pt}not\_\hspace{0pt}contain} may forbid only renderings that are clinically wrong under any reading, with correct unit, dose and frequency conversions explicitly protected; (3) a note-content rule, excluding conversational closure, reassurance boilerplate and patient-education phrasing from required facts unless clinically decisive; and (4) transcript-only traps, derived from what the transcript shows and never from meta-knowledge of how scribes fail. All 57 consultations were re-run on the new prompt rather than only the 19 that had dropped, because a stratum has to be one instrument rather than a mixture of two.

Two further corrections were needed, and neither is a change to the prompt. \textbf{The input-view correction}: the parsed PriMock57 release carries a \texttt{presenting\_\hspace{0pt}complaint} header that is the actor's case-card prompt rather than a fact about the consultation that happened, and it was reaching the extraction and critic views, killing sheets unfixably. It was struck from every input view and the full 57 re-run on the same prompt text. \textbf{The excision rule}: deleting a contested \texttt{must\_\hspace{0pt}not\_\hspace{0pt}contain} item can never make a sheet wrong, only less strict, so a final-pass rule deletes such items with a per-item log where they are a sheet's only remaining material issues - 9 items across 7 sheets were excised, and 3 sheets were refused excision because their residuals touched required facts or traps.

{
{\small\begin{longtable}[]{@{}
  >{\raggedright\arraybackslash}p{(\linewidth - 6\tabcolsep) * \real{0.2500}}
  >{\raggedright\arraybackslash}p{(\linewidth - 6\tabcolsep) * \real{0.2500}}
  >{\raggedright\arraybackslash}p{(\linewidth - 6\tabcolsep) * \real{0.2500}}
  >{\raggedright\arraybackslash}p{(\linewidth - 6\tabcolsep) * \real{0.2500}}@{}}
\toprule\noalign{}
\begin{minipage}[b]{\linewidth}\raggedright
instrument
\end{minipage} & \begin{minipage}[b]{\linewidth}\raggedright
dropped
\end{minipage} & \begin{minipage}[b]{\linewidth}\raggedright
drop rate
\end{minipage} & \begin{minipage}[b]{\linewidth}\raggedright
what changed
\end{minipage} \\
\midrule\noalign{}
\endhead
\bottomrule\noalign{}
\endlastfoot
first prompt & 19 of 57 & \textbf{33.3\%} & the original blind-extraction prompt \\
second prompt & 13 of 57 & 22.8\% & the four targeted fixes above (15.8\% excluding the newly diagnosed metadata class) \\
corrected input view & 11 of 57 & 19.3\% & the \texttt{presenting\_\hspace{0pt}complaint} header struck from every extraction, critic and revision view \\
+ excision rule & 4 of 57 & \textbf{7.0\%} & contested \texttt{must\_\hspace{0pt}not\_\hspace{0pt}contain} items deleted rather than whole sheets dropped \\
\end{longtable}\addtocounter{table}{-1}}
}

Kept-sheet composition moves with it: 38, then 44, then 46, then \textbf{53 of 57 kept}. All 5 consultations whose drops were caused wholly or partly by the header are resolved. The \textbf{4 remaining drops} are documented per consultation identifier in the released \texttt{provenance/\hspace{0pt}primock\_\hspace{0pt}instrument\_\hspace{0pt}trajectory.json}: three are genuine internal contradictions the revision cycle could not fix, each with a residual touching a required fact or a trap so that the excision rule could not apply, and one is a consultation where the revision call never returned a sheet at all.

Three things about this trajectory are worth a replicator's attention. It was fixed before the run that the final rate would be reported whatever it turned out to be and that the extraction prompt would not be iterated again after the second version, which is why the last two corrections are an input correction and a deletion rule rather than a third prompt. The other three strata stayed on the first prompt, because their drop rates were already at or near the final level and the fixes are specific to speech-recognition artefacts, so the corpus carries a per-stratum instrument difference and we disclose it here. And both prompt versions ship with the release, so the trajectory is checkable rather than asserted: real recorded consultations needed corrections that authored ones did not.

\subsection{Blind recovery of authored ground truth, and the sheet-size asymmetry}

The instrument that writes the answer key needs its own accuracy measured, and the 30 consultations we wrote ourselves make that possible: each already carries the fact sheet its authors produced alongside it, so re-extracting a sheet blind from the transcript alone and comparing it against that authored sheet is a direct test.

\textbf{Recovery is 646 of 650 authored required facts, 99.4\%} (95\% bootstrap interval {[}98.8\%, 99.9\%{]}, computed by resampling whole consultations so that each consultation counts as one observation), with a \textbf{per-consultation minimum of 93.8\%} - so the recovery is not an average hiding one bad consultation. The 650 facts are those of the 27 consultations whose blind re-extractions passed the critic panel (of the 30 re-extracted; the three drops are in A.1's table). The reading is that blind extraction is a faithful substitute for an authored sheet, that the extracted strata are not systematically thinner than the authored one, and that pooling the strata in the main analysis is licensed.

The size comparison went the other way, and we did not anticipate it.

{
{\small\begin{longtable}[]{@{}lll@{}}
\toprule\noalign{}
per consultation & authored sheet & blind-extracted sheet \\
\midrule\noalign{}
\endhead
\bottomrule\noalign{}
\endlastfoot
required facts & 24.1 & \textbf{42.4} \\
forbidden assertions & 16.1 & \textbf{24.0} \\
salience traps & 7.0 & \textbf{17.5} \\
\end{longtable}\addtocounter{table}{-1}}
}

Extracted sheets are 1.8 times fatter on required facts and 2.5 times fatter on traps. That is a size difference between strata, and it cuts against our own hypothesis rather than for it: a higher completeness bar on the extracted strata means a judge could appear to miss more omissions there for reasons that have nothing to do with whether judges can see absence. We report it as it stands.

\textbf{The consolidation pass} narrows it. Every extracted sheet's required facts were classified as \texttt{core} (a competent clinician would consider the note deficient without it) or \texttt{contextual}, with the classifier never shown a target count and no curving of any kind. Pairs and clean notes are built from the core view only; the full sheet is retained alongside it and nothing is deleted. Per-stratum mean core counts against the authored sheets' 24.1: \textbf{PriMock 24.8}, ACI-Bench 29.0, trap-blind 31.1, authored re-extraction 32.6. The PriMock stratum lands on the authored envelope; the others narrow substantially but stay fatter, and that is reported rather than forced. On traps, \textbf{2,154 of 2,296 traps are eligible for use as pair targets}, with 142 peripheral-importance traps excluded from injection.

\textbf{The reverse check found gaps in the human sheets}: asking which core facts the blind extraction found that the authored sheets had missed gives 13 gaps across 7 of the 30 authored consultations, mostly safety-netting and pertinent negatives. Blind extraction caught note-worthy content the human authors missed. This is reported as a result; the authored sheets and the 90 pairs built from them are left untouched rather than repaired, so nothing downstream shifts because of it.

\subsection{The severity rubric, printed in full}

\subsubsection{Severity without a rubric}

Grading an omission by how much it matters is only useful if the grade is stable. Natively it is not. Two frontier models - successive generations of one model family - grading the same 171 semantically matched traps independently agree at \textbf{Cohen's kappa 0.177}, 95\% interval {[}0.04, 0.31{]}, where 0 is the agreement expected by chance alone. Raw agreement is 66\%, which sounds tolerable until the chance baseline is taken out. The disagreement is one-sided rather than noisy: of 52 traps that one generation graded \texttt{supporting}, the other graded \textbf{37 as \texttt{critical}}. The kappa implementation was checked against perfect-agreement, systematic-disagreement, textbook and independence cases before it was used on real data.

\subsubsection{Severity with the rubric}

We wrote a clinical rubric with three grade definitions, worked anchors and a four-step decision procedure (printed below; it also ships verbatim, with its hash, in the released \texttt{validation/\hspace{0pt}sitting\_\hspace{0pt}severity\_\hspace{0pt}items.json}), and re-graded \textbf{all 683 traps (211 authored plus 472 extracted)} with two independent model graders from different families.

\begin{itemize}
\tightlist
\item
  \textbf{Cross-family agreement with the rubric - the two graders whose consensus the benchmark ships - is kappa 0.662}, 95\% interval {[}0.59, 0.73{]}, against \textbf{0.177 for the unanchored same-family pair}, with non-overlapping intervals.
\item
  The gain is not an artefact of the easier design: the original, harder cross-sheet matched-pair design still reads 0.50 to 0.56 under the rubric, with about 0.16 of the residual gap shown by control to be item-matching noise.
\item
  \textbf{The rubric moves one family's own grades substantially}: kappa 0.39 between that family's rubric-anchored grades and the earlier generation's unanchored run. The rubric is doing real work on how the model grades, not relabelling grades it would have produced anyway.
\item
  \textbf{Both rubric-anchored graders grade more severely than either model does natively.} On the 472 extracted-sheet traps graded by both runs, one family's rubric run gives 302 critical, 158 supporting and 12 peripheral, against 243, 190 and 39 for the same family's native grading - 64\% critical against 51\%. The action and safety tests fire often. We disclose this wherever a result is conditioned on severity: the paired finding is that severity is contested between frontier models until a written rubric anchors it, and that the rubric shifts the distribution towards critical.
\item
  \textbf{92 of 683 remaining disagreements (13.5\%) are resolved conservatively to the lower grade} and flagged in the data, because the second family is the more severe grader of the two. The consensus labels are the working severity axis.
\item
  \textbf{The rubric has since been graded blind by an independent clinician, and that is the only check from outside the author group anywhere in this paper.} Independent here means not an author of this study and with no involvement in it at any stage. It is not a claim about how the clinician was found, and this paper does not describe that. The companion census applies the same written rubric to its own verified findings, and a fresh sample of them was put to that clinician: \textbf{9 of 12 grades exact against the rubric and 3 one grade apart}. Read with the same census's earlier sitting by a physician author, \textbf{25 of 32 grades are exact and no disagreement anywhere exceeds a single grade}, so the rubric reproduces to within a grade under two clinicians who graded disjoint material. What the pair does not establish is a direction. The two clinicians lean opposite ways against the same rubric (three of twelve below it for the non-author clinician, four of twenty above it for the author), so no claim in this paper rests on the rubric leaning one way, and the one-grade gap Section 7 reports for the pipeline's audit-stage grades is read as a property of that rater as much as of the rubric. The check is of the rubric only. No adjudication of a judge output in this paper has been read by anyone outside the author group, and Section 8 states that gap.
\end{itemize}

\subsubsection{Severity as it enters the benchmark}

The same rubric, run verbatim, grades the facts the pairs are built from. The evaluation split carries \textbf{613 graded facts: 166 critical, 393 supporting, 54 peripheral}, from two cross-family graders on the verbatim rubric, with disagreement taking the lower grade and flagging the fact. \textbf{The two graders disagreed on 93 of 613 facts (15.2\%)}. A later pass graded the optimiser's development split as well, so the full severity artefact holds \textbf{880 facts (243 critical, 558 supporting, 79 peripheral, 137 flagged = 15.6\%)} at \textbf{\$44.19 over 1,760 calls}.

Peripheral facts are scarce in this corpus. Under a rubric whose third test is whether losing the fact materially weakens the note as a clinical record, almost everything a scribe records earns at least \texttt{supporting}. \textbf{54 peripheral facts out of 613}, and \textbf{36 of those 54 come from contextual required-fact items} - exactly the class the core view drops - which is why every earlier build of the benchmark contained no peripheral pairs at all.

Where a trap already carried an importance grade from its own fact sheet and was also graded fresh under the rubric, the two agree on 113 of 158 trap facts (71.5\%), with disagreements running both ways; the per-fact record is in the released severity artefact (\texttt{pairs/\hspace{0pt}factorial\_\hspace{0pt}severity.json} in the data repository).

\subsubsection{The rubric, verbatim}

The following is the rubric as the graders received it. The file's administrative header (internal file references and an author name) is omitted, and one internal workstream code in the final sentence is replaced in brackets; the graded text is otherwise unaltered, and the released file carries the same redactions.

\begin{lstlisting}
# Severity rubric - importance grading for salience traps and findings

## The question being graded

For a trap or finding, grade the clinical importance of the note getting this
RIGHT vs getting it wrong (omitting it, fabricating it, or altering it):
**if the note carried this error, how much would it matter for the patient or
the next clinician who reads the note?**

Grade the consequence of the error IN THE NOTE, not the drama of the topic.
A mundane-sounding fact can be critical (a drug allergy); a dramatic-sounding
one can be peripheral (a vividly described but self-resolved symptom with no
bearing on the plan).

## The three grades

**critical** - the error would plausibly change clinical action, delay or
misdirect care, or create a safety risk. Anchors:
- omitted working diagnosis or a hedged impression hardened into certainty
- omitted red-flag safety-netting ("return if the headache becomes sudden")
- omitted or wrong drug allergy, interaction, or contraindication
- wrong drug, dose, frequency, or duration; a conditional plan made definite
- fabricated examination on a remote consult; invented findings or results
- omitted pertinent negative that licenses the management decision (the denied
  cardiac features behind treating chest pain as musculoskeletal)
- wrong patient identity attributes (name, sex, age used clinically)

**supporting** - degrades the note's completeness, clarity, or defensibility,
but is unlikely to change what happens next. Anchors:
- omitted duration/onset detail that colours but does not gate the plan
- omitted social context relevant to follow-up (lives alone, occupation) where
  no immediate action hangs on it
- compressed phrasing that loses nuance without inverting meaning
- omitted secondary symptom that neither supports nor threatens the diagnosis

**peripheral** - no plausible clinical consequence. Anchors:
- conversational colour, rapport, patient-education phrasing
- administrative closure ("no further questions")
- redundant restatement of something already captured elsewhere in the note

## Decision procedure (apply in order)

1. **Action test**: would a competent GP, or the next clinician reading the
   note, plausibly DO something different if this error stood? Yes -> critical.
2. **Safety test**: even if action today is unchanged, does the error remove a
   safety net or create a latent risk (allergy, red flag, follow-up trigger)?
   Yes -> critical.
3. **Record-quality test**: does it materially weaken the note as a clinical
   record (completeness, defensibility, handover value)? Yes -> supporting.
4. Otherwise -> peripheral.

Tie-breaks: if genuinely torn between two grades after the procedure, take the
LOWER grade (conservative - protects the severity analysis from grade
inflation). Grade each item independently; do not curve to a distribution.

## What this rubric is for (and not)

It anchors model graders and the author-clinician to one written standard so
grades are comparable across models and time. It does not claim to settle
clinical importance in general - cross-grader disagreement UNDER this rubric
is itself a reported result [of this study].
\end{lstlisting}

The physician author's blinded validation of these grades is in Appendix E.

\subsection{The reference-note repair loop}

The clean note that every pair is built from is a repaired version of the corpus's own clinician reference note. The repair runs per consultation, after that consultation's fact sheet has passed the critic panel, and it sees three things: the transcript, the audited fact sheet, and the original clinician note.

It works in two steps. \textbf{The audit} lists every discrepancy between note and transcript, classified as \texttt{missing\_\hspace{0pt}fact}, \texttt{contains\_\hspace{0pt}error}, \texttt{unsupported} or \texttt{hardened\_\hspace{0pt}uncertainty} (the transcript hedged, the note is definite), each graded \texttt{material} - could change patient understanding or management - or \texttt{minor}. \textbf{The repair} fixes every material discrepancy with minimal edits, preserving the note's structure and style so the result reads as the same clinician's note corrected rather than a rewrite; both versions stay in the released record. A verification pass re-checks the repaired note, and notes that fail are fixed once more and re-checked.

\textbf{Every reference note in both public strata carries at least one material discrepancy with its own transcript}: PriMock \textbf{53 of 53 (100\%)}, 95\% interval {[}93.2\%, 100\%{]}, and ACI-Bench \textbf{45 of 45 (100\%)}, {[}92.1\%, 100\%{]}. The density is what matters more than the bare 100\%.

{
{\small\begin{longtable}[]{@{}lll@{}}
\toprule\noalign{}
& PriMock (53 notes) & ACI-Bench (45 notes) \\
\midrule\noalign{}
\endhead
\bottomrule\noalign{}
\endlastfoot
discrepancies, all severities & 1,030 & 703 \\
of which material & 566 & 320 \\
mean per note, all severities & 19.43 & 15.62 \\
\textbf{mean material per note} & \textbf{10.68} & \textbf{7.11} \\
\end{longtable}\addtocounter{table}{-1}}
}

By kind, with material counts in brackets: PriMock missing fact 574 (310), contains error 195 (139), unsupported 154 (60), hardened uncertainty 107 (57); ACI-Bench missing fact 356 (185), unsupported 193 (42), contains error 98 (57), hardened uncertainty 56 (36). Missing facts dominate on both. Graded on the rubric of A.4, PriMock's 1,030 discrepancies are 345 critical, 500 supporting and 185 peripheral, and ACI-Bench's 703 are 143 critical, 374 supporting and 186 peripheral. Every discrepancy ships with its verbatim evidence quote in the released audit record (\texttt{provenance/\hspace{0pt}wd\_\hspace{0pt}r3\_\hspace{0pt}report.json} in the code repository); Section 3 quotes one worked case.

\textbf{Convergence.} No note needed zero edits: the mean edit count is 18.74 per PriMock note and 15.69 per ACI-Bench note. Of the 53 PriMock notes, 19 passed verification first time, 34 needed a further fix and 23 of those then passed, leaving \textbf{11 still failing} the strictest residual checks. Of the 45 ACI-Bench notes, 18 passed first time, 27 needed a further fix and 21 then passed, leaving \textbf{6 still failing}. So \textbf{17 of the 98 repaired notes still fail verification on ultra-strict residuals}; all 17 are logged by identifier and none is silently kept.

Three framing points belong with these numbers. The 100\% is by our audit instrument's standard, a single-model auditor applying a strict rubric, so we report the discrepancy density with its verbatim evidence quotes and its severity split rather than resting on the saturated proportion. PriMock's reference notes are terse case summaries by design, which inflates the missing-fact counts specifically. And the instrument doing the auditing is the same extraction machinery whose blind recovery of human-authored facts is 99.4\% in A.3, so the checker has itself been checked.

\subsection{Pair construction, the single-edit invariant, and what verification rejected}

\subsubsection{Four classes and what a single edit means for each}

The corpus carries four pair classes. The older three-value \texttt{type} field (\texttt{add} / \texttt{change} / \texttt{omit}) is retained unchanged for compatibility, and \texttt{class} is the field that distinguishes the two kinds of omission.

{
{\small\begin{longtable}[]{@{}
  >{\raggedright\arraybackslash}p{(\linewidth - 2\tabcolsep) * \real{0.5000}}
  >{\raggedright\arraybackslash}p{(\linewidth - 2\tabcolsep) * \real{0.5000}}@{}}
\toprule\noalign{}
\begin{minipage}[b]{\linewidth}\raggedright
class
\end{minipage} & \begin{minipage}[b]{\linewidth}\raggedright
the edit
\end{minipage} \\
\midrule\noalign{}
\endhead
\bottomrule\noalign{}
\endlastfoot
\texttt{add} & one span asserting something the clean note does not support \\
\texttt{change} & one span's assertion altered \\
\texttt{omit-complete} & one fact removed from the note entirely, at every place it is stated \\
\texttt{omit-partial} & one fact's primary statement removed, its other statements left in place \\
\end{longtable}\addtocounter{table}{-1}}
}

The invariant that keeps a pair interpretable is that exactly one thing changed. What ``one thing'' means depends on the class, and getting that right was the hardest definitional problem in the build. For \texttt{add} and \texttt{change} the unit is a \textbf{span}: one non-equal block in the diff, with a blast-radius cap. For \texttt{omit-complete} the unit is a \textbf{fact}: removing a fact that lives in three sections is one edit by any standard a clinician would recognise and three edits by the diff, so the one-block rule is replaced by two conditions - every non-equal block is a deletion, or a repaired wound no longer than what it replaced, and every deleted span is a mapped statement of the target fact. For \texttt{omit-partial} the unit is a span again, the primary site, algorithmically identical to \texttt{add} and \texttt{change} and semantically its opposite, because here a residual is required rather than forbidden.

The \textbf{primary site} is the note's home statement of the fact: the site carrying it most completely and specifically, the one a reader would cite. It is nominated by the cross-family verifier, with recorded fallbacks to the mapper's nomination and then a deterministic order rule; which rule fired is recorded per fact. Every \texttt{omit-partial} pair carries residual metadata - the number of surviving mentions, each with its span, section and strength, plus the strongest surviving strength and the sections involved. Strength has three grades, strongest first: \texttt{explicit} (states the fact plainly in the note's own clinical terms), \texttt{paraphrase} (states the same fact in different words) and \texttt{partial} (states only part of it, losing a defining detail such as a value, a laterality, a hedge, or the half that drives action). There is deliberately no ``implied'' grade, because a span a reader could only infer the fact from is not a statement of it, and treating one as a site would delete unrelated clinical content.

\subsubsection{The one permitted revision of the edit constants}

The algorithmic single-edit constants took their one permitted revision, exercised before any semantic edit check ran, and the revision is disclosed here in full because it changed which pairs the build could produce. The first version of the constants rejected \textbf{210 of 411 pairs as built (51\%)}, including \textbf{20 of the 90 preserved authored pairs, 13 of them omissions} - the study's headline error class. The dominant false-failure classes were one-block contiguous section deletes of three to five units and adds or omissions realised as natural one-to-one line replacements, both of which are single-location edits under the rule's own stated intent. The second version raised the pure insert or delete cap from 2 to 6 units and the replace cap from 2 to 3 units a side, and dropped two proxies, word-operation purity and a maximum-regions count, because edit direction and atomicity are now verified semantically instead. Three things stayed frozen: the single non-equal block invariant that localises the edit, the 30-changed-word blast-radius cap, and the regeneration suffix's two-unit limit. The second version's fail count as built is \textbf{97}. No further revision is permitted, and none was taken.

\subsubsection{The site map}

Neither omission class can be built without knowing where in the note a fact is stated, so a mapping stage runs before any injection. Candidate facts per consultation are the pair-eligible omission-mode traps first, then core required facts with \texttt{load\_\hspace{0pt}bearing:\ high}, capped per consultation and seeded on the study seed. For each candidate the mapper locates every place in the clean note that states it - the smallest verbatim span, its section, its strength, its character offset - and the map is built by one model family and checked by another, asking whether each listed span really states the fact, whether any site was missed, and whether the primary nomination is right.

Over \textbf{134 consultations (112 evaluation plus 22 in the optimiser's development pool)} at 14 candidate facts each: \textbf{1,876 facts mapped, 1,791 present in the note, 85 absent, 2,617 sites located}, at \textbf{\$155.63 over 537 calls}.

\textbf{574 of the 1,791 present facts (32.0\%) are stated in more than one place}, at a mean of 1.461 sites per present fact. Redundancy reaches the facts that matter clinically:

{
{\small\begin{longtable}[]{@{}ll@{}}
\toprule\noalign{}
candidate bucket & facts stated more than once \\
\midrule\noalign{}
\endhead
\bottomrule\noalign{}
\endlastfoot
supporting traps & 71 of 116 = 61.2\% \\
critical traps & 80 of 155 = 51.6\% \\
peripheral traps & 8 of 23 = 34.8\% \\
core required facts, high & 276 of 813 = 33.9\% \\
core required facts, medium & 31 of 135 = 23.0\% \\
contextual required facts & 108 of 549 = 19.7\% \\
\end{longtable}\addtocounter{table}{-1}}
}

The second call is not optional at this scale: the verifier \textbf{amended 419 of 1,791 facts (23\%)} - dropping a span that did not state the fact, adding one the mapper missed, or moving the primary nomination - and 14 verification failures were excluded from construction. Of the 2,617 sites, 1,224 align to whole units and can be deleted with no model call at all. The full verdict counts, the per-stratum redundancy split and the sites-per-fact distribution are in the released site map (\texttt{pairs/\hspace{0pt}fact\_\hspace{0pt}sites.json}).

\subsubsection{Allocation and build}

The allocator ran to a target of 42 pairs for each of the nine trace-by-severity combinations over the 112 evaluation consultations, capped at 4 target facts and 6 pairs per consultation, giving \textbf{277 pairs planned from 177 target facts} (ACI-Bench 125, PriMock 83, authored 50, trap-blind 19). The build cost \textbf{\$7.29 over 173 calls}, or \textbf{\$0.026 a pair}, and the edit ledger totals: \textbf{104 pairs were pure programmatic span deletions with no model call at all, 65 needed a grammar tidy, 77 needed a collateral cleanup} where deleting a span orphaned a neighbouring clause (340 words dropped in total, every edit logged), and on \textbf{31 the attempted cleanup was rejected by the validator} so the raw deletion stood. Those four numbers sum to the 277 planned pairs. A grammar tidy is punctuation-level repair and nothing more: in one logged case, deleting the clause recording that vomiting came first left ``onset Sunday evening.; nausea settling.'', and the tidy made it ``onset Sunday evening; nausea settling.'' - no word of remaining content changes. Construction left \textbf{exactly one residual site on all 151 partial pairs}, by design.

\subsubsection{Verification standards, per class}

\texttt{add} and \texttt{change} pairs go through the unchanged edit check. The omission classes have their own standards.

{
{\small\begin{longtable}[]{@{}
  >{\raggedright\arraybackslash}p{(\linewidth - 2\tabcolsep) * \real{0.5000}}
  >{\raggedright\arraybackslash}p{(\linewidth - 2\tabcolsep) * \real{0.5000}}@{}}
\toprule\noalign{}
\begin{minipage}[b]{\linewidth}\raggedright
class
\end{minipage} & \begin{minipage}[b]{\linewidth}\raggedright
semantic checks
\end{minipage} \\
\midrule\noalign{}
\endhead
\bottomrule\noalign{}
\endlastfoot
\texttt{omit-complete} & \texttt{truly\_\hspace{0pt}absent} (the removed fact was present in the clean note and survives nowhere in the errored note, with no partial or paraphrased residual mention - verbatim the same clause the earlier build used, so the two are directly comparable); \texttt{spans\_\hspace{0pt}all\_\hspace{0pt}state\_\hspace{0pt}fact} (every removed span states the target fact, the other half of ``one edit means one fact''); \texttt{no\_\hspace{0pt}collateral\_\hspace{0pt}loss} (every other fact the clean note recorded is still recorded); \texttt{reads\_\hspace{0pt}naturally} (the result reads as a note that never recorded the fact, not as a note visibly cut about) \\
\texttt{omit-partial} & \texttt{primary\_\hspace{0pt}site\_\hspace{0pt}removed} (the home statement is gone); \texttt{residual\_\hspace{0pt}present} (at least one mention survives, quoted and graded - the inverse of \texttt{truly\_\hspace{0pt}absent}, and where the residual metadata comes from); \texttt{no\_\hspace{0pt}collateral\_\hspace{0pt}loss}; \texttt{reads\_\hspace{0pt}naturally} \\
\end{longtable}\addtocounter{table}{-1}}
}

Free algorithmic checks gate all of it first (deletion-only diff, no new content, removed and residual spans as required, every mapped site of every other fact still present). One instrument detail matters for anyone reproducing this: the omission checks are given the errored note and \textbf{no transcript}, because every question is note-internal, and supplying the transcript made the auditor fail \texttt{truly\_\hspace{0pt}absent} on the grounds that the fact still appears in the transcript, which it always does.

Verification is a cross-family panel of two models, with a third call from the auditor at a different seed on any field the two split on and a per-field majority deciding: \textbf{679 calls, \$33.92}.

{
{\small\begin{longtable}[]{@{}
  >{\raggedright\arraybackslash}p{(\linewidth - 8\tabcolsep) * \real{0.2000}}
  >{\raggedright\arraybackslash}p{(\linewidth - 8\tabcolsep) * \real{0.2000}}
  >{\raggedright\arraybackslash}p{(\linewidth - 8\tabcolsep) * \real{0.2000}}
  >{\raggedright\arraybackslash}p{(\linewidth - 8\tabcolsep) * \real{0.2000}}
  >{\raggedright\arraybackslash}p{(\linewidth - 8\tabcolsep) * \real{0.2000}}@{}}
\toprule\noalign{}
\begin{minipage}[b]{\linewidth}\raggedright
class
\end{minipage} & \begin{minipage}[b]{\linewidth}\raggedright
verified
\end{minipage} & \begin{minipage}[b]{\linewidth}\raggedright
passed the algorithmic checks
\end{minipage} & \begin{minipage}[b]{\linewidth}\raggedright
panel unanimous
\end{minipage} & \begin{minipage}[b]{\linewidth}\raggedright
verified and unanimous
\end{minipage} \\
\midrule\noalign{}
\endhead
\bottomrule\noalign{}
\endlastfoot
\texttt{omit-complete} & \textbf{71 of 126 (56.3\%)} & 121 of 126 & 90 of 126 & 60 \\
\texttt{omit-partial} & \textbf{114 of 151 (75.5\%)} & 145 of 151 & 130 of 151 & 101 \\
\end{longtable}\addtocounter{table}{-1}}
}

\subsubsection{The complete class's failure breakdown}

\textbf{55 of 126 complete-removal attempts (44\%) failed verification}, and the breakdown says why. Across those 55 the semantic checks failed as \texttt{truly\_\hspace{0pt}absent} 29, \texttt{reads\_\hspace{0pt}naturally} 28, \texttt{no\_\hspace{0pt}collateral\_\hspace{0pt}loss} 18 and \texttt{spans\_\hspace{0pt}all\_\hspace{0pt}state\_\hspace{0pt}fact} 15 - totalling 90 across 55 pairs because a pair can fail more than one check. The partial class's 37 rejections are dominated by \texttt{reads\_\hspace{0pt}naturally} (29). Taking all 92 rejected pairs together, the commonest single cause is \texttt{reads\_\hspace{0pt}naturally} alone, then \texttt{truly\_\hspace{0pt}absent}; every rejected pair is itemised with its failing checks in the released dataset, and rejected pairs are \textbf{excluded rather than patched}.

\textbf{The compound-fact caveat.} Part of the complete class's rejection rate is a property of the targets rather than of clinical notes. A trap's text plus its correct-handling line can name several things at once (``omitting the past surgical \emph{and} medical history''), and when the target is a compound like that, ``every site of \emph{the} fact'' is ill-defined and the mapper spreads across more of the note than it should. This is the single biggest identified driver of the 44\%: it is what \texttt{spans\_\hspace{0pt}all\_\hspace{0pt}state\_\hspace{0pt}fact} (15 failures) and \texttt{no\_\hspace{0pt}collateral\_\hspace{0pt}loss} (18 failures) are mostly catching. The cheap fix for a future build is to restrict complete-omission targets to required facts, which are single statements by construction, and leave traps to the partial class. We did not do it: it is a scope decision taken after the build, and the affected pairs are excluded rather than reworked. What Section 3 reads off the rejection rate does not rest on those cases, because the two commonest failing checks are the other two - \texttt{truly\_\hspace{0pt}absent} at 29 failures, meaning the fact was still there, and \texttt{reads\_\hspace{0pt}naturally} at 28, meaning the note was visibly cut about - and neither is a consequence of an ill-defined target.

\subsubsection{Two things the panel's second family bought}

\textbf{Family disagreement is concentrated and informative.} The panel split on at least one field in 36 of 126 complete pairs and 21 of 151 partial pairs, most often on \texttt{truly\_\hspace{0pt}absent} (22 pairs), where one family read a surviving fragment as a residual and the other did not. Three seeds of a single model would have re-sampled one model's blind spots instead of surfacing them. One asymmetry is disclosed: two of the three calls come from the same family, so a lone dissent from the other family forces a re-check rather than failing a pair by itself, which is why both the majority rate and the stricter unanimous rate appear in the table above.

\textbf{Construction's residual count is a property of the map, not of the note.} Construction left exactly one mapped residual site on every partial pair, but the panel's own count of surviving mentions found more than one in \textbf{71 of the 151}, and its strength grading differs from construction's on the margins. A sceptic reading the errored note finds mentions the mapper did not map. Every row records both counts and both gradings, so an analysis can control for it; the distributions are per-pair fields in the released set.

\subsection{The two carried-over cohorts, recorded per pair}

\begin{table}[!htbp]
\centering
\small
\setlength{\tabcolsep}{4pt}
\begin{tabular}{@{}p{0.78\linewidth}r@{}}
\toprule
\multicolumn{2}{@{}l}{\textbf{The released set}}\\
\addlinespace[1pt]
\hspace*{1em}Pairs released\newline\hspace*{1em}{\footnotesize a pair is one edited note and the verified-clean twin of the consultation it came from} & 500\\
\hspace*{1em}Pairs in the evaluation set\newline\hspace*{1em}{\footnotesize the remaining 5 were seen by the prompt optimiser and are excluded for every judge, so all judges score an identical item set} & 495\\
\hspace*{1em}Consultations\newline\hspace*{1em}{\footnotesize 112 of them in the evaluation set} & 117\\
\hspace*{1em}Verified-clean twins\newline\hspace*{1em}{\footnotesize one per evaluation consultation; the same note with the fact intact} & 112\\
\hspace*{1em}Matched couples: the same fact removed at two different trace levels\newline\hspace*{1em}{\footnotesize the build proposed 122 pairs naming a partner; in 45 couples both members survived verification (90 pairs)} & 45\\
\midrule
\multicolumn{2}{@{}l}{\textbf{What the evaluation set's 495 pairs test}}\\
\addlinespace[1pt]
\hspace*{1em}Omissions\newline\hspace*{1em}{\footnotesize complete 150, fragment trace 86, restatement trace 57} & 293\\
\hspace*{1em}Commissions\newline\hspace*{1em}{\footnotesize additions 95, alterations 107} & 202\\
\midrule
\multicolumn{2}{@{}l}{\textbf{Where the 500 pairs came from}}\\
\addlinespace[1pt]
\hspace*{1em}Add-or-change controls, carried over untouched & 202\\
\hspace*{1em}Built new against the fact-site map & 185\\
\hspace*{1em}Complete omissions from the earlier build\newline\hspace*{1em}{\footnotesize built by model rewrite rather than span deletion, and graded by the older single-arm severity pass} & 79\\
\hspace*{1em}Relabelled partial-omission seeds\newline\hspace*{1em}{\footnotesize surviving-mention counts run 1 to 9, because they predate the site map} & 34\\
\bottomrule
\end{tabular}
\caption{\textbf{Composition of the released set.} What the benchmark holds, what its evaluation pairs test, and where the pairs came from. All 202 add-or-change controls are pairs carried over from an earlier build, while most omission pairs are newly built against the fact-site map, so a comparison between the two error classes is also a comparison across construction cohorts; A.7 records the four sources per pair.}
\label{tab:composition-a}
\end{table}

Table~\ref{tab:composition-a} sets out what the released set holds, what its evaluation pairs test, and where the 500 pairs came from. This section records the cohorts behind that last panel. The released set pools pairs from four sources, and two of the four were built by an earlier procedure. Rather than describe this in a footnote, the differences are fields on every pair, so any analysis can restrict to a uniform cohort. The relevant fields are \texttt{source}, \texttt{class}, \texttt{residual\_\hspace{0pt}level}, \texttt{severity}, \texttt{severity\_\hspace{0pt}source}, \texttt{residual.n\_\hspace{0pt}surviving} and a per-pair \texttt{provenance} block naming the builder and its inputs.

{
{\small\begin{longtable}[]{@{}
  >{\raggedright\arraybackslash}p{(\linewidth - 6\tabcolsep) * \real{0.2500}}
  >{\raggedright\arraybackslash}p{(\linewidth - 6\tabcolsep) * \real{0.2500}}
  >{\raggedright\arraybackslash}p{(\linewidth - 6\tabcolsep) * \real{0.2500}}
  >{\raggedright\arraybackslash}p{(\linewidth - 6\tabcolsep) * \real{0.2500}}@{}}
\toprule\noalign{}
\begin{minipage}[b]{\linewidth}\raggedright
source
\end{minipage} & \begin{minipage}[b]{\linewidth}\raggedright
pairs
\end{minipage} & \begin{minipage}[b]{\linewidth}\raggedright
how built
\end{minipage} & \begin{minipage}[b]{\linewidth}\raggedright
severity grades
\end{minipage} \\
\midrule\noalign{}
\endhead
\bottomrule\noalign{}
\endlastfoot
newly built graded pairs & 185 & programmatic span deletion off the site map, verified by the cross-family panel & two cross-family rubric graders with the conservative tie-break, all 185 \\
earlier-build complete omissions & 79 & model rewrite under the earlier injection prompts, not span deletion, and with no site map available & 66 inherited trap importance grades, 13 rubric grades from the earlier freeze \\
relabelled partial-omission seeds & 34 & earlier-build omission attempts that failed the whole-note absence check and are verified partial omissions (A.8) & 24 inherited trap importance grades, 10 rubric consensus grades \\
commission controls & 202 & earlier build, carried over untouched & inherited grades \\
\end{longtable}\addtocounter{table}{-1}}
}

Two consequences follow, and both are visible per pair rather than argued.

\textbf{The 79 earlier-build complete omissions carry older severity grades}: built by model rewrite rather than span deletion off the map, graded mostly on the older single-grader path. They satisfy \texttt{omit-complete} by outcome rather than construction - each passed the same whole-note absence clause, so they are complete omissions of facts that lived in exactly one place.

\textbf{The 34 relabelled partials vary in how much trace survives}: they predate the site map, so the panel's count of surviving mentions runs from 1 to 9 against the newly built partials' uniform 1. Their residual strength is graded on the same three-level scale, so the trace level is comparable even though the count is not.

The newly built column is the internally uniform one and the pooled column is the bigger one, and both are reported wherever the distinction can matter. Section 3 states the one consequence that changes how a reader uses the headline result: all 202 commission controls are carried-over pairs while most omission pairs are newly built, so the headline asymmetry compares across construction cohorts, and the robustness slice restricted to the uniformly built omission pairs is what bounds it. Restricted to the 185 newly built omission pairs, the best of the eight judge designs reads 0.610 paired on omissions against the pooled 0.634, and the commission-minus-omission gap widens from 0.305 to 0.329. Every omission figure falls a little, because the carried-over cohort is slightly easier than the newly built one, and the per-fact rule of Section 7.5 reads 16.7\% on the 90 newly built held-out omission pairs against 20.6\% pooled, with the false-alarm cell unchanged because clean notes carry no cohort.

\subsection{The earlier frozen pair set, as a snapshot}

The graded benchmark of Section 3 replaced an earlier frozen pair set as the evaluation material. That earlier set is where the commission controls, the 79 complete omissions and the 34 partial seeds come from, and its audit produced several measurements the current build inherits. It is recorded here as a snapshot rather than as a live artefact.

\subsubsection{Composition}

{
{\small\begin{longtable}[]{@{}ll@{}}
\toprule\noalign{}
& pairs \\
\midrule\noalign{}
\endhead
\bottomrule\noalign{}
\endlastfoot
constructed & 411 \\
excluded by the answer-key audit & 78 \\
surviving corpus & \textbf{333} \\
frozen evaluation set & \textbf{281} \\
routed to the optimiser's development pool, audited identically & 52 \\
clean notes excluded, unrepaired after three repair rounds & 3 \\
\end{longtable}\addtocounter{table}{-1}}
}

The type mix over the 333 is add 111, change 129, omit 93. The 78 exclusions, by cause: 27 omission pairs that failed three regenerations \emph{and} the programmatic deletion fallback; 20 add and 5 change pairs that failed three regenerations; 17 authored-stratum pairs that failed the semantic majority on a decisive field, the authored stratum being preserved verbatim and never rewritten; and 9 pairs orphaned by the 3 excluded notes.

The frozen 281-pair file and an earlier published freeze of 280 pairs both ship in the release under their recorded hashes. Those hashes are not a way to tell the two builds apart: the freeze artefact embeds its own timestamp, so identical data hashes differently, and a third hash in the verification artefact is this same data re-frozen. Compare pair content, not hashes.

\subsubsection{Method measurements from the same run}

The run's own QA record - first-pass defect and failure rates by stratum, audit reproducibility across seeds, checklist coverage of verified omissions, and the auditor's false-flag decay at three-seed confirmation - ships with the release, under one standing caveat: no human adjudicated this run's flags, so its defect rates are auditor-majority upper bounds. The notes were repaired until they passed rather than the generator being rewritten, and that is disclosed rather than left implicit. The datasheet in Appendix F quotes the rates a user of the benchmark needs.

\subsubsection{The absence-clause correction, and how the 34 partial seeds arose}

A freeze-decision document argued that 36 excluded omission pairs had failed on bookkeeping rather than on substance: the edit check's absence clause conjoined genuine absence with ``corresponds to at least one required-fact item'', and trap-targeted omissions often have no such index. The clause was split so that the index is still recorded but gates nothing, and 40 pairs were re-verified at the same auditor, seeds and lever. The projection did not hold. \textbf{2 pairs were restored and 38 still failed.} In most of the 36 the index was in fact found, and the check still failed on the first clause: a partial or paraphrased residual of the removed fact survives in the errored note. The auditor's own words, on pairs the document had held up as demonstrably good: \emph{``still partially reflected by `No clinical signs of pulmonary oedema today' in the Impression''}; \emph{``a partial residual mention remains in the errored Impression: it still says findings point to a chest-wall origin''}.

The re-run was stable, which is what makes either verdict trustworthy: 31 of the 36 returned a character-for-character identical failure set, and of the 4 control pairs excluded on the single-edit clause alone, 3 got \emph{stricter} - removing the required-fact anchor made the auditor more willing to find residuals, not less. The changed prompt is hash-logged in the release beside its predecessor.

Two consequences carry forward. The frozen set tests only omissions of facts that lived in exactly one place. And single-point omission is often infeasible on a real clinical note, because the same fact is recorded in more than one section - which is the same conclusion the 44\% rejection rate of A.6 reaches from the other end. The 34 pairs are not broken; they are verified partial omissions, and they enter the released set as the relabelled partial-seed cohort of A.7.

\subsubsection{One property that constrained what the frozen set could measure}

The 281 frozen pairs carry \textbf{255 critical, 26 supporting and 0 peripheral} severity grades, from 218 inherited trap importance grades and 63 rubric grades, because the consolidation pass of A.3 deliberately excluded peripheral-importance traps from injection targets. A severity axis with two populated levels out of three is close to a two-group comparison rather than a rank correlation, and we said so rather than reporting a rank statistic as though it had range. The graded build of Section 3 exists partly to fix this: it carries \textbf{345 critical, 116 supporting and 39 peripheral} across its 500 pairs (the 293 evaluation omissions are Table 1's 151, 103 and 39; the 202 commission controls and the five excluded pairs carry the remaining 194 critical and 13 supporting).

\section{The three prompt-optimisation campaigns}

Section 6 reports that letting an optimiser rewrite a judge's prompt relocates the judge's operating point without creating the missing omission signal. This appendix carries the record behind that sentence at the depth the null needs: the three campaigns' designs and data, the held-out discipline of the confirmatory campaign, its deviations from the optimiser's published procedure, and - in B.5 - the confirmatory seed and winner printed verbatim. The candidate populations, acceptance tables and lineage records ship in the released repository (\texttt{gepa/\hspace{0pt}lineage.jsonl}, \texttt{gepa/\hspace{0pt}v2\_\hspace{0pt}results.json}, \texttt{gepa/\hspace{0pt}v3\_\hspace{0pt}lineage.jsonl}, \texttt{gepa/\hspace{0pt}v3\_\hspace{0pt}results.json}), each record carrying the candidate's full instruction text; the acceptance-against-held-out comparison for both confirmatory targets is Table~\ref{tab:gepa}, in B.4 below.

The optimiser is reflective prompt evolution, published as GEPA \citep{agrawal2026gepa}. It works in a loop: run the judge prompt over a sample of training examples, hand a second model natural-language traces of what the judge got wrong, let that model propose a rewritten prompt, score the rewrite, and keep it if it scores better. The prompt being rewritten belongs to the \emph{student}, which here is always the judge under test; the model that reads the failures and writes the rewrite is the \emph{reflector}. Both are pinned in Appendix D. The student ran at temperature 1.0, no reasoning effort and the 1,024-token answer cap throughout, matching the settings the judge designs of Section 4 used; the reflector ran at high reasoning effort in the first campaign and medium in the third, with a 16,000-token budget. That split carries interpretive weight, because the student's budget is the one setting under which every candidate of every campaign was scored, and one of the winners behaves differently outside it. B.7 records the re-run.

\subsection{What evolves, and what stays fixed}

Only one block of text changes. Everything around it - how the transcript and the note are laid into the prompt, and the format the answer must come back in - is fixed by the harness and identical for every candidate within a campaign, so a comparison between two candidates is a comparison between two instruction blocks and nothing else. The prompts of B.5 are those instruction blocks. Three wrappers are in play across the three campaigns; all three ship as release files named in B.5, so a reader can reconstruct any judge call exactly.

\subsection{The exploratory campaign}

The first campaign optimised a whole-note judge against a training pool built for it: \textbf{79 judgements over 22 consultations} - 16 additions, 22 alterations, 22 clean notes, 14 complete omissions and 5 partial omissions, 50 of them from PriMock57 and 29 from our authored scenarios. The pool shares no consultation with the 112 evaluation consultations; the overlap is asserted at load and recorded as zero in the run's own artefacts. No evaluation pair was judged at any point in this campaign.

It started from three seeds - the completeness-scored criterion of the eight designs (\texttt{seed\_\hspace{0pt}fc\_\hspace{0pt}score}), the engineered judge of Section 4 (\texttt{seed\_\hspace{0pt}engineered}), and the deployed faithfulness judge with an affirmative omission instruction added (\texttt{seed\_\hspace{0pt}v14\_\hspace{0pt}incl}) - ran \textbf{40 iterations}, and produced \textbf{21 candidates} (the 3 seeds plus 18 accepted mutations, with 22 further proposals rejected at a minibatch pre-filter), with zero unparseable judgements. The search objective mixed severity-weighted omission detection, commission detection and false alarms, with feasibility capped at 10\% false alarms on the pool's 22 clean notes.

Two facts from its population carry into the body, and both survive any inspection of the released lineage. The first is the trade the population makes: \textbf{every candidate that reached 30\% or better severity-weighted omission detection carried a false-alarm rate of at least 22.7\%}, which on 22 clean notes is five of them. The frontier sits between the winner (\texttt{gepa-04}, a mutation of the engineered judge's seed accepted at iteration 4) at 28.6\% detection with 9.1\% false alarms and the cheapest infeasible candidate at 31.4\% with 22.7\%. The second is the asymmetry, which is where Section 5 gets its earliest sighting: \textbf{all 21 candidates detect additions and alterations better than omissions} on the same severity-weighted measure, by margins running from 14.6 to 49.3 percentage points - the deployed judge, both deployable seeds, the winner, and every accepted and rejected mutation in between.

The winner's apparent gain was a shift of the score distribution, not better separation: against the seeds, severity-weighted omission detection rose from 11.4\% to 28.6\% at 2 of 22 clean notes flagged, while the gap between the mean clean-note score and the mean score on notes with omissions is 0.36 to 0.55 points on a ten-point scale for the winner and all three seeds alike, and on complete omissions the winner ranks the flawed note below its twin \emph{least} often of the four (4 of 14). On data the optimiser had never seen - the study's evaluation material, 586 omission and 224 clean-note judgements - the property that made it feasible collapsed: \textbf{23.0\% of omissions flagged (135/586) at a false-alarm rate of 21.9\% (49/224)}, a gap of 0.35 standard errors, with paired discrimination on omissions of \textbf{0.549} and an area under the detection curve of 0.535, where 0.5 is a coin flip. No significance testing was run on the 79-judgement pool and none should be read into its numbers: 14 complete omissions and 22 clean notes mean one example moves a rate by five to seven percentage points.

\subsection{The second campaign}

The second campaign aimed directly at the strongest of the eight designs - the completeness-scored eight-sample judge - on a rebuilt pool of \textbf{176 judgements over the same 22 consultations}, 97 of them graded omission pairs. It ran 11 iterations and accepted one mutation, which at the design's own eight samples reads 0.418 paired on omissions against the baseline seed's 0.685 - below a coin flip, and below its own parent - and nothing in the campaign separates from the baseline (the widest gap in its tables is 3.45 points at p=0.45, and it belongs to a seed rather than to anything the search found). No prompt file was written, because writing one would have recorded a discovery that did not happen. The pool's 22 clean twins bound what the campaign can say: every false-alarm rate moves in steps of one twenty-second, so the supported claim is that prompt optimisation of this design did not beat the design's own baseline prompt here, not that prompt optimisation cannot help. Both comparison tables, the trajectory and the stall-rule extension are in the released record (\texttt{gepa/\hspace{0pt}v2\_\hspace{0pt}results.json}).

\subsection{The confirmatory campaign}

\begin{sidewaystable}
\centering
\footnotesize
\setlength{\tabcolsep}{4pt}
\begin{tabular}{@{}p{0.26\linewidth}p{0.27\linewidth}p{0.20\linewidth}>{\raggedright\arraybackslash}p{0.17\linewidth}@{}}
\toprule
Prompt & paired omissions & best detection at $\leq$10\% false alarms & false alarms\\
\midrule
\multicolumn{4}{@{}p{0.97\linewidth}@{}}{\textbf{Leg A - the per-fact check module inside the two-stage pipeline} (15 iterations, 9 candidates, 8 accepted, 7 rejected; winner \texttt{v3p-01} at iteration 1)}\\
\multicolumn{4}{@{}p{0.97\linewidth}@{}}{\textit{Acceptance set: 20 consultations, 48 omission pairs, 20 clean twins}}\\
\quad seed: the check prompt as shipped & 0.792 & 33.3\% (16/48) & 10.0\% (2/20) \\
\quad winner \texttt{v3p-01} & 0.844 & 33.3\% (16/48) & 5.0\% (1/20) \\
\textit{\quad difference} & \textit{+5.2 pp [$-$4.2, +16.2], 8 wins to 3, p = 0.23} & \textit{0.0 pp, the same notes flagged} & \textit{$-$5.0 pp} \\
\multicolumn{4}{@{}p{0.97\linewidth}@{}}{\textit{Held-out set: 47 consultations, 131 omission pairs, 47 clean twins}}\\
\quad seed: the same shipped prompt, measured before the search & 0.801 (replicate one) & 16.0\% (21/131) & 8.5\% (4/47) \\
\quad winner \texttt{v3p-01} & 0.779 & 16.8\% (22/131) & 8.5\% (4/47) \\
\textit{\quad difference} & \textit{$-$2.3 pp [$-$7.4, +2.7], 8 wins to 13, p = 0.38} & \textit{+0.8 pp, 3 flags to 2, p = 1.0} & \textit{0.0 pp} \\
\addlinespace
\multicolumn{4}{@{}p{0.97\linewidth}@{}}{\textbf{Leg B - the monolithic faithfulness-and-completeness scoring judge} (15 iterations, 9 candidates, 7 accepted, 8 rejected; winner \texttt{v3m-08} at iteration 14)}\\
\multicolumn{4}{@{}p{0.97\linewidth}@{}}{\textit{Acceptance set: 20 consultations, 48 omission pairs, 20 clean twins}}\\
\quad first seed: the faithfulness-and-completeness scoring criterion, one sample & 0.615 & 2.1\% (1/48) & 5.0\% (1/20) \\
\quad second seed: the hand-engineered omission prompt, one sample & 0.646 & 18.8\% (9/48) & 5.0\% (1/20) \\
\quad winner \texttt{v3m-08}, one sample & 0.688 & 25.0\% (12/48) & 5.0\% (1/20) \\
\textit{\quad difference against the first seed} & \textit{+7.3 pp [$-$7.0, +25.0], 13 wins to 10, p = 0.68} & \textit{+22.9 pp, 11 flags to 0, p $<$ 0.001} & \textit{0.0 pp} \\
\multicolumn{4}{@{}p{0.97\linewidth}@{}}{\textit{Held-out set: 47 consultations, 131 omission pairs, 47 clean twins}}\\
\quad first seed at eight samples, run 1 of 3, measured before the search & 0.622 & 13.0\% (17/131) & 6.4\% (3/47) \\
\quad winner \texttt{v3m-08} at eight samples & 0.626 & 8.4\% (11/131) & 6.4\% (3/47) \\
\textit{\quad difference} & \textit{+0.4 pp [$-$12.6, +13.3], 40 wins to 36, p = 0.73} & \textit{$-$4.6 pp, 2 flags to 8, p = 0.11} & \textit{0.0 pp} \\
\textit{\quad yardstick: the same held-out baseline, its three runs} & \textit{0.622 / 0.649 / 0.531} & \textit{13.0\% / 10.7\% / 7.6\% (17, 14, 10 of 131)} & \textit{6.4\% / 8.5\% / 6.4\% (3, 4, 3 of 47)} \\
\bottomrule
\end{tabular}
\caption{\textbf{Two optimisation targets, what the acceptance set said about each and what the held-out set said.} Paired discrimination on omissions is the tie-adjusted share of pairs in which a judge scores the note with the omission below its own verified-clean twin, so 0.500 is a coin flip. Detection is read at the most sensitive threshold whose false-alarm rate stays within 10\%, and is always quoted beside that false-alarm rate. The acceptance set is 20 consultations, 48 omission pairs and 20 clean twins, and the optimiser accepted or rejected candidates on it; the held-out set is 47 consultations, 131 omission pairs and 47 clean twins, split at consultation level because two pairs from one consultation share a transcript, a clean twin and a fact list, committed before any optimisation and scored once, at the end. Differences are in points of paired discrimination, so +0.004 reads as +0.4. Intervals are bootstraps over consultations, sign tests count only the pairs the two prompts ordered differently, and the detection comparison is an exact McNemar test on the flag decisions. Leg B searched at one sample per note and confirmed at eight, so its two blocks are the same criterion asked a different number of times, and the comparison that carries the result is winner against seed within each block. Leg A's acceptance-set gain, the one the search was accepted on, is not distinguishable from noise on the acceptance set itself (8 wins to 3, p = 0.23). The last row is the yardstick: the same held-out baseline run three times, spanning 0.531 to 0.649 paired and 7.6\% to 13.0\% detection, so a single run of this design carries about $\pm$0.06 of paired noise - fifteen times leg B's held-out difference of +0.004. The held-out seed row's 0.801 paired is replicate 1 of the two-stage tier's three runs; Section 7.2 and Table~\ref{tab:practitioner} quote that tier as the mean of the three, 0.786.}
\label{tab:gepa}
\end{sidewaystable}

Both earlier campaigns drew the reflection traces and the acceptance decision from a single small pool, which is the overspecialisation the optimiser's own documentation warns about, and B.2 is what it cost. The third campaign fixes that at full scale, runs the published mechanics rather than an approximation of them, and points the search at a compound system - the setting the optimiser's authors report it is strongest on - as well as at a whole-note judge. It is the campaign Section 6 rests on. It ran on 2026-08-14 in a single session, with zero unparseable judgements and no killed runs.

\subsubsection{The partition is the experiment}

The split was built and committed before any optimisation, and disjointness was asserted at every load: against the held-out subset, and against the 154 pair ids and 22 consultations the earlier campaigns had trained on, at both pair and consultation level. Everything is split at \textbf{consultation} level, because two pairs from one consultation share a transcript, a clean twin and a fact list, so a pair-level split would leak all three.

{
{\small\begin{longtable}[]{@{}
  >{\raggedright\arraybackslash}p{(\linewidth - 12\tabcolsep) * \real{0.1429}}
  >{\raggedright\arraybackslash}p{(\linewidth - 12\tabcolsep) * \real{0.1429}}
  >{\raggedright\arraybackslash}p{(\linewidth - 12\tabcolsep) * \real{0.1429}}
  >{\raggedright\arraybackslash}p{(\linewidth - 12\tabcolsep) * \real{0.1429}}
  >{\raggedright\arraybackslash}p{(\linewidth - 12\tabcolsep) * \real{0.1429}}
  >{\raggedright\arraybackslash}p{(\linewidth - 12\tabcolsep) * \real{0.1429}}
  >{\raggedright\arraybackslash}p{(\linewidth - 12\tabcolsep) * \real{0.1429}}@{}}
\toprule\noalign{}
\begin{minipage}[b]{\linewidth}\raggedright
split
\end{minipage} & \begin{minipage}[b]{\linewidth}\raggedright
consultations
\end{minipage} & \begin{minipage}[b]{\linewidth}\raggedright
pairs
\end{minipage} & \begin{minipage}[b]{\linewidth}\raggedright
notes
\end{minipage} & \begin{minipage}[b]{\linewidth}\raggedright
complete / partial-weak / partial-strong omissions
\end{minipage} & \begin{minipage}[b]{\linewidth}\raggedright
additions / alterations
\end{minipage} & \begin{minipage}[b]{\linewidth}\raggedright
role
\end{minipage} \\
\midrule\noalign{}
\endhead
\bottomrule\noalign{}
\endlastfoot
held-out test & 47 & 151 & 198 & 62 / 36 / 33 & 10 / 10 & Section 7's confirmation subset. Touched once per target, at the end \\
reflect & 45 & 180 & 225 & 55 / 31 / 14 & 37 / 43 & minibatches and reflection traces; the reflector reads nothing else \\
accept & 20 & 85 & 105 & 27 / 13 / 8 & 17 / 20 & acceptance, per-example rewards, choice of winner; the reflector never sees it \\
\end{longtable}\addtocounter{table}{-1}}
}

The held-out baselines were already on disk before the campaign began, bought in the runs of Sections 5 and 7, so the final comparison cost one run of the candidate and nothing else. The analysis path here recomputes the two-stage pipeline's held-out figures from those stored records and reproduces them exactly. Those figures are replicate one throughout, which is the baseline run the campaign's winners were measured against: 0.801 paired on omissions (Section 7.2's published 0.786 is the mean of three primary-family runs spanning 0.771 to 0.801), 0.935 on complete removals, 0.792 on a fragment trace, 0.561 on a restatement, and 16.0\% detection (21/131) at 8.5\% false alarms (4/47).

\subsubsection{The mechanics that were run}

The published procedure was followed on the points that matter to whether a search can overfit its own validator: candidates are kept on a per-example Pareto frontier rather than by a single aggregate score, parents are sampled in proportion to the examples on which they lead, acceptance is promiscuous (a child enters on a minibatch improvement and the frontier prunes afterwards), merge is enabled once six candidates exist, feedback to the reflector is textual and per-failure, and minibatches are drawn consultation-first so every flawed note's own clean twin sits in the same batch. The acceptance objective's primary term is paired discrimination on omissions, with detection at a false-alarm rate of 10\% or below as secondary - paired is primary because the acceptance split holds 20 clean twins, so its false-alarm axis moves in five-point steps, too coarse to steer a search with.

\subsubsection{Target one: the per-fact checker inside the two-stage pipeline}

This is the one place a gain looked most plausible, because the pipeline of Section 7 is where the study's strongest omission signal sits and its check stage asks a small, well-posed question. The student was that pipeline with exactly one paragraph mutable. Extraction stayed frozen and cached, so the held-out run scored against the identical fact lists the baseline had been scored on, and the check stage ran at no reasoning effort through the pipeline's own code path. The search ran 15 iterations: 8 candidates accepted (7 mutations and 1 merge), 7 rejected at the minibatch, 3 merge invocations. The winner, \texttt{v3p-01}, appeared at \textbf{iteration 1}, and nothing later beat it.

Table~\ref{tab:gepa} carries the comparison. On the acceptance split the mutation gains 5.2 points of paired discrimination and halves the false-alarm rate at identical detection. On the 131 held-out omissions it is \textbf{2.3 points worse} (cluster bootstrap over the 47 consultations {[}-0.074, +0.027{]}; sign test on the 21 discordant notes, 8 its way against 13, p=0.38) and \textbf{0.8 points better on detection}, which is one extra omission out of 131 (on the flag decisions, 3 against 2, p=1.0). The interval's upper end is what bounds the result: the held-out data excludes gains larger than about 2.7 points of paired discrimination. Neither prompt's detection separates from its own false-alarm rate - 1.38 standard errors for the candidate and 1.27 for the baseline - which is Section 7's picture unchanged.

The detail worth carrying is that \textbf{the acceptance-split gain was not significant on the acceptance split either}: sign test 8 wins against 3, p=0.23, cluster bootstrap {[}-4.2, +16.2{]} points on 48 omissions and 20 clean twins. The held-out set did not overturn a real effect. It declined to confirm a difference that was never distinguishable from noise on the sample used to accept it.

No prompt file was written. The rule set before the run was that a prompt file is emitted only when the winner beats its baseline on the primary measure on held-out data, and it does not.

\subsubsection{Target two: the completeness-scored whole-note judge}

The second target was the same judge the second campaign aimed at, at one sample during the search with the held-out confirmation at eight. Two seeds: the completeness-scored criterion, which the harness rebuilds byte for byte from the design's own prompt file, and the hand-written seed, which is the engineered judge's content re-expressed in that design's frame. 15 iterations: 7 candidates accepted (4 mutations and 3 merges), 8 rejected at the minibatch, 3 merge invocations. The winner, \texttt{v3m-08}, appeared at \textbf{iteration 14} and is itself a merge of two descendants of the hand-written seed.

The nine-candidate population for this leg is likewise in the released lineage (\texttt{legs.mono.valid.all\_\hspace{0pt}candidates}); the winner, \texttt{v3m-08}, appeared at iteration 14 as a merge of two descendants of the hand-written seed.

A decision rule fixed before the run said to buy the held-out confirmation only if the winner beat the completeness-scored seed on the acceptance split by more than two points on both measures. It fired: \textbf{+7.3 points paired (0.688 against 0.615) and +22.9 points of detection (25.0\%, 12/48, against 2.1\%, 1/48)}, and it clears the hand-written seed too, by 4.2 points paired and 6.2 points of detection. So the confirmation was bought.

Table~\ref{tab:gepa} carries the held-out result beside the baseline's own three replicates on the same pairs. Against the matched single replicate the winner is \textbf{+0.004 paired} (sign test 40 against 36 over 76 informative pairs, p=0.73; cluster bootstrap {[}-0.126, +0.133{]}) and \textbf{4.6 points worse on detection} (on the flag decisions, 2 against 8, p=0.11). The yardstick is the baseline's own spread on these same 131 pairs: paired 0.531 to 0.649 and detection 7.6\% to 13.0\%, so a single run of this architecture carries roughly ±0.06 of paired noise here, about fifteen times the difference the search produced.

\subsubsection{What the search wrote}

Both targets produced longer, more specific, entirely sensible prompts. The first target's seed is 564 characters and its winner 3,183, with the most elaborate candidate in the population at 15,117; the winner names the clinically decisive part of a fact, polarity, certainty, dose and attribution, and tells the checker to preserve hedging. It is good advice. What it bought is visible in the held-out score distributions: mean score on clean notes 0.9587 against the baseline's 0.9612, mean score on notes with omissions 0.9315 against 0.9323, so the gap between a clean note and a flawed one is \textbf{0.0272 against 0.0289}. The prompt moved the whole distribution down about a quarter of a percentage point and left the separation where it found it - the same operating-point shift the levers of Section 6 produce, one level down, at the per-fact lookup rather than the note-level judgement.

One aside: the first target's winner has better commission calibration held out than the baseline (0.725 against 0.650, and 0.700 against 0.600 on the 10 addition pairs) - not a finding at n=20, but it points where Section 7's fabrication diagnostic pointed, at an enumerating judge's fabrication blindness being reachable by wording in a way its omission blindness is not.

\subsubsection{Departures from the optimiser's published procedure}

Four, all deliberate and all recorded in the run's artefacts.

\begin{enumerate}
\def\labelenumi{\arabic{enumi}.}
\tightlist
\item
  \textbf{Merge is implemented as a reflector-mediated crossover} of two frontier candidates that lead on different examples, rather than the published module swap. Both targets evolve exactly one module - the first freezes extraction on purpose - so a module swap has nothing to swap between.
\item
  \textbf{The second target's hand-written seed is not character-for-character identical} to the engineered judge of Section 4 that shares its content: it is that prompt's content re-expressed in the frame the eight designs use, with the transcript and note moved to the front. The completeness-scored seed \emph{is} identical, character for character, to that design's own prompt.
\item
  \textbf{The acceptance objective's primary term is paired discrimination rather than single-note detection}, for the sample-size reason stated with the mechanics above.
\item
  \textbf{The search ran at one sample per judgement on both targets}; only the second target's held-out confirmation ran at eight.
\end{enumerate}

The two earlier campaigns departed further, and in ways the third was built to fix: both drew reflection traces and acceptance decisions from one pool, the first accepted candidates on an 8-example unstratified minibatch rather than full-pool scores, and the second searched at four samples with one seed left unconfirmed at eight and extended its own stall rule mid-run. Every departure is recorded in the released campaign records.

\subsection{The confirmatory seed and winner, verbatim}

B.4 records the confirmatory campaign growing a 564-character seed into 3,183 characters of clinically sensible instruction that bought nothing on held-out data. That pair is reproduced verbatim below, exactly as stored in the campaign's lineage record, so the reader can see the object the claim is about. \textbf{Every other seed, winner and harness wrapper of all three campaigns ships verbatim in the released repository} - the campaign lineage files carry every candidate's full instruction text with its hash - and the inventory below identifies each one so a reader can pull and verify any of them byte for byte: the inventory - every seed, winner and wrapper of all three campaigns, with character counts and SHA-256 prefixes - is in the released lineage records named at the top of this appendix, and the three harness wrappers are the release files \texttt{gepa/\hspace{0pt}gepa\_\hspace{0pt}data.py}, \texttt{w2\_\hspace{0pt}prompts/\hspace{0pt}grid\_\hspace{0pt}FC\_\hspace{0pt}score.txt} and \texttt{w2\_\hspace{0pt}prompts/\hspace{0pt}pipeline\_\hspace{0pt}check\_\hspace{0pt}binary.txt}, so any judge call in any campaign can be reconstructed exactly.

\subsubsection{The confirmatory seed, verbatim}

The check-stage paragraph as shipped, 564 characters:

\begin{lstlisting}
For each reference fact, decide whether the clinical note CAPTURES it - stated verbatim, stated
in different words, or unambiguously implied by what the note does say. A faithful paraphrase, a
correct unit or dose conversion, an accepted abbreviation, and the same fact recorded in an
unexpected section of the note all count as captured. Anything the note simply does not carry, in
any form, is not captured.

Judge only whether each fact is there. Do not judge the note's style, length, ordering or
formatting, and do not penalise it for carrying extra material.
\end{lstlisting}

\subsubsection{The confirmatory winner, verbatim}

\texttt{v3p-01}, the mutation that gained 5.2 points on the acceptance split and was 2.3 points worse held out, 3,183 characters:

\begin{lstlisting}
For each numbered reference fact, judge whether the note conveys the same clinical proposition, not merely whether it contains related words.

First identify the fact's load-bearing content: the patient/finding/action involved; polarity or negation; timing/currentness; site/laterality; value, dose, route or frequency if relevant; who said or did it when attribution matters; certainty level such as possible/probable/confirmed/ruled out/not established; and any stated reason, red flag, pertinent negative, diagnosis, plan, urgency, or "only/no other" qualifier that changes the clinical meaning.

Mark a fact present when the note carries that load-bearing content explicitly, by faithful paraphrase, by accepted abbreviation, by a conventional clinical phrase, or by an unavoidable implication from a heading or context. It may be in any section and need only be recorded once. Do not require conversational wording, duplicate mentions, exact order, or unnecessary detail. A broad but clinically standard statement can cover a narrower fact when its scope clearly includes it, such as a relevant normal examination phrase, a system-wide negative, or a medication/allergy/status entry under the appropriate heading.

Mark a fact absent when the note does not let a reader recover the load-bearing proposition from the note alone. In particular:
- A mere mention of the same topic, symptom, medication, diagnosis, or plan is not enough if the required relationship is missing.
- A surviving fragment is not enough if the fact depends on the dropped element, such as the working diagnosis, explanatory link, reason for urgency, dose, current medication status, "no other" qualifier, red flag, or pertinent negative.
- If the note states an incompatible version of the fact, mark the reference fact absent: opposite polarity; present vs denied; assessed vs not assessed; different dose/value/timing/site; wrong source or attribution where source matters; or a symptom term used in a different clinical context.
- Preserve certainty. A possible, suspected, differential, working, or to-be-excluded diagnosis is not captured by a note that records it as confirmed; a confirmed diagnosis is not captured by a note that records only a query; "not established" or "not ruled out" is not captured by documenting the finding as absent.
- For medication, contraception, allergy, past history, social history, and occupation facts, require the status the fact asserts. Mentioning an item somewhere else does not by itself establish that it is current, the only medication, absent, denied, or part of the formal medication/allergy/history list unless the note's wording or section unambiguously does so.
- For patient ideas, concerns, expectations, clinician explanations, counselling, safety-netting, and plans, require the substantive content being attributed, not just that discussion occurred.

Be generous about faithful clinical compression, abbreviations, and section placement, but do not fill in missing qualifiers from medical common sense or from what would usually be true. The question is whether this note actually carries this fact, with its clinically important meaning intact.
\end{lstlisting}

\subsection{The portable lesson, and three corollaries}

Across two targets, 30 iterations and 15 accepted candidates, the confirmatory campaign moved the held-out estimate by -2.3 and +0.4 points of paired discrimination, neither distinguishable from zero, against a baseline whose own run-to-run spread on the same pairs is 0.531 to 0.649. The narrow claim is the one Section 6 makes: prompt optimisation, run properly, did not beat the prompts we already had, on either a whole-note judge or the per-fact checker inside the pipeline.

The methods result is more portable, and it is the reason this appendix prints the acceptance tables next to the held-out ones. \textbf{A 20-consultation acceptance split will hand you a five-to-seven-point improvement that a 47-consultation held-out set does not see.} Both the first and the third campaign found that; the first found it only after emitting a prompt file, and that prompt then went on to read 21.9\% false alarms on data it had never seen. Three corollaries follow, and they are cheap to apply.

\textbf{Split at the unit that shares structure, not at the row.} Two pairs from one consultation share a transcript, a clean twin and a fact list, so a pair-level split leaks all three and the validator stops being independent of the training signal.

\textbf{Measure the held-out baseline before the search starts.} The held-out figures for both baselines here were already bought and on disk before the campaign began, so the final comparison cost one run of the candidate and the held-out set was touched exactly once, at the end.

\textbf{Emit the artefact only on the held-out result.} The third campaign's first target was chosen on a 5.2-point acceptance-split gain that its own sample could not resolve - 8 wins against 3, p=0.23 - so a rule keyed to the acceptance split would have shipped it, and the held-out set says there was nothing to ship. The rule used instead - a prompt file is written only when the winner beats its baseline on the primary measure on held-out data - meant no file was written for either target. The only prompt file any of the three campaigns emitted is the first campaign's, written under the loose rule, and B.2 is what happened to it.

\subsection{The budget every candidate was scored in, and what happens when it is lifted}

One setting is common to all three campaigns and appears in none of their result tables. Every candidate prompt was executed by the student with no reasoning effort inside the 1,024-token answer cap of Section 4 - \texttt{gepa\_\hspace{0pt}optimize.py}, \texttt{gepa\_\hspace{0pt}v2.py} and \texttt{gepa\_\hspace{0pt}v3.py} each pass \texttt{reasoning\_\hspace{0pt}effort="none"} at their judging call - while only the reflector, which reads the failure traces and writes the rewrite, was given room to deliberate. A candidate could therefore be proposed with an instruction it had no budget to carry out, and then be scored as though the instruction had been tried.

That is what happened to the exploratory campaign's winner. \texttt{gepa-04}'s instruction block leads with an enumerate-then-check procedure, and B.2 records it reading 23.0\% of omissions flagged at 21.9\% false alarms on data it had never seen, paired 0.549. Re-run unchanged, character for character - the same released prompt file, \texttt{gepa/\hspace{0pt}judge\_\hspace{0pt}prompt.txt}, SHA-256 prefix \texttt{bb47c133} - on the 151-pair held-out subset at high reasoning effort with a 12,000-token cap, three seeds, it reads \textbf{0.669 paired against 0.546 for its own effort-none control on that same subset} (change +0.123, 95\% interval {[}+0.059, +0.189{]}, computed by resampling whole consultations so that each consultation counts as one observation). That budgeted configuration is the evolved prompt of Section 7.6. Its clean-note score distribution changes shape too, with 45, 46 and 45 of the 47 clean twins landing on exactly 10 out of 10, which is the calibration behind the operating point Section 7 reports at 32.1\% detection and 2.1\% false alarms. The prompt text is identical character for character across the two runs, and only the reasoning budget it executed under differs.

The same re-run carries the control that keeps this from being a claim about compute in general. The engineered judge of Section 4, the strongest single-prompt judge we could write by hand, gains \textbf{+0.013 {[}-0.047, +0.071{]}} when re-run the same way, an interval straddling zero on a prompt already at 0.635, while the plain completeness-scored criterion gains \textbf{+0.101 {[}+0.048, +0.153{]}}. A reasoning budget is worth about a tenth of a point of discrimination to some prompts and nothing to others. Table~\ref{tab:compute} carries the full ladder with the compute-matched comparison against the pipeline, and Section 7 the reading. Every prompt text involved is unchanged from B.5 and the released inventory, and the re-run's records, per-replicate tables and clustered intervals ship in the data repository under \texttt{judgements/\hspace{0pt}judges/\hspace{0pt}w2-power/\hspace{0pt}}, beside the study's other judge families.

Two things this does not establish. It is one already-selected prompt given a budget, not a search conducted at one: a search with a larger student budget would make different acceptance decisions over a different candidate population, and it was not run. And the clean-note calibration is a behaviour observed on three seeds on this corpus, not a property the prompt guarantees, so a deployment would have to re-establish the threshold on its own clean notes (Section 8).

\subsection{The exploratory winner, verbatim}

The prompt behind Section 7's single-call operating point - \texttt{gepa-04}, the exploratory campaign's winner, released as \texttt{gepa/\hspace{0pt}judge\_\hspace{0pt}prompt.txt} (SHA-256 prefix \texttt{bb47c133}, 9,071 characters) - is reproduced in full below, exactly as stored, since it is one of the two objects the paper's central result rests on. Its instruction block opens with the enumerate-then-check procedure B.7 discusses. No seed contained that procedure: the engineered judge's seed that B.2 names, this prompt's parent, asks the judge to weigh what the note leaves out but never to inventory the transcript first, and the reflector wrote the inventory step in the rewrite that produced this candidate, accepted at iteration 4 of the campaign's forty. The rest of the block is the seed's content reworked around it. The braces mark the two slots the harness fills per call.

\begin{lstlisting}
CONSULTATION TRANSCRIPT:
{transcript}

CLINICAL NOTE (AI-generated from the transcript above):
{note}

---

You are a clinical safety reviewer assessing whether a clinical note is faithful to, and complete from, the consultation transcript it was generated from.

Your central question is: would a competent clinician reading only the note know the clinically load-bearing facts established in the consultation, without being misled by anything the note added, inverted, hardened, or changed?

Do not judge by how polished the note is. Judge by clinical content.

Review method: compare in both directions

1. Transcript-to-note audit: first build a mental inventory of the clinically important facts established in the consultation, then check whether the note carries them.
2. Note-to-transcript audit: check every clinically meaningful assertion in the note against the transcript. A wrong assertion is as serious as a missing one.

Clinically important facts to look for include, when present in the transcript:

- Presenting problem(s), key chronology, duration, progression, severity, location, laterality, triggers, functional impact, and associated symptoms.
- Pertinent positives and pertinent negatives, especially red-flag questions and answers.
- Symptoms that broaden the differential even if they are not the main complaint, such as exertional breathlessness, chest pain, syncope, fever, weight loss, neurological symptoms, bleeding, severe pain, or suicidal/self-harm thoughts.
- Examination findings, observations, abnormal findings, and important normal findings used to justify reassurance or conservative management.
- Working diagnosis, differential diagnosis, level of uncertainty, and the reasoning behind tests, referrals, or conservative treatment.
- Medication names, doses, routes, frequency, duration, changes, stopped medication, allergies, adverse reactions, contraindications, pregnancy status, and other safety-relevant history.
- Investigations requested, results discussed, referrals made, urgency, target service, reason for referral, and any timeframes.
- Follow-up and safety-netting: what to do, when to do it, which symptoms should trigger action, and where to seek help.
- Safeguarding, capacity, mental health risk, social circumstances, occupation/driving advice, or administrative outcomes when these affect care.

Omission types you must catch

A fact can be omitted even when the general topic is still mentioned. Ask what a reader would actually know from the note.

Count it as an omission if the note preserves only a vague shell but loses the load-bearing part, for example:

- It says "bloods requested" but not the suspected condition or clinical reason that makes the results actionable.
- It says "referral made" but not that it was urgent, cancer-suspected, same-day, or driven by red-flag features.
- It mentions "safety-net given" but omits the specific trigger, route, or timeframe that matters.
- It lists a symptom area but drops the severity, duration, exertional nature, laterality, neurological feature, bleeding, fever, weight loss, or other detail that changes risk.
- It records a medication but omits the dose, frequency, duration, or change where that would affect safe prescribing.
- It records a diagnosis but drops the clinician's uncertainty or differential when that uncertainty was important to management.
- It mentions general aches or symptoms elsewhere but fails to carry a specific red-flag answer, such as pain/stiffness with neck movement, exertional symptoms, waking from sleep, vomiting, neurological symptoms, or self-harm denial.

Altered-content types you must catch

Penalise clinically meaningful changes even if nothing is "missing" in the ordinary sense. In particular, actively check for:

- Flipped negation or polarity: "denies vomiting" becoming "vomiting"; "not waking from sleep" becoming "waking from sleep"; "no suicidal ideation" becoming suicidal ideation; "no chest pain" becoming chest pain, or the reverse.
- Wrong value: incorrect dose, frequency, duration, age, temperature, blood pressure, date, number of episodes, timeframe, or investigation result.
- Wrong laterality or site: left/right, upper/lower, unilateral/bilateral, temporal/occipital, arm/leg, etc.
- Wrong subject: patient versus partner, child, parent, family history, clinician, or previous doctor.
- Hardened uncertainty: "possible", "likely", "to exclude", or "watchful waiting" turned into a definitive diagnosis or exclusion when that would affect care.
- Fabricated findings, diagnoses, allergies, examination signs, advice, referrals, or medications not supported by the transcript.
- Over-reassurance: "no red flags" or "normal examination" when the transcript contains a positive, equivocal, unasked, or undocumented red-flag point that matters.

Use the transcript as the source of truth. If the transcript is genuinely ambiguous, do not penalise a clinically reasonable paraphrase. But do penalise the note if it creates false certainty, a false denial, a false positive, or a misleadingly complete statement.

What not to penalise

Clinical notes legitimately compress. Do not penalise:

- Paraphrase into clinical language.
- Standard abbreviations and conventional normal-exam phrasing, if consistent with what happened.
- Combining repeated statements into one entry.
- Recording a fact once rather than every time it was said.
- Dropping conversational colour, pleasantries, hesitations, filler, or non-clinical digressions.
- Omitting exhaustive symptom detail when the load-bearing clinical content is still present.
- Different ordering, formatting, or note style.
- Minor wording differences in safety-netting or plans when the same trigger, urgency, and route remain clear.
- A concise note that a careful clinician would still regard as faithful and sufficient.

Severity calibration

Grade the consequence of the missing or wrong content, not the emotional intensity of the consultation.

Critical:
Use this when the absence or alteration could plausibly change diagnosis, triage, urgency, treatment, prescribing safety, investigation, referral, follow-up, safeguarding, or risk assessment; or when it removes a safety net or creates a latent safety risk. This includes important red flags, pertinent negatives used to rule out serious disease, allergies, medication changes/doses, abnormal examination findings, urgent referral reasons, escalation advice, and safety-net triggers. Any critical issue must receive a score low enough to be flagged for human review.

Supporting:
Use this when the issue materially weakens the clinical record, reasoning, handover, or defensibility, but is unlikely by itself to change immediate care. Examples include missing context for conservative management, incomplete chronology, or a non-urgent rationale that would help future interpretation. Supporting issues usually should not be flagged unless several together materially undermine the note.

Peripheral:
Use this for details with no meaningful bearing on diagnosis, management, risk, handover, or defensibility. Do not penalise peripheral omissions.

Scoring calibration

- A faithful and complete note should receive full credit, with no speculative issues added.
- Do not reduce the score merely because the note is concise or uses different wording.
- One critical omission, inversion, fabrication, or clinically material alteration is enough to make the note fail human-review threshold.
- Multiple critical problems, or a fabricated/inverted red flag, medication safety issue, allergy issue, or urgent-care issue, should receive a substantially lower score.
- Supporting issues may reduce the score modestly; multiple supporting issues can reduce it further if their combined effect materially weakens the record.
- If you are genuinely torn between supporting and peripheral, choose peripheral and do not penalise.
- If you are genuinely torn between critical and supporting, choose supporting only when patient care, safety-netting, triage, prescribing, or diagnostic risk would probably be unchanged.

Final check before scoring

Ask yourself:

1. What clinically important facts did the transcript establish?
2. Would a reader of the note know those facts?
3. Did the note change any yes/no answer, dose, timeframe, subject, laterality, diagnosis, urgency, or safety-net instruction?
4. Did a surviving vague mention hide the loss of the load-bearing detail?
5. Are any concerns merely stylistic compression or conversational detail?

Only penalise differences that matter clinically or materially weaken the record.

Return ONLY a JSON object on one line, with no other text before or after it:
{"score": <integer 0-10>, "omissions": [{"missing": "<what the note is missing, or what it asserts wrongly>", "severity": "critical|supporting|peripheral", "why": "<one line: what could go wrong because of it>"}]}

The score runs one way and one way only: 10 = the note fully satisfies the criterion, 0 = a severe violation. Leave "omissions" empty if you are penalising nothing.
\end{lstlisting}

\section{Pipeline diagnostics and deployment parameters}

This appendix carries the working detail behind Section 7: what each stage of the enumerate-then-check pipeline does and what it is allowed to see, the reasoning-effort deviation and the experiment that resolved it, two diagnostics that looked like mechanisms and did not survive the held-out run, the economics of the quote-verified second look, and the parameters a team would need to run either tier themselves. Two sets of pairs appear throughout and are never pooled. The \textbf{exploratory pass} covers 38 pairs against 32 clean notes, 70 note-judgements per tier. The \textbf{confirmation} covers 151 held-out pairs - 131 omissions (62 complete, 36 partial-weak, 33 partial-strong) plus 10 addition and 10 change pairs - against 47 clean twins over 47 consultations, 198 note-judgements per tier per replicate. Both tiers carry three replicates at seeds 11, 22 and 33, with extraction and audit cached across all three and zero parse failures in any tier or replicate, and the diagnostics in this appendix read replicate 1. Pair-id overlap between the two is zero. Following Section 7, \textbf{the two-stage pipeline} is B2 in the released data and \textbf{the three-stage pipeline} is B3. Those codes are the released data's names for the two tiers, and the pipeline as a whole is route one in the released data and the figures.

\subsection{The stages, and what each one is allowed to see}

The pipeline runs the same operations the benchmark's construction used, at inference time, in four stages - the three that name the three-stage tier, plus a second look that fires only on flagged absences. \textbf{Extraction} reads the transcript and writes a numbered list of clinically material facts, each with an id and its supporting evidence; it runs once per consultation and is cached to disk. \textbf{The audit} (three-stage pipeline only) reads the transcript and that candidate list, drops facts that are unsupported, duplicated or not something a note would be expected to record, and grades each survivor's severity against the written rubric of Section 3 (printed in Appendix A); it also runs once per consultation and is cached. \textbf{The check} is the only stage that sees the note under judgement: it receives the note and the numbered fact list and returns one verdict per fact id, present-or-absent on the two-stage pipeline and full/partial/absent on the three-stage pipeline. \textbf{The second look} (three-stage pipeline only) receives the note and only those facts the check judged absent, and may rescue a fact by finding it in the note.

Three contract details matter for anyone reproducing this. Verdicts are keyed to fact ids rather than returned as a list, so a mis-count is a set difference rather than a silent misalignment. Ids that come back unanswered trigger one targeted retry asking only about them, and anything still unanswered makes the record a parse failure rather than a fraction over an unknown denominator. In the confirmation neither tier needed that retry on any of its 198 note-judgements, and neither produced a parse failure. The second look must return the verbatim note text carrying any fact it claims the check missed, and a rescue is accepted only if that quote actually occurs in the note after whitespace and case normalisation, at eight characters or more. An unverifiable quote is counted and the absence stands. And the fact list shown to the check stage carries ids and fact text only: severity is withheld deliberately, because a checker told that a fact is critical is a checker being nudged about how hard to look.

Scoring and thresholds. The two-stage aggregate is the share of facts present. The three-stage aggregate is severity-weighted capture, crediting full 1.0, partial 0.5 and absent 0.0, with weights critical 4, supporting 2 and peripheral 1. Those weights are an indicative guess at relative clinical cost rather than a measured quantity, and the unweighted variant is recorded beside every note so nothing rests on them. Both tiers write their flag field as null on purpose: no flag threshold was fixed before the run, and every operating point in Section 7 and below is swept in analysis and quoted with the false-alarm rate at that same threshold.

What the stages produce, on the confirmation's 47 consultations: extraction yields 39.2 facts per note, and the audit cuts that to 37.6. The three-stage pipeline's 198 note-judgements therefore rest on 7,443 per-fact verdicts, 6,548 full, 675 partial and 220 absent after the second look; the two-stage pipeline's rest on 7,762 binary verdicts, 7,291 present and 471 absent.

\subsection{Blindness is a structural assertion, not a convention}

Extraction and audit must never see the note under judgement, because a fact list contaminated by the note is a fact list built from the thing it is used to score. Two mechanisms enforce that. The prompt builders for both stages take a transcript, and for the audit a candidate fact list, and nothing else; no note text is in scope inside either function. Before either call is made, \texttt{assert\_\hspace{0pt}blind} compares the built prompt against every note of that consultation: it normalises whitespace and case, takes three 120-character probes from each note (start, middle, end), and raises rather than running if any probe occurs in the prompt. A note is a generated structured document rather than a span of the transcript, so a hit would be a real leak instead of incidental overlap. Because the assertion raises rather than degrading, a completed run is itself the record that it never fired: all 37 extraction and 37 audit calls of the confirmation completed.

One further separation follows from the pinned roles rather than from an assertion: the model that writes the fact list is from a different family from the model that checks the note against it (Appendix D), so the enumerated facts are not the checking family's own.

\subsection{The reasoning-effort confound, and the experiment that removed it}

The eight monolithic judge designs ran their judge at no reasoning effort. The pipeline's check stage did not, and the reason is recorded in the runner. A two-consultation wiring pilot (4 note-judgements per tier) found that at no reasoning effort the checker answered a 36-fact keyed question in about 190 output tokens with zero reasoning tokens, roughly five tokens per verdict, and that across the pilot's four three-stage note-judgements it never once used the trichotomy's middle label: 147 full and 3 absent, no partials. A four-call re-probe at medium effort on the two clean notes of that pilot, costing \$0.06 and stored nowhere, produced 1,100 to 1,200 reasoning tokens and did use ``partial''. The check stage was therefore moved to medium effort, which left the confound plainly on the table: any win over the monolithic designs might be deliberation rather than task restructuring.

We tested it directly by re-running the two-stage pipeline on the identical 38-pair exploratory subset at no reasoning effort, changing nothing else. Effort was not the source of the gain. At no effort the tier scores 0.828 paired discrimination on omissions, where 0.5 is a coin flip, with swept detection of 31.2\% at a 9.4\% false-alarm rate; at medium effort on the same pairs it scores 0.813 paired and 25.0\% detection at the same 9.4\% false alarms. It is also cheaper, \$0.0074 against \$0.0346 per note-judgement, a factor of 4.65. The confirmation therefore ran the two-stage pipeline at no reasoning effort, the same setting the monolithic designs used, which removes the confound for that tier outright. The three-stage pipeline kept medium effort because its trichotomy demonstrably needs it. The records agree: all 198 two-stage confirmation note-judgements carry zero reasoning tokens, while the three-stage pipeline averages 1,834 reasoning tokens per note-judgement across its check and second-look calls combined.

\subsection{The trichotomy diagnostic did not replicate}

The analysis script re-scores the same purchased verdicts under variants. The most informative of them, ``lenient'', credits a partial verdict as full, so only an outright absent counts against a note; it is the cleanest read available on whether the trichotomy's middle label is doing work. On the exploratory pass it reached \textbf{46.9\% detection at 6.2\% false alarms} (32 omission pairs, 32 clean notes), the strongest number that pass produced. On the 131 held-out omission pairs against 47 clean twins it reaches \textbf{22.9\% at 8.5\%}. It more than halved. These variants are attribution diagnostics computed on verdicts already bought, not a menu of configurations to choose from, and picking the best of them would have been tuning on the exploratory pass, which is exactly what the halving punishes.

The trichotomy also failed the other half of what it was expected to do. It was designed on the expectation that the largest gain would fall on partial omissions, where the removed fact leaves some trace behind. At a swept threshold holding false alarms at or below 10\%, the three-stage pipeline detects 19.4\% of the 62 complete omissions, 17.4\% of the 69 partial omissions (weak and strong pooled) and 12.1\% of the 33 partial-strong ones. On the paired measure the ordering is the same and wider: complete 0.863 against 0.652 for partials pooled, and within partials 0.819 weak against 0.470 strong. What the audit stage actually bought was the severity grades, which are the ingredient the per-fact rule of Section 7 needs and the cheaper tier cannot supply: all 471 of the two-stage pipeline's missing-fact entries carry no severity at all.

\subsection{The quote-verified second look: real work on facts, no change at note level}

On the confirmation, the second look examined \textbf{299 flagged absences and rescued 79 of them, 26.4\%}, with \textbf{zero} rejections for an unverifiable quote; the quote-verification rail was live throughout and never had to fire, and no second-look call errored. Its effect on the headline is a wash. Re-scoring the same verdicts without it gives 0.752 paired discrimination on omissions with the stage against 0.763 without, and swept detection of 18.3\% against 17.6\% on the 131 held-out omission pairs.

It is not free. The stage fires only on note-judgements carrying at least one flagged absence, which was 124 of the confirmation's 198, so it adds 0.63 calls per note-judgement rather than one. Those 124 calls cost \$2.77 of the \$8.71 the three-stage pipeline spent on note-level calls across the confirmation's 198 note-judgements, which is \textbf{\$0.014 of that pipeline's \$0.044 marginal cost per note, or 32\%}, for no measurable change in either direction. The 26.4\% flip rate says the second look is doing real work on individual facts; the note-level figures say that work cancels out. The stage stays inside the three-stage pipeline Section 7 reports and prices, and Appendix D records its share of that pipeline's per-note cost separately for this reason.

\subsection{Fabrication anti-correlation on the addition pairs, at n=10}

The confirmation's 10 addition pairs - a fabricated fact inserted into an otherwise clean note - give the three-stage pipeline a paired score of \textbf{0.300}, below the 0.5 of a coin flip, which means it scores the note containing the fabrication above its own clean twin more often than not. The two-stage pipeline on the same 10 pairs scores 0.600, the three-stage pipeline on the 10 change pairs 0.700, and the exploratory pass's 3 addition pairs gave the three-stage pipeline 0.833. The mechanism is plausible enough on its face: a coverage score rewards a note that says more, and neither tier ever asks whether the note contains content the transcript does not support, which is the commission-side blindness Section 7 describes. At 10 pairs this is a signal to size properly rather than a finding, and it is reported that way here.

\subsection{Deployment parameters}

The full deployment parameterisation ships with the release rather than here: per-stage model roles, reasoning-effort settings, what each stage receives, call schedules, caching keys, and the measured per-note costs in both framings. The pipeline runner (\texttt{w2\_\hspace{0pt}pipeline.py}), its prompt files (\texttt{w2\_\hspace{0pt}prompts/\hspace{0pt}}) and the run manifests carry every setting, and Appendix D records the per-note prices with their measurement dates. Three facts from that record stay in the paper because results rest on them. At the one-note-per-consultation price, the audit stage is 71\% of the three-stage pipeline's \$0.45 per note, which is why Section 7 names folding severity grading into the extraction call, and a cheaper extraction model, as the two untested cost levers. The confirmation bought 37 extractions and 37 audits for its 47 consultations, because 10 of those consultations also appear in the exploratory pass and their cached stages were reused - so the confirmation is a fully held-out test of the judging and only a partial one of the extraction, as Section 8 states. And latency was not measured: the harness records token receipts, cost and a completion timestamp per call, but no per-call wall-clock duration, so no latency figure exists and none should be inferred from the timestamps.

\section{Reproducibility}

This appendix records what a replicator needs in order to regenerate the study: which model ran in which role, on what transport and at which settings, and how runs were seeded, hashed and identified. Section 9 lists the harness, prompts, judge configurations, model pins and run manifests among the released artefacts, and this appendix is the description of those objects. The per-note prices a practitioner would pay are in Section 7, and are reconciled to their measurement dates in D.3 below.

The full versions of what this appendix summarises are release material rather than print. The lock file itself (\texttt{models.lock.json}) and the validator every manifest was written to satisfy (\texttt{check\_\hspace{0pt}manifests.py}) ship with the harness; the run manifests themselves, with their per-call logs and the prompt and dataset digests every manifest records, ship in the data repository beside the judgements they describe. Transports, hashes and the manifest schema are therefore checkable in the release rather than printed here. Of the run manifests, those of the released benchmark judge runs ship, complete runs only; incomplete runs are excluded at packaging. Each manifest also records the state of the working tree its run started from, including the exact uncommitted paths where there were any.

\subsection{Model pins}

Every call that stands behind a number in this paper goes through a single function that resolves a named role against a lock file, \texttt{models.lock.json}, and refuses to run if the role is not pinned there. The lock records, per role, the provider slug, the provider-routing block sent with each request, and the credit price per million tokens at pin time. It was pinned on 2026-07-29, and extended twice while the study ran: the construction role gained an explicit routed entry on 2026-08-10 when construction moved off a subscription path onto metered credits, and the second-family judge was added on 2026-08-17. Slugs and prices were verified live against the provider's model listing on each of those three dates.

\begin{sidewaystable}
\centering
\footnotesize
\setlength{\tabcolsep}{4pt}
\begin{tabular}{@{}>{\raggedright\arraybackslash}p{0.95in}>{\raggedright\arraybackslash}p{1.45in}>{\raggedright\arraybackslash}p{1.12in}>{\raggedright\arraybackslash}p{1.80in}>{\raggedright\arraybackslash}p{0.62in}>{\raggedright\arraybackslash}p{2.36in}@{}}
\toprule
role (lock key) & pinned model & transport and route & pinned settings & credit price in / out per Mtok & where it ran\\
\midrule
\texttt{judge-primary} & \texttt{openai/\hspace{0pt}gpt-5.4} & OpenRouter, provider order \texttt{["openai"]}, fallbacks off & temperature 1.0, reasoning effort none, 1,024 max output tokens; three disclosed deviations, each recorded per run (the three-stage pipeline's check stage at medium effort, the census verification panel at high, and the cached per-consultation extractions and checklists at temperature 0) & \$2.50 / \$15.00 & the judge in all eight ablation designs, every reference judge, the RAGAS-style recipe, the external commission anchor, the check stage of both pipeline tiers in the primary family, and the student prompt in all three optimisation campaigns (188 run manifests)\\
\addlinespace
\texttt{auditor} & \texttt{openai/\hspace{0pt}gpt-5.5} & OpenRouter, \texttt{["openai"]}, fallbacks off & temperature 1.0, reasoning effort high, except medium in the census work and as the reflection model of the confirmatory optimisation campaign & \$5.00 / \$30.00 & the answer-key audit and its semantic checks, the omission-pair verification panel, the severity grading, the pipeline's audit stage and its quote-verified second look, and the reflection model in the optimisation campaigns (175 manifests)\\
\addlinespace
\texttt{constructor} & \texttt{anthropic/\hspace{0pt}claude-opus-5} & OpenRouter, \texttt{["anthropic"]}, fallbacks off & temperature 1.0, reasoning effort medium, 16,000 max output tokens & \$5.00 / \$25.00 & benchmark construction (reference-note repairs, regenerations, severity grades, omission injection, the fact-site map), the census discovery passes and cluster labelling, and the pipeline's extraction stage (112 manifests)\\
\addlinespace
\texttt{judge-gemini} & \texttt{google/\hspace{0pt}gemini-3.1-pro-preview} & OpenRouter, \texttt{["google-vertex"]}, fallbacks off & temperature 1.0, reasoning effort minimal, 1,024 max output tokens & \$2.00 / \$12.00 & the second-family replication of two designs, and the check stage of both pipeline tiers in the transfer run (24 manifests)\\
\addlinespace
\texttt{judge-opus} & \texttt{anthropic/\hspace{0pt}claude-opus-4.8} & OpenRouter, \texttt{["anthropic"]}, fallbacks off & temperature 1.0, 1,024 max output tokens; high reasoning effort in the verification panel, none sent in the second-family probe & \$5.00 / \$25.00 & the second member of the cross-family omission-verification panel, and the second-family candidate probe (9 manifests)\\
\addlinespace
\texttt{judge-qwen} & \texttt{qwen/\hspace{0pt}qwen3-235b-a22b-2507} & OpenRouter, no provider constraint sent & temperature 1.0, 1,024 max output tokens, on a non-thinking variant that takes no reasoning parameters & \$0.09 / \$0.55 & the open-weights row in the census work (1 manifest)\\
\addlinespace
\texttt{author-offplan} & \texttt{openai/\hspace{0pt}gpt-5.5} & OpenRouter, \texttt{["openai"]}, fallbacks off & reasoning effort none unless a spec says otherwise & \$5.00 / \$30.00 & authoring, verification and critic support; no run manifest records a call under this role name, since work of that kind that became paper-bound ran under \texttt{auditor}\\
\bottomrule
\end{tabular}
\end{sidewaystable}

The manifest counts in the table were recomputed on 2026-08-17 from the 424 run manifests then present; at the 25 Aug freeze the tree holds 569 run manifests, 539 of them complete; the 240 describing the released benchmark judge runs ship in the data repository, alongside the census pipeline's own manifests under \texttt{judgements/\hspace{0pt}census/\hspace{0pt}}. A run that used two roles appears in both rows. Two rows carry pins that were used lightly and are reported as such rather than as further judge families: the open-weights row appears in one run, and the second Anthropic row in nine. The Anthropic route lists no seed support, so seeding there is best-effort and the run-to-run spread, not the seed, is what stands in for determinism.

\textbf{The second-family judge and its reasoning setting.} The eight ablation designs run their judge at no reasoning effort. The second family's endpoint refuses that setting outright, returning an HTTP 400 with ``Reasoning is mandatory for this endpoint and cannot be disabled'' when probed on 2026-08-17, so ``minimal'', the nearest available analogue, is what is pinned. Judge family and reasoning setting therefore move together in every second-family comparison, and no run in this study can separate them; Section 8 states the confound and Appendix G carries the results it qualifies. In the pipeline transfer run only the check stage moved to this family: extraction and audit were read from cache and are identical, character for character, to the primary-family run, and the second look stayed on the auditor model, so fact lists and severity grades are the same objects across both checkers. Checker reasoning effort is not matched tier for tier across families (in the primary family the two-stage pipeline ran at none and the three-stage at medium, while the second family ran both at minimal), so deliberation is confounded with family in both directions there as well.

\textbf{Models used outside the lock.} Two, both outside the judging path: the census clustering's embedding model (named with its projection and clustering parameters in the companion census paper's method appendix) and the text-to-speech renderer for the synthesised consultation audio, \texttt{gpt-4o-mini-tts}, named in the capture script rather than in the lock. Both sit outside the study's credit ledger: they were paid by card rather than credits and are not itemised in the study's artefacts. One further category needs stating rather than pinning: the three scribe products under audit are commercial systems whose internal models their vendors do not publish, so there is no pin to record for them and none is claimed.

\subsection{Seeds}

Replicates carry run seeds 11, 22 and 33. Within a run, each item takes \texttt{base\_\hspace{0pt}seed\ =\ run\_\hspace{0pt}seed\ x\ 100,000\ +\ item\_\hspace{0pt}index\ x\ 10}, and sample \emph{i} of an ensemble takes \texttt{base\_\hspace{0pt}seed\ +\ i}, so an eight-sample judgement occupies \texttt{base\_\hspace{0pt}seed+0} to \texttt{base\_\hspace{0pt}seed+7} and no two judgements can collide. Seeds that fix a selection rather than a sample are recorded the same way: the held-out confirmation subset was drawn at seed 41, the audit-stage severity validation pack at seed 20260817, and the clustering's dimensionality reduction at seed 42 with a stability sweep over 42, 43 and 44.

Seeds are accepted on the primary judge's route and on the second family's, and they do not make output reproducible at the token level. On the routing probe of 2026-07-30 the same prompt at temperature 0 and at temperature 1 returned different text on repeat calls while the verdict was stable in 4 of 4 probes. This is why the released raw completions, rather than the seeds, are what a replicator should check against: a rerun should reproduce our verdicts and our rates, not our strings.

\subsection{The per-note prices in Section 7, and when they were measured}

Section 7's price ladder is quoted per note, on the pinned routes of D.1, and every figure in it comes from the provider's own usage accounting for the runs named below rather than from a price list. These are what these models cost on those dates, and they move with any repricing.

{
{\small\begin{longtable}[]{@{}
  >{\raggedright\arraybackslash}p{(\linewidth - 6\tabcolsep) * \real{0.2500}}
  >{\raggedright\arraybackslash}p{(\linewidth - 6\tabcolsep) * \real{0.2500}}
  >{\raggedright\arraybackslash}p{(\linewidth - 6\tabcolsep) * \real{0.2500}}
  >{\raggedright\arraybackslash}p{(\linewidth - 6\tabcolsep) * \real{0.2500}}@{}}
\toprule\noalign{}
\begin{minipage}[b]{\linewidth}\raggedright
price as quoted
\end{minipage} & \begin{minipage}[b]{\linewidth}\raggedright
what it buys
\end{minipage} & \begin{minipage}[b]{\linewidth}\raggedright
how it is composed
\end{minipage} & \begin{minipage}[b]{\linewidth}\raggedright
measured
\end{minipage} \\
\midrule\noalign{}
\endhead
\bottomrule\noalign{}
\endlastfoot
\$0.005 & the engineered single-call judge, whose own critical-omission list is Section 7's cheapest flag that names a fact & one call per note at the judge settings of Section 4 & reference-judge run receipts, 12--13 August 2026 \\
about \$0.03 to \$0.07 & the RAGAS-style recipe with a calibrated threshold & \$0.0305 per note measured, at 1.98 calls per note with the transcript fact extraction cached per consultation & cost read of 2026-08-14, over records bought 10--14 August 2026 \\
\$0.036 & the eight-sample monolithic judge the two-stage pipeline is compared against & eight independent calls summing to \$0.0360, against \$0.0045 for a single call; the provider's prompt cache landed on all eight, so effective input cost is about 3.1 times a single call rather than eight & routing probe of 2026-07-30, at prices pinned 2026-07-29 \\
\$0.046 & the evolved prompt (route two in the released data) & one call per note at high reasoning effort under a 12,000-token cap & reasoning-budget run receipts, 18 and 24 August 2026 \\
\$0.094 & the two-stage pipeline at one note per consultation & extraction \$0.0868 per consultation plus \$0.0077 per note for the single keyed check call, which is \$0.0945 & confirmation run receipts, 13--14 August 2026 \\
\$0.45 & the three-stage pipeline at one note per consultation & extraction \$0.0868 plus audit \$0.3194 per consultation, plus \$0.0440 per note for the check and the second look, which is \$0.4502; the audit stage is 71\% of that total & confirmation run receipts, 13--14 August 2026 \\
\$0.028 and \$0.140 & the same two tiers amortised over the 4.21 notes per consultation the held-out subset holds & \$0.0283 and \$0.1404 fully loaded & confirmation run receipts, 13--14 August 2026 \\
\end{longtable}\addtocounter{table}{-1}}
}

Two conventions travel with those numbers. The one-note-per-consultation framing and the amortised framing must both be labelled wherever either is quoted: the extraction and audit stages run once per consultation and are cached, so the two framings differ by more than a factor of three on both tiers and describe different deployments. A practitioner evaluating one note per consultation pays the first, and a service evaluating several notes against one transcript pays the second. And within the three-stage tier the second look accounts for \$0.014 of the \$0.044 marginal per-note cost, 32\% of that tier's per-note spend, for no measurable change in detection in either direction (Appendix C).

The \$0.03 to \$0.07 range for the RAGAS-style recipe spans its two measured prices: \$0.0305 per note with the per-consultation extraction cached across a consultation's notes, and \$0.069 measured with the keyed contract and full token budgets (from the baselines runs' own receipts, released beside their manifests in the data repository).

\section{Clinician validation in full: both sittings}

Every other measurement in this study is a model judging a model. Two sittings put a doctor under that stack. The first, on 14 August 2026, covered 70 items in a sitting that spanned six stages of the pipeline and the benchmark and is the source of the adjudication reported in Section 7. The second, on 17 August 2026, covered 24 items and validates the one thing the first did not reach: the severity grades the deployable per-fact rule actually fires on. This appendix carries both in full, and reports the standing conflict with them: the rater is a physician author of this study who designed the instruments they are grading and knew the study's claims before they started.

Two rules govern how the two sittings are read together. First, they measure different instruments and their numbers are never pooled. The first sitting's severity stage validates the \textbf{benchmark's} severity grades, which are the axis every severity-conditioned result in Sections 5 to 7 is conditioned on. The second validates the \textbf{audit stage's own} grades of its own extracted facts, a separate run of the same written rubric and what the per-fact rule of Section 7 fires on. That rule flags the note if any fact graded critical is judged absent. Second, blinding differs by stage and the label travels with every rate. Where an item has to state the machine verdict in order to be answerable at all, the answer is an \textbf{endorsement} - a clinician checked and did not object - and where the item can be answered without the verdict on screen, the answer is a blinded \textbf{measurement}. Stages A, B, C and G of the first sitting are endorsements; D, E and F are measurements, and so is the second sitting's blind pass.

\subsection{The 70-item sitting, 14 August 2026}

\subsubsection{The pack, its re-cut, and how the keys were recovered}

The pack was generated at seed 20260809 by a deterministic builder, \texttt{build\_\hspace{0pt}sitting\_\hspace{0pt}pack.py}, drawing its items from the study's final artefacts. The builder is released in the code repository; the pack itself prints withheld note excerpts and is not released. It was sat in one uninterrupted session on 2026-08-14, 14:37 to 15:35 UTC - 58 minutes against the pack's 75-minute budget - and all 70 items were answered, with four free-text comments. Abstention was offered explicitly on the adjudication sections and recorded as an abstention rather than a guess.

The pack was re-cut before it was sat. An 11 August version of the same instrument is kept verbatim in the study's internal archive: same shape and same section design, but half its sections sampled artefacts the study had since replaced, so it was regenerated on 14 August against the final ones and only the regenerated pack was answered.

{
{\small\begin{longtable}[]{@{}
  >{\raggedright\arraybackslash}p{(\linewidth - 8\tabcolsep) * \real{0.2000}}
  >{\raggedright\arraybackslash}p{(\linewidth - 8\tabcolsep) * \real{0.2000}}
  >{\raggedright\arraybackslash}p{(\linewidth - 8\tabcolsep) * \real{0.2000}}
  >{\raggedright\arraybackslash}p{(\linewidth - 8\tabcolsep) * \real{0.2000}}
  >{\raggedright\arraybackslash}p{(\linewidth - 8\tabcolsep) * \real{0.2000}}@{}}
\toprule\noalign{}
\begin{minipage}[b]{\linewidth}\raggedright
section
\end{minipage} & \begin{minipage}[b]{\linewidth}\raggedright
what the clinician was doing
\end{minipage} & \begin{minipage}[b]{\linewidth}\raggedright
items
\end{minipage} & \begin{minipage}[b]{\linewidth}\raggedright
answer codes
\end{minipage} & \begin{minipage}[b]{\linewidth}\raggedright
pack's minute budget
\end{minipage} \\
\midrule\noalign{}
\endhead
\bottomrule\noalign{}
\endlastfoot
\textbf{A} fact-list extraction audit & does the cited quote support the extracted fact & 10 & \texttt{y} / \texttt{n} / \texttt{?} & 8 \\
\textbf{B} repaired-note spot checks & was the flag a real defect and did the repair fix it & 8 & \texttt{y} / \texttt{n} / \texttt{?} & 9 \\
\textbf{C} rejected-pair adjudications & was the pair rightly rejected & 6 & \texttt{y} / \texttt{n} / \texttt{?} & 9 \\
\textbf{D} panel-killed findings & was a cut finding really not a finding & 10 & \texttt{y} / \texttt{n} / \texttt{?} & 11 \\
\textbf{E} severity grades & grade the fact's severity & 20 & \texttt{c} / \texttt{s} / \texttt{p} & 18 \\
\textbf{F} judge disagreements & is the fact in the note or not & 10 & \texttt{y} / \texttt{n} / \texttt{?} & 13 \\
\textbf{G} realism and fidelity & rate realism or fidelity & 3 & \texttt{1}-\texttt{5} & 4 \\
\textbf{Q} standing methodology questions & three open questions & 3 & free text & 3 \\
\end{longtable}\addtocounter{table}{-1}}
}

Keys were recovered after the fact rather than held alongside the answers: we re-ran the builder's own per-section draws over the same artefacts at the same seed and checked each recovered key item by item against the pack's printed headers and fact text. Every key was recovered deterministically and nothing was left unscoreable for want of one - the artefact records 64 of 64 recovered keys, the six items with no key by construction being section G's three ratings and section Q's three free-text answers. The full join of answer to key is released as \texttt{validation/\hspace{0pt}sitting\_\hspace{0pt}results.json} in the data repository.

Three conventions apply to every rate below. Each carries a Wilson 95\% interval, quoted every time, because n is 6 to 20 per stage and the interval is what the result actually is. Abstentions are reported as abstentions and never imputed: there were two, both in section D and none anywhere else. And ``agreement'' always means agreement with a named machine verdict, which is not the same thing as correctness.

\subsubsection{Blinding, stage by stage}

{
{\small\begin{longtable}[]{@{}
  >{\raggedright\arraybackslash}p{(\linewidth - 6\tabcolsep) * \real{0.2500}}
  >{\raggedright\arraybackslash}p{(\linewidth - 6\tabcolsep) * \real{0.2500}}
  >{\raggedright\arraybackslash}p{(\linewidth - 6\tabcolsep) * \real{0.2500}}
  >{\raggedright\arraybackslash}p{(\linewidth - 6\tabcolsep) * \real{0.2500}}@{}}
\toprule\noalign{}
\begin{minipage}[b]{\linewidth}\raggedright
stage
\end{minipage} & \begin{minipage}[b]{\linewidth}\raggedright
what it audits
\end{minipage} & \begin{minipage}[b]{\linewidth}\raggedright
how it was blinded
\end{minipage} & \begin{minipage}[b]{\linewidth}\raggedright
reads as
\end{minipage} \\
\midrule\noalign{}
\endhead
\bottomrule\noalign{}
\endlastfoot
\textbf{A} & the deployed extractor's inference-time fact list & not blindable: the item prints the fact and its quote, which is the verdict under audit & endorsement \\
\textbf{B} & the reference-note repair loop & not blindable: the item prints the flag and the repair & endorsement \\
\textbf{C} & the omission-pair verification panel & not blindable: the item prints the rejection and its reasons & endorsement \\
\textbf{D} & the taxonomy panel's cuts & the audit classifier's four buckets were interleaved and never named on screen & measurement \\
\textbf{E} & the benchmark's severity rubric & no machine grade printed, and no indication of which items the two graders split on & measurement \\
\textbf{F} & the three-stage pipeline against the best monolithic judge & neither which judge said what, nor whether the note shown was an original or an edited one & measurement \\
\textbf{G} & scenario realism and fact-sheet fidelity & a rating, not an adjudication; the rater designed the material & endorsement \\
\end{longtable}\addtocounter{table}{-1}}
}

\subsubsection{The per-stage rates}

{
{\small\begin{longtable}[]{@{}
  >{\raggedright\arraybackslash}p{(\linewidth - 10\tabcolsep) * \real{0.1667}}
  >{\raggedright\arraybackslash}p{(\linewidth - 10\tabcolsep) * \real{0.1667}}
  >{\raggedright\arraybackslash}p{(\linewidth - 10\tabcolsep) * \real{0.1667}}
  >{\raggedright\arraybackslash}p{(\linewidth - 10\tabcolsep) * \real{0.1667}}
  >{\raggedright\arraybackslash}p{(\linewidth - 10\tabcolsep) * \real{0.1667}}
  >{\raggedright\arraybackslash}p{(\linewidth - 10\tabcolsep) * \real{0.1667}}@{}}
\toprule\noalign{}
\begin{minipage}[b]{\linewidth}\raggedright
stage
\end{minipage} & \begin{minipage}[b]{\linewidth}\raggedright
stage under judgement
\end{minipage} & \begin{minipage}[b]{\linewidth}\raggedright
n
\end{minipage} & \begin{minipage}[b]{\linewidth}\raggedright
what agreement means
\end{minipage} & \begin{minipage}[b]{\linewidth}\raggedright
rate
\end{minipage} & \begin{minipage}[b]{\linewidth}\raggedright
95\% interval
\end{minipage} \\
\midrule\noalign{}
\endhead
\bottomrule\noalign{}
\endlastfoot
\textbf{A} & the deployed extractor's inference-time fact list & 10 & the cited quote supports the extracted fact & 10/10 = 100\% & 72.2--100\% \\
\textbf{B} & the reference-note repair loop & 8 & the flag was a real defect and the repair fixed it & 8/8 = 100\% & 67.6--100\% \\
\textbf{C} & the omission-pair verification panel & 6 & the panel was right to reject the pair & 6/6 = 100\% & 61.0--100\% \\
\textbf{D} & the taxonomy panel's cuts & 10 & \emph{false-kill rate}: a cut finding was a real omission & 0/10 = 0\% & 0--27.8\% \\
\textbf{E} & the benchmark's two-grader severity rubric & 20 & exact grade match & 14/20 = 70\% & 48.1--85.5\% \\
\textbf{F} & the three-stage pipeline against the best monolithic judge & 10 & the \textbf{pipeline's} per-fact verdict matches & 10/10 = 100\% & 72.2--100\% \\
\textbf{G} & scenario realism, extraction fidelity & 3 & a 1--5 rating, not an agreement & 5, 5, 5 & - \\
\end{longtable}\addtocounter{table}{-1}}
}

\subsubsection{A, B and C: the three endorsements, and what they rule out}

\textbf{A, the deployed extractor (10 of 10).} One fact per consultation over the 47 held-out confirmation consultations, drawn from the 1,833 facts the pipeline wrote at inference time, with no filtering for interestingness. On all ten the clinician judged that the quote the extractor cited supports the fact it wrote. The pipeline's own audit stage had kept all ten; corpus-wide that stage drops 9 facts as unsupported, 62 as not something a note would be expected to record and 43 as duplicates, so clinician and audit agree on 10 of 10. The limit is the sampling. The draw contains no fact the audit rejected, so this measures the \textbf{extractor's} precision and says nothing about whether the audit catches the bad ones. What 10 of 10 rules out is an extractor whose quotes support its facts less than about seven times in ten; a 90\% extractor is entirely compatible with it.

The per-item facts, quotes and recovered keys are in the released sitting artefact.

Stratum labels in the tables below are the corpus strata of Section 3: \texttt{primock} and \texttt{aci} are the two public consultation corpora, \texttt{authored} and \texttt{trapblind} the two sets of scenarios we wrote ourselves, the second holding deliberate documentation traps.

\textbf{B, the repair loop (8 of 8).} Eight repaired reference notes, stratified by corpus stratum and by how heavily the auditor flagged them, from single-flag notes up to three-round repairs. Every one was judged a real defect properly fixed. The two failure modes the item was written to separate - an auditor that fires on nothing, and a repair that is cosmetic - are both absent from this sample. At n=8 a defect rate under about a third would not show.

The per-item flags, repair rounds and recovered keys are in the released sitting artefact.

\textbf{C, the verification panel (6 of 6).} Six of the 92 rejected pairs, one per failure signature. All six rejections were upheld. This is the endorsement that props up a headline directly: the 44\% rejection rate on the complete-removal class (Section 3) is a finding about clinical notes only if the panel's standard is right, and a clinician reading one pair per failure mode agreed with it every time. The lower bound is 61\%, which excludes a panel that is wrong more than a third of the time and nothing finer.

The per-item removed facts, failing checks and recovered keys are in the released sitting artefact.

\subsubsection{\texorpdfstring{D: the taxonomy panel's cuts, and the audit classifier's \texttt{wrongly\_\hspace{0pt}cut} bucket}{D: the taxonomy panel's cuts, and the audit classifier's wrongly\_cut bucket}}

Ten of the 40 audited omission refusals from the companion census paper's taxonomy work, interleaved so the audit classifier's bucket never leaked, each asking ``reading the note, is this a real omission the study should have counted?''.

They called none of the ten a reportable omission: 8 fair cuts, 2 abstentions, 0 real omissions. The \textbf{false-kill rate is 0/10 = 0\% {[}0, 27.8{]}}, one-sided 95\% upper bound 25.9\%; on answered items only it is 0/8 = 0\% {[}0, 32.4{]}, and on the seven items outside the \texttt{wrongly\_\hspace{0pt}cut} bucket discussed next it is 0/7 = 0\% {[}0, 35.4{]}, both abstentions counted in that denominator. For the taxonomy's headline this is supportive: a clinician reading ten of the panel's kills finds no reportable omission among them, so the panel's overall survival rate of 10.48\% of candidates, and 6.1\% on omissions specifically, is not obviously a panel suppressing real findings.

\textbf{Agreement with the audit classifier's own bucketing is weaker, and every disagreement sits in one bucket.} Of the eight answered items they match the bucket's implied answer on five, \textbf{5/8 = 62.5\% {[}30.6, 86.3{]}}, and all three misses are items the classifier had placed in \texttt{wrongly\_\hspace{0pt}cut} - the bucket whose whole job was to find the panel's mistakes. It found three in forty, and the clinician rejects all three. So the audit's 7.5\% wrongly-cut share (3 of 40) should be read as an upper bound produced by a model, not as a measured false-kill rate: on the only sample a clinician has seen, the model auditing the panel was harsher on the panel than the doctor was.

\textbf{The \texttt{present\_\hspace{0pt}elsewhere} check barely happened, and that is the important gap.} Of the three items in that bucket they answered one, agreeing the cut was fair, and abstained on two. The taxonomy's \texttt{present\_\hspace{0pt}elsewhere} share of 25\%, which is the live caveat on the whole omission rate, is therefore essentially unvalidated by this sitting and the sitting must not be cited in support of it.

\textbf{Both abstentions are the pack's doing.} Eight of the ten D items printed only the part of the note covering the point at issue; two printed the note in full. D4's comment is exactly that: \emph{``think i would need to see more of the note to judge really, can't see anything heere''}. D1 carries no comment but is the same shape, a \texttt{present\_\hspace{0pt}elsewhere} item where the fact could be sitting in the unprinted remainder. Both are abstentions on the pack rather than on the finding, and the honest denominator for this stage is 8. The same fact qualifies the eight answers: they were given without the whole note in view, so on a \texttt{wrongly\_\hspace{0pt}cut} item ``fair cut'' is a judgement that the fact did not have to be in the note, not a check that it was somewhere else in it.

The full per-item table - each item's note, the claimed omission, the audit classifier's bucket with its implied answer, and the clinician's answer against it - is in the released sitting artefact (\texttt{validation/\hspace{0pt}sitting\_\hspace{0pt}results.json}), item by item.

\subsubsection{E: the benchmark's severity grades}

Twenty facts drawn from the benchmark's severity strata, graded blind against the written rubric of Section 3 (printed verbatim in Appendix A). The key is the consensus grade of the benchmark's two graders, which are two models from different families running the same rubric, with a split resolved to the lower grade.

\begin{itemize}
\tightlist
\item
  \textbf{Exact agreement 14/20 = 70.0\% {[}48.1, 85.5{]}.}
\item
  \textbf{Weighted agreement, an adjacent grade counting half: 85.0\%.}
\item
  \textbf{Linear-weighted kappa 0.63}, where kappa is chance-corrected agreement and 0 is what two raters would reach by guessing (observed agreement 0.850, expected 0.595; percentile bootstrap 95\% 0.32--0.86 over 10,000 seeded resamples, indicative at n=20). Unweighted Cohen's kappa 0.55.
\item
  \textbf{No disagreement is more than one grade.} Six adjacent, zero two-grade.
\end{itemize}

Confusion matrix, rows the machine consensus, columns the clinician:

{
{\small\begin{longtable}[]{@{}llll@{}}
\toprule\noalign{}
machine ~clinician & critical & supporting & peripheral \\
\midrule\noalign{}
\endhead
\bottomrule\noalign{}
\endlastfoot
\textbf{critical} (7) & 3 & 4 & 0 \\
\textbf{supporting} (7) & 1 & 6 & 0 \\
\textbf{peripheral} (6) & 0 & 1 & 5 \\
\end{longtable}\addtocounter{table}{-1}}
}

\textbf{The disagreement lives in the machine's critical row.} Four of the six disagreements are the clinician moving a machine \texttt{critical} down to \texttt{supporting} (E1, E7, E9, E10); of the seven facts the two graders called critical they kept three. The other two run upward, one \texttt{supporting} to \texttt{critical} (E19) and one \texttt{peripheral} to \texttt{supporting} (E8). Their overall mix is 4 critical / 11 supporting / 5 peripheral against the machine's 7 / 7 / 6 - the same middle-heavy shape, with them the more conservative grader at the top. That is the opposite of the direction the rubric's resolve-downward rule was written to guard against. If the critical stratum is over-inclusive by anything like the four-in-seven this sample shows, every severity-conditioned result is conditioned on a slightly generous critical class. At seven machine-critical items that is something to check, not a correction to apply.

\textbf{The items the two graders split on are the ones they match best, which is the reverse of the natural worry.} On the 8 items where the graders disagreed and the tie was resolved downward, \textbf{7/8 = 87.5\% {[}52.9, 97.8{]}}; on the 12 they agreed on, \textbf{7/12 = 58.3\% {[}32.0, 80.7{]}}. Fisher's exact test gives two-sided p = 0.32, so this is not a significant difference and may well be noise, but it is worth recording because a reader will assume the split cases are the shaky ones. Here they are not.

Their three borderline comments all sit on items they \textbf{agreed} with, and two of the three say they would have accepted the adjacent grade. They are the qualitative version of the gap between 70\% exact and 85\% weighted:

\begin{itemize}
\tightlist
\item
  \textbf{E4} (machine critical, they said critical): \emph{``should document those symtpoms, don't need each one exactly written literally, but it's quite critical to document that they were well overall and that that general vibe of things was asked about''}
\item
  \textbf{E12} (machine peripheral, they said peripheral): \emph{``could really be supporting or peripheral''}
\item
  \textbf{E17} (machine critical, they said critical): \emph{``i might accept supporting though too''}
\end{itemize}

The twenty facts, their machine-consensus keys, the clinician's answers and the exact/adjacent scoring are in the released sitting artefact, item by item. Some items grade a handling instruction rather than a bare fact, because for those the material at risk is defined by what a correct note would have recorded; the grading task is the same either way.

\subsubsection{F: the judge disagreements, adjudicated}

Ten notes where the three-stage pipeline's per-fact verdict and the best monolithic judge's note-level flag reach opposite conclusions, drawn one per consultation from the 35 such notes in the confirmation run (21 the pipeline flags and the monolithic judge does not, 13 the reverse, and 1 clean note the per-fact rule fires on). The monolithic judge here is the strongest of the eight designs, the completeness-scored eight-sample judge (\texttt{FC-score-k8} in the released data), read across three replicates. The clinician was told neither which judge said what nor whether the note in front of them had been edited, and was told explicitly that \texttt{?} was a real answer, because how much of this disagreement is genuine ambiguity is itself a finding.

\textbf{They used \texttt{?} zero times, and on all ten their verdict matches the pipeline's per-fact verdict and contradicts the monolithic judge's.} Pipeline \textbf{10/10 = 100\% {[}72.2, 100{]}}, monolithic judge \textbf{0/10 = 0\% {[}0, 27.8{]}}, and on the 10 discordant pairs \textbf{p = 0.002}.

The full adjudication table - each item's case type, the note's pair class and benchmark severity, the fact at issue, and all three verdicts (pipeline, monolithic across its three replicates, clinician) - is in the released sitting artefact (\texttt{validation/\hspace{0pt}sitting\_\hspace{0pt}results.json}), item by item.

Two severity instruments are in play in that record and they must not be conflated. The benchmark severity above is the two-grader benchmark grade, which is the axis section E validates. The \texttt{critical} grade the per-fact rule of Section 7 actually fires on is the \textbf{pipeline audit stage's own} grade of its own extracted fact, a different run of the same written rubric, which is why item F8 sits here as a \texttt{supporting} pair whose extracted fact the pipeline graded critical. Stage E does not validate the grades the rule fires on; it validates the axis the results are conditioned on. Nothing in this sitting puts a clinician under the audit stage's severity grading, which is what the second sitting in E.2 exists to do.

\textbf{The six cases where the pipeline flags and the monolithic judge is silent are like-for-like, and they are the study's claim adjudicated on real cases.} In all six the fact the pipeline named absent is the fact construction actually removed, the strongest monolithic design did not flag the note at all across three replicates, and the clinician confirms the fact is not in the note. A missing clindamycin allergy, a missing distal pulse check, a missing positive Murphy's sign: the monolithic judge read the whole note and scored it fine.

\textbf{Two of the three reverse cases are not like-for-like and should not be quoted as clean losses for the monolithic judge.} Item F4's note is a \texttt{change} pair and item F5's carries a partial-strong omission of a fact the item never shows, so on those two the note-level flag can be earned by a defect the clinician was not asked about, while the pipeline is right about the specific fact. F2 is the clean one: an unedited note flagged in one replicate of three, with the clinician confirming the fact is present. Restricting to the \textbf{8 items where the monolithic judge's position is attributable to the fact in front of them}: pipeline 8/8 = 100\% {[}67.6, 100{]}, monolithic judge 0/8 = 0\% {[}0, 32.4{]}, p = 0.008. Both the 10--0 headline and the conservative 8--0 reading hold.

\textbf{And the false alarm is not a false alarm.} Item F6 is the confirmation run's single false positive under Section 7's per-fact rule, recomputed here from the three-stage pipeline's confirmation records (\texttt{confirm-B3.jsonl}) as 27 of 131 detection and 1 of 47 on clean notes, exactly the figures Section 7 reports, on the clean reference note for \texttt{aci\textbar{}D2N199}. The flagged fact is ``patient is aged 74''. The transcript says it. The repaired reference note never states an age at all. The clinician's verdict is that the fact is not captured - a real omission of the note. But the rule's trigger is a fact graded critical, and this fact graded supporting in both of the clinician's passes (the second sitting's reconciliation, below), so under the rule as stated the flag stands as a false alarm, exactly as the body counts it. This is one case rather than a rate. It also says something section B's 8 of 8 does not: the repaired reference notes are complete against their fact sheets, not against everything the transcript contains, and a judge that enumerates from the transcript will find the difference.

\subsubsection{G: realism and fidelity}

Two items rated realism and one rated fidelity. The realism items are trap-blind consultations - transcripts written by a model to contain a specific documentation difficulty, with the writer not told what that difficulty was - and the question was whether each would pass as a transcript of a real UK primary care consultation, where 5 means ``you would not know''. Both were rated \textbf{5}. The fidelity item is a real PriMock57 consultation shown beside the fact sheet a model extracted from it, asking whether the sheet is a fair account of what was said, where 5 means ``nothing important missed or invented''. It was also rated \textbf{5}. This is where the author conflict bites hardest, because they are rating material generated to their own design, so it is reported as three ratings by one interested rater and nothing more. It moves no result. The separate test for trap inflation, which found no detectable inflation with an interval spanning zero, is where that question is actually answered (the companion census paper's method appendix).

{
{\small\begin{longtable}[]{@{}
  >{\raggedright\arraybackslash}p{(\linewidth - 4\tabcolsep) * \real{0.3333}}
  >{\raggedright\arraybackslash}p{(\linewidth - 4\tabcolsep) * \real{0.3333}}
  >{\raggedright\arraybackslash}p{(\linewidth - 4\tabcolsep) * \real{0.3333}}@{}}
\toprule\noalign{}
\begin{minipage}[b]{\linewidth}\raggedright
item
\end{minipage} & \begin{minipage}[b]{\linewidth}\raggedright
what was rated
\end{minipage} & \begin{minipage}[b]{\linewidth}\raggedright
rating
\end{minipage} \\
\midrule\noalign{}
\endhead
\bottomrule\noalign{}
\endlastfoot
G1 & trap-blind scenario \texttt{tb\_\hspace{0pt}psoriasis\_\hspace{0pt}new}, realism & 5 \\
G2 & trap-blind scenario \texttt{tb\_\hspace{0pt}lipids\_\hspace{0pt}results\_\hspace{0pt}review}, realism & 5 \\
G3 & extracted fact sheet for primock / day3\_consultation02 (31 facts, 17 traps), fidelity & 5 \\
\end{longtable}\addtocounter{table}{-1}}
}

\subsubsection{The three position answers, verbatim}

\textbf{Q1, on the flag rule} (``is `any single critical fact absent' the rule you would want a scribe monitor to fire on?''):

\begin{quote}
``i think the critical grade omissions are pretty important so yeah it is noteworthy if even just 1 is missing. but if 5 are missing then obviously that is also way worse''
\end{quote}

They endorse the per-fact rule as it stands and want magnitude carried beside it. The rule is unchanged; reporting the count of absent critical facts alongside each flagged note is one line over verdicts already purchased.

\textbf{Q2, on the case where a strong mention survives elsewhere in the note} (``publish it as a named open problem, or is it the case you would most want caught?''):

\begin{quote}
``yeah that's the right call''
\end{quote}

The proposal stands: report the class rather than pooling it away, as an open problem (Section 7).

\textbf{Q3, on the rubric's rule of taking the lower grade when torn} (``does that match how you would grade in practice?''):

\begin{quote}
``yeah i think that is right''
\end{quote}

Consistent with their own grading in section E, which moved four machine-critical facts down and only two up.

\subsubsection{What this validation cannot do}

\begin{enumerate}
\def\labelenumi{\arabic{enumi}.}
\tightlist
\item
  \textbf{One rater, one sitting, no re-test.} There is no within-rater reliability figure and no way to separate their standard from the instrument's. Stage E's kappa is a two-rater figure in which one of the raters wrote the rubric.
\item
  \textbf{They are an author.} Stages A, B, C and G are endorsements of positions the items had to state to be answerable. D, E and F are the sections where blinding was structural and are the only ones that should be described as independent measurements.
\item
  \textbf{Every stage holds 6 to 20 items.} Four stages read 100\% with lower bounds of 61 to 72\%: that rules out gross failure and nothing finer. The intervals are the result; the point estimates are not.
\item
  \textbf{The sampling is seeded and stratified for coverage, not adversarial.} Stage A drew one fact per consultation with no filtering for interestingness and no fact the audit stage had rejected. Stage F drew one note per consultation stratified by stratum and by how much of the fact survived, not by which judge was likely to be right.
\item
  \textbf{The pack itself cost two answers.} Eight of the ten section D items printed a window on the note rather than the whole note, and both abstentions came from that. A rebuilt D section printing the full note would be the way to test the \texttt{present\_\hspace{0pt}elsewhere} finding properly.
\item
  \textbf{The \texttt{present\_\hspace{0pt}elsewhere} bucket is unvalidated.} One answer and two abstentions across its three items, so nothing in this sitting supports or undermines the taxonomy's 25\% share for it, and it must not be cited either way.
\item
  \textbf{Stage E does not reach the grades the per-fact rule fires on.} It validates the benchmark's severity axis, not the pipeline audit stage's own grades of its own extracted facts. That is the gap E.2 closes.
\item
  \textbf{Nothing here validates the omission-detection numbers themselves.} It validates the material they are measured on (A, B, C), the panel that produced the taxonomy rates (D), the severity axis they are conditioned on (E) and the judge design comparison (F). Detection and false-alarm rates remain what Sections 5 to 7 measured.
\end{enumerate}

\subsection{The 24-item audit-stage severity sitting, 17 August 2026}

\subsubsection{Why a second sitting}

The per-fact rule of Section 7 fires on the audit stage's own severity grade of its own extracted fact. The first sitting never scored those grades: it validated the benchmark's grades, which are a separate run of the same written rubric over different material. A second blinded pass covers the audit stage's grades directly. The two sittings are separate instruments and their numbers are never pooled.

\subsubsection{The pack}

Twenty-four audit-stage-graded facts from the 47-consultation confirmation run, one item per consultation, at deterministic seed 20260817 (pack \texttt{01db4358fb60}, builder \texttt{build\_\hspace{0pt}severity\_\hspace{0pt}pack.py}). The draw is deliberately weighted towards the class the rule depends on. \textbf{10 rule-fired facts} - graded critical by the audit stage \emph{and} judged absent - were drawn from the 31 such facts in the run, including the \texttt{aci\textbar{}D2N199} ``patient is aged 74'' clean-note false alarm. Beside them sit 5 random machine-critical, 5 random machine-supporting and 4 random machine-peripheral facts. Each item printed the full transcript. The machine grade was structurally blinded, and a byte-identity check over the stripped item cards verified that nothing on screen varied with the grade. Sat 2026-08-17, 16:06 to 16:14 UTC, 7 minutes 49 seconds, all 24 answered, zero abstentions, zero comments.

\subsubsection{The blind pass}

{
{\small\begin{longtable}[]{@{}
  >{\raggedright\arraybackslash}p{(\linewidth - 6\tabcolsep) * \real{0.2500}}
  >{\raggedright\arraybackslash}p{(\linewidth - 6\tabcolsep) * \real{0.2500}}
  >{\raggedright\arraybackslash}p{(\linewidth - 6\tabcolsep) * \real{0.2500}}
  >{\raggedright\arraybackslash}p{(\linewidth - 6\tabcolsep) * \real{0.2500}}@{}}
\toprule\noalign{}
\begin{minipage}[b]{\linewidth}\raggedright
stratum
\end{minipage} & \begin{minipage}[b]{\linewidth}\raggedright
n
\end{minipage} & \begin{minipage}[b]{\linewidth}\raggedright
exact
\end{minipage} & \begin{minipage}[b]{\linewidth}\raggedright
weighted (adjacent = half)
\end{minipage} \\
\midrule\noalign{}
\endhead
\bottomrule\noalign{}
\endlastfoot
\textbf{rule-fired (all machine-critical)} & 10 & \textbf{1/10 = 10\% {[}1.8, 40.4{]}} & 50\% \\
random machine-critical & 5 & 2/5 = 40\% & 70\% \\
random machine-supporting & 5 & 3/5 = 60\% & 80\% \\
random machine-peripheral & 4 & 3/4 = 75\% & 88\% \\
\textbf{machine-critical pooled} & 15 & \textbf{3/15 = 20\% {[}7.0, 45.2{]}} & 57\% \\
all 24 (stratified, critical-heavy) & 24 & 9/24 = 38\% {[}21.2, 57.3{]} & 67\% \\
\end{longtable}\addtocounter{table}{-1}}
}

Confusion matrix, rows the audit stage's grade, columns the clinician's blind grade:

{
{\small\begin{longtable}[]{@{}llll@{}}
\toprule\noalign{}
audit stage ~clinician & critical & supporting & peripheral \\
\midrule\noalign{}
\endhead
\bottomrule\noalign{}
\endlastfoot
\textbf{critical} (15) & 3 & 11 & 1 \\
\textbf{supporting} (5) & 0 & 3 & 2 \\
\textbf{peripheral} (4) & 0 & 1 & 3 \\
\end{longtable}\addtocounter{table}{-1}}
}

Their mix is 3 critical / 15 supporting / 6 peripheral against the audit stage's 15 / 5 / 4. Linear-weighted kappa is 0.28 and is \textbf{not comparable to the first sitting's 0.63}: this sample is deliberately critical-heavy, which biases kappa, so the per-stratum rows are the result and the kappa is not. There is one two-grade disagreement, S02 (``he had been coughing a lot for the last two days'', machine critical against clinician peripheral), where the first sitting had none. The two near-duplicate distal pulse-examination facts, S14 and S16 from different consultations, both drew ``supporting'', which is internally consistent grading.

The random strata carry 4 and 5 items each and are quoted as counts for that reason; only the three rows with intervals above carry enough weight to read as rates.

\subsubsection{The unblinded reconciliation}

The same day, the 15 blind disagreements were re-adjudicated unblinded, with the machine grade and the audit stage's recorded reasoning shown per item. \textbf{Five revised to critical} - S02 (the cough sitting beside calf findings, a pulmonary embolism picture), S03 (avoiding non-steroidal anti-inflammatories on renal risk), S07 (the blood-pressure safety-net), S11 (nil by mouth before possible surgery) and S23 (work of breathing in a febrile child) - all cases where the critical grade rests on consultation context that a fast blind pass had not weighed. \textbf{Seven stood} on the first reconciliation pass (S13 a bare age, S14 and S16 the normal pulse examinations, S20 family screening, and S08, S10, S06), and \textbf{three more stood on a follow-up confirmation} (S12, S17 and S22, all random machine-critical items the clinician kept at supporting).

On the pulse-examination pair they gave the reasoning verbatim:

\begin{quote}
``if the pulses were absent i think it would be critical, but given they are present it's slightly more on supporting, but very borderline''
\end{quote}

Reconciled: \textbf{rule-fired criticals confirmed 6/10 {[}31.3, 83.2{]}}, and \textbf{rule-fired facts graded clinically material - critical or supporting - 10/10}. Machine-critical pooled 8/15. Overall exact agreement 14/24 = 58\%, weighted 79\%. \textbf{Every residual disagreement is a single grade}, and the audit stage is the more severe grader in 9 of the 10 residuals; the single exception is S06 (a machine \texttt{peripheral} the clinician put at supporting).

All 24 items - stratum, the fact graded, the audit stage's grade, the blind answer, the reconciled answer and its status - are in the released validation artefact. Fact wording there is the audit stage's own, including the US trade names the ACI-Bench consultations use (Motrin is ibuprofen, Tylenol paracetamol).

\subsubsection{What the two passes say}

\textbf{Blind, the audit stage's critical class looked substantially generous}: 1 of 10 rule-fired criticals confirmed. \textbf{Reconciled against the audit stage's own reasoning, half that gap was consultation context the fast blind pass had not weighed}, and the final position is 6 of 10 rule-fired criticals confirmed, with the residual generosity being one grade on routine content - a bare age, normal examination findings, family screening. Both passes are reported, blind first.

\textbf{The flags are clinically material.} All 10 of 10 rule-fired facts are critical or supporting on reconciliation, so a flag names something a clinician would want in the note even where the ``critical'' label is one grade high.

\textbf{Detection and false-alarm rates are unaffected.} They are measured against constructed absence, not against severity truth. What changes is what a flag means: a material missing fact that the audit stage grades critical, with that grade sitting about one grade above this clinician's on routine content. Aligning the audit stage's critical threshold with clinician grades and re-scoring the already-purchased per-fact verdicts is a cheap lever, and it is named as such in Section 9.

\textbf{The direction is the rater's as much as the instrument's.} The same written rubric has since been graded blind by an independent clinician - not an author of this study and with no involvement in it - on a fresh sample of findings in the companion census: 9 of 12 grades exact and 3 one grade apart. Across the two census sittings 25 of 32 grades are exact and no disagreement anywhere exceeds a single grade, so the rubric reproduces to within a grade under two clinicians who graded disjoint material. What the pair does not establish is a direction, because the two clinicians lean opposite ways against it (Section 3). The one-grade lean recorded above is therefore read as a property of this rater as much as of the audit stage, and no claim in this paper rests on the rubric leaning one way.

\subsubsection{Caveats}

One rater, an author, one 8-minute blind sitting plus a same-day unblinded reconciliation, and no re-test. The rows that carry the reading hold 10 and 15 items, so the intervals are wide and the intervals are the result. The reconciliation is unblinded by construction and is reported beside the blind pass, never instead of it. And the ``patient is aged 74'' fact (S13 here, item F6 in the first sitting) drew supporting in both passes, so the first sitting's observation stands - the note really does lack a transcript fact - while the flag remains a false alarm under the rule's critical-only trigger, the supporting grade now confirmed twice.

\textbf{Artefacts} (paths in the released data repository's \texttt{validation/\hspace{0pt}} directory unless noted). First sitting: \texttt{sitting\_\hspace{0pt}results.json} (the per-item join, per-stage rates with Wilson intervals, the severity confusion matrix and kappa, the adjudication table, the position answers verbatim), the raw answer export \texttt{sitting\_\hspace{0pt}answers.json}, and the deterministic builder \texttt{build\_\hspace{0pt}sitting\_\hspace{0pt}pack.py} in the code repository - the pack itself prints withheld note excerpts and is not released. Second sitting: \texttt{build\_\hspace{0pt}severity\_\hspace{0pt}pack.py} (code repository), \texttt{severity\_\hspace{0pt}answers.json}, \texttt{severity\_\hspace{0pt}pack\_\hspace{0pt}keys.json} and \texttt{severity\_\hspace{0pt}validation\_\hspace{0pt}results.json}.

\section{Release index}

Section 9 states what we release and what we withhold. This appendix names the release's two homes and its licences, sets out the withholding policy, and carries the compressed datasheet - composition, verification standards and known limitations. The field-level schemas, the scoring protocol and the usage guide live in the bundle's dataset card, which ships as the data repository's \texttt{README.md}, and nothing in the card claims more than Section 9 does.

The benchmark is released as \textbf{OmissionBench}, in two parts. The data - everything the datasheet in F.3 describes - is at \texttt{huggingface.co/\hspace{0pt}datasets/\hspace{0pt}ComposoAI/\hspace{0pt}OmissionBench} under CC BY 4.0, and carries a DataCite DOI minted at publication against the released revision. The harness is at \texttt{github.com/\hspace{0pt}composo-ai/\hspace{0pt}omission-bench} under the MIT licence, archived to Zenodo at its tagged releases under DOI \texttt{10.5281/\hspace{0pt}zenodo.22160954}, so the code has a citable snapshot independent of the hosting account. Paths below are relative to the data repository's root unless the row names the code repository.

\subsection{What the bundle holds}

{
{\small\begin{longtable}[]{@{}
  >{\raggedright\arraybackslash}p{(\linewidth - 2\tabcolsep) * \real{0.5000}}
  >{\raggedright\arraybackslash}p{(\linewidth - 2\tabcolsep) * \real{0.5000}}@{}}
\toprule\noalign{}
\begin{minipage}[b]{\linewidth}\raggedright
path
\end{minipage} & \begin{minipage}[b]{\linewidth}\raggedright
contents
\end{minipage} \\
\midrule\noalign{}
\endhead
\bottomrule\noalign{}
\endlastfoot
\texttt{pairs/\hspace{0pt}dataset\_\hspace{0pt}v2.json} & the 500 graded pairs (495 evaluation), each with its clean note, its errored twin, the fact that was injected or removed, its severity and trace grades, and its build-cohort tag \\
\texttt{pairs/\hspace{0pt}fact\_\hspace{0pt}sites.json} & the fact-site map: the sites where each fact is stated in each clean note \\
\texttt{pairs/\hspace{0pt}factorial\_\hspace{0pt}severity.json} & the per-fact severity grades, with both graders' verdicts behind each consensus \\
\texttt{transcripts/\hspace{0pt}} & 141 transcripts, one per consultation, by stratum 57 PriMock57, 45 ACI-Bench, 30 authored and 9 trap-blind: authored and trap-blind scenarios (ours), PriMock57 and ACI-Bench in derived form (upstream CC BY 4.0, attribution in F.4) \\
\texttt{fact\_\hspace{0pt}sheets/\hspace{0pt}} & the per-consultation fact sheets in full and core views, including the authored answer keys \\
\texttt{judgements/\hspace{0pt}findings/\hspace{0pt}} & the census's 618 verified findings, products as letters, with the transcript-side evidence quoted \\
\texttt{judgements/\hspace{0pt}judges/\hspace{0pt}} & every benchmark judge run - the ablation grid, the reference judges, the RAGAS-style recipe, the pipeline tiers, the second-family runs, the reasoning-budget re-runs (\texttt{w2-power/\hspace{0pt}}) and the optimisation campaigns - with raw completions and one manifest per run \\
\texttt{judgements/\hspace{0pt}census/\hspace{0pt}} & the census pipeline's own run manifests, the four-standards panel's prompts, read-outs and lenient-arm record store, and the deduplication groupings beside the findings; its per-call logs judge the withheld notes and are withheld with them \\
\texttt{notes/\hspace{0pt}structure/\hspace{0pt}} & per-note count metadata for the census notes; no note text \\
\texttt{validation/\hspace{0pt}} & the four clinician sittings' per-item records: the judge-adjudication and severity sittings, the census precision sitting (\texttt{precision-sitting/\hspace{0pt}}) and the second clinician's sitting (\texttt{second-clinician-sitting/\hspace{0pt}}) \\
code repository & the MIT harness: capture scripts, every prompt including superseded versions (the deployed faithfulness judge's three wordings under \texttt{w2\_\hspace{0pt}prompts/\hspace{0pt}} with SHA-256 hashes; the as-shipped text is also printed in Appendix H), optimisation lineages, judge configurations, \texttt{models.lock.json}, provenance records, and the text-to-speech recipe that renders the audio \\
\end{longtable}\addtocounter{table}{-1}}
}

\textbf{Anonymisation of the harness itself.} The study was not run anonymously, so some of its file names and configuration keys carry product names. The packaging step rewrites them to the letters and then fails the build on any case-insensitive product-name match anywhere in the bundle. A released capture script therefore names its driving mode - a documented API, or audio played to the product in real time - and never the product; the letter-to-product mapping stays in the private repository and is not released in any form.

\subsection{What we withhold}

What we withhold is the note text the three products wrote. The reason is contractual rather than clinical: the corpus contains no patient data at all (Section 9), and it is the products' terms of service, which differ from one another, that decide whether their output can be republished. We treat all three the same way rather than release asymmetrically, partly because a release carrying one product's notes and not another's would make products identifiable by their absence.

Three things keep the withheld half honest. Released findings carry the transcript-side evidence for every claim and never the note's own words, and the packaging step fails the build on any released content that reproduces a run of withheld note text, on any schema field carrying a note body, and on any product name anywhere in the bundle - the checks themselves are described in the released \texttt{VERIFICATION.md}. A replicator with their own product accounts can regenerate equivalent notes end to end with the released harness. And researchers who need more than the release carries can contact the authors.

\subsection{The benchmark datasheet, compressed}

The full datasheet is the bundle's dataset card; this section carries what a reader of the paper needs - composition, verification and limitations. Intended use is evaluating LLM judges and ambient scribe systems; it is explicitly not for clinical use of any kind or as training data for clinical systems, and nothing in it is a real patient encounter or a clinical record.

\subsubsection{Composition and construction}

Composition is Table~\ref{tab:composition-a}'s job, in A.7, and construction is Appendix A's; this datasheet does not repeat them. In one sentence each: the corpus is 145 consultations over four strata (30 authored, 10 trap-blind with 9 kept, 57 PriMock57 with 53 kept, 48 ACI-Bench with 45 kept), of which 134 carry a verified clean note - 112 evaluation, 22 in a separate judge-optimisation split - and the released set is 500 pairs, 495 in the evaluation set - 293 omissions (150 complete, 86 fragment trace, 57 restatement trace) and 202 commissions - plus the 112 clean twin notes they are scored against. Construction runs transcript-first through nine steps - deterministic subsample (seed 20260728, the 67 previously-used ACI-Bench encounters excluded), blind extraction, three-critic audit, consolidation, reference-note repair, site mapping, single-error injection, the optimisation-split carve, and cross-family verification with failures excluded rather than patched - shown with its counts in Figure~\ref{fig:construction}. The repaired reference note ships inside each pair record as the clean note; the originals belong to the upstream corpora's own releases; the 27 blind re-extractions of authored transcripts are an instrument-comparison artefact from which no pair is built.

\subsubsection{Verification standards}

Nothing in the dataset is described as clean by construction. Each layer has an instrument, a measured rate on a stated denominator, and a published failure list.

{
{\small\begin{longtable}[]{@{}
  >{\raggedright\arraybackslash}p{(\linewidth - 4\tabcolsep) * \real{0.3333}}
  >{\raggedright\arraybackslash}p{(\linewidth - 4\tabcolsep) * \real{0.3333}}
  >{\raggedright\arraybackslash}p{(\linewidth - 4\tabcolsep) * \real{0.3333}}@{}}
\toprule\noalign{}
\begin{minipage}[b]{\linewidth}\raggedright
layer
\end{minipage} & \begin{minipage}[b]{\linewidth}\raggedright
standard
\end{minipage} & \begin{minipage}[b]{\linewidth}\raggedright
measured
\end{minipage} \\
\midrule\noalign{}
\endhead
\bottomrule\noalign{}
\endlastfoot
fact sheets & three critics, one revision cycle, drop if material issues remain & sheets dropped: PriMock57 4 of 57 (7.0\%) after three disclosed instrument corrections, ACI-Bench 3 of 48 (6\%), trap-blind 1 of 10, authored re-extraction 3 of 30 \\
fact-sheet fidelity & blind extraction against the authored sheet on the same 30 transcripts & 646 of 650 authored facts recovered, 99.4\% {[}98.8, 99.9{]}, per-consultation minimum 93.8\% \\
reference notes & audited against transcript and core sheet, repaired, re-verified & 53 of 53 PriMock57 and 45 of 45 ACI-Bench notes carry at least one material discrepancy with their own transcript, at means of 10.7 and 7.1 per note; 17 of the 98 repaired notes still fail an ultra-strict residual check and are logged \\
clean notes & four-section checklist audit by a different model family, key in hand & first-pass defect rate by stratum: authored 11 of 30 (36.7\%), trap-blind 3 of 9 (33.3\%), PriMock57 49 of 53 (92.5\%), ACI-Bench 36 of 45 (80.0\%); 88 notes repaired, 3 excluded \\
pairs, earlier build & deterministic single-edit check, then a semantic edit check at three seeds with a per-field majority & first-pass failure 182 of the 402 pairs that reached the check (45.3\%) {[}40.5, 50.2{]}, out of 411 constructed; 78 of the 411 excluded \\
pairs, graded-omission build & cross-family panel of two model families, third call on any field they split on, per-field majority & verified 71 of 126 complete-removal attempts (56.3\%) and 114 of 151 partial removals (75.5\%); 92 pairs dropped, itemised with their failing checks \\
severity grades & written rubric, two independent cross-family graders, conservative tie-break & Cohen's kappa 0.662 {[}0.59, 0.73{]} over the 683 traps regraded under the rubric, against 0.177 {[}0.04, 0.31{]} over 171 matched traps graded without it, intervals non-overlapping; 92 of the 683 (13.5\%) resolved to the lower grade \\
\end{longtable}\addtocounter{table}{-1}}
}

Two disclosures belong with these numbers. The audit rates are auditor-majority rates with no human adjudication in the loop, so a defect rate is an upper bound on a human-confirmed one; the empirical check on auditor strictness is that 15.3\% of one-seed screen flags die at three-seed majority confirmation (57 of 373), rising to 29.6\% of the 98 flags in the open-ended section. And audit reproducibility is 0.794 over the escalated checklist sections and 0.725 including the open-ended one, both conditional on the screen having flagged something at all (222 of 1,024 section-audits escalated), and therefore a lower bound over disputed sections rather than a whole-corpus figure.

\subsubsection{Known biases, gaps and exclusions}

\textbf{Population and setting.} English-language UK primary care and US ambulatory care only. Two of four strata are consultations we wrote; the largest public stratum is UK primary care with actor patients. No secondary care, no specialty outside general practice, and no note in the corpus carries an electronic-record structured field, so the benchmark cannot speak to structured-field omissions at all.

\textbf{No real patients anywhere.} Actors, simulation and fiction throughout. That is an ethical strength and an ecological limitation at once: acted and simulated consultations are cleaner and more purposeful than real ones.

\textbf{Audio provenance is mixed.} One stratum is genuine recordings, three are synthesised speech, and the recordings predate the products by several years. All capture ran at native speed on unmodified audio, after a speed-up test was rejected when silence-trimming produced a critical medication substitution. The audio itself is regenerable from the released transcripts with the harness's text-to-speech recipe and pins.

\textbf{Selection asymmetry between strata.} First-pass pair failure was 18.9\% on the authored stratum against 52.9\% on PriMock57 and 52.3\% on ACI-Bench, because the authored stratum is preserved verbatim and its failing pairs are excluded rather than regenerated, while public-stratum notes were repaired and pairs regenerated up to three times. A surviving public-stratum pair has passed a harsher filter, so judge and scribe scores should not be compared across strata.

\textbf{Two construction cohorts are pooled, and recorded per row.} The 202 commission controls and 79 complete omissions carried over were built by model rewrite rather than by span deletion off the site map, and the 79 carry severity grades from an older single-arm grading; the 34 relabelled partial seeds vary in residual count from 1 to 9 because they predate the site map. Every pair records which cohort it came from, so a user can restrict to the internally uniform column - noting that all commission controls sit in the carried-over cohort, so the uniform column has no commission arm.

\textbf{Residual count is a property of the map, not of the note.} Construction left exactly one mapped residual site on all 151 partial pairs, but the verification panel's own reading found more than one surviving mention in 71 of them. Both counts are in each record, so an analysis can control for it; the released trace labels use the map's count.

\textbf{Thin and unpopulated cells.} The restatement-trace peripheral cell holds 2 pairs and should be treated as unpopulated, and the fragment-trace peripheral cell is thin at 8. A restatement trace needs a fact whose second mention is explicit or a full paraphrase, and peripheral facts are both the rarest in the corpus and the least redundant.

\textbf{Excluded content, with counts.}

{
{\small\begin{longtable}[]{@{}
  >{\raggedright\arraybackslash}p{(\linewidth - 4\tabcolsep) * \real{0.3333}}
  >{\raggedright\arraybackslash}p{(\linewidth - 4\tabcolsep) * \real{0.3333}}
  >{\raggedright\arraybackslash}p{(\linewidth - 4\tabcolsep) * \real{0.3333}}@{}}
\toprule\noalign{}
\begin{minipage}[b]{\linewidth}\raggedright
what
\end{minipage} & \begin{minipage}[b]{\linewidth}\raggedright
count
\end{minipage} & \begin{minipage}[b]{\linewidth}\raggedright
why
\end{minipage} \\
\midrule\noalign{}
\endhead
\bottomrule\noalign{}
\endlastfoot
consultations dropped at the fact-sheet gate & 8 of 145 & material issues remained after one revision cycle; Figure~\ref{fig:construction}'s 11 of 145 counts sheets, adding the 3 authored consultations whose blind re-extraction dropped while their own authored sheets stand \\
clean notes excluded & 3 of 137 & never converged after three repair rounds \\
pairs excluded at the earlier build's answer-key audit & 78 of 411 & 27 omit pairs failed three regenerations and the deletion fallback, 20 add and 5 change failed three regenerations, 17 authored pairs failed the semantic majority and are never rewritten, 9 were orphaned by the excluded notes \\
graded-omission pairs dropped for failing verification & 92 of 277 & itemised with their failing checks in the released dataset \\
encounters tagged as overlapping an external benchmark & 14 of 48 & retained in the benchmark, excluded from any analysis that also uses that benchmark \\
\end{longtable}\addtocounter{table}{-1}}
}

\textbf{Everything is model-produced except the transcripts.} Fact sheets, clean notes, injections, repairs, site maps and audits are all model outputs, checked by models from a different family. Human clinical review is by an author-clinician and is disclosed as such; its scope and its limits are in Appendix E.

Anonymisation and safety, licences, versioning and citation are F.2's and the head of this appendix's job, restated in full in the dataset card: the three systems are Scribe A, B and C everywhere, the released set carries \texttt{dataset\_\hspace{0pt}version:\ factorial-v1} (built 2026-08-12, superseding the earlier 281-pair freeze whose digests identify the carried-over cohorts' source), the holdout from the prompt optimiser is asserted structurally at build time, and a number computed over the benchmark should be quoted with the dataset version, because a later revision mints its own DOI and the two are not interchangeable.

\subsection{Per-corpus attribution}

Both public corpora are published under CC BY 4.0, and both are redistributed here in derived form. The transcripts are normalised, and for ACI-Bench subsampled. The clean notes for those two strata are the corpora's own reference notes audited against their transcripts and repaired, so every pair built on them is an adaptation of upstream material, not only of ours. The fact sheets derived from those strata are adaptations too.

CC BY 4.0 requires four things of a derived release: name the creator, carry the licence with a link to it, state that changes were made and what they were, and impose no further restriction on downstream users. The release carries one fixed attribution string per corpus - naming the creator, the licence, the source record, and the changes made (transcripts normalised; for ACI-Bench a 48-encounter subsample drawn at seed 20260728; the clinician reference notes audited against their own transcripts and repaired; single-error variants generated from the repaired notes) - reproduced verbatim in the bundle's \texttt{ATTRIBUTION.md} beside the licence file, in the dataset card, and in the \texttt{\_\hspace{0pt}attribution} header of every derived JSON artefact. For ACI-Bench the CC BY 4.0 grant is carried on the corpus's figshare record, which the string cites; for PriMock57 it is the repository's own licence file. Everything else in the bundle is our own work released under CC BY 4.0, so a downstream user carries one licence across the whole bundle and two attribution lines.

Both upstream licences were verified against their published sources at packaging time, and CC BY 4.0 expressly permits redistribution of adapted material with attribution, which is what this bundle is.

\section{The second judge family, in full}

This appendix carries every number behind the cross-family sentences in Section 5, the transfer paragraph in Section 7, and the first limitation in Section 8: what was run on a second judge model family, on which sets of pairs, and what came back.

Its scope should be read before its numbers. This is one additional judge model family, and on it we ran two of the eight judge designs as monolithic judges plus one replicate of each pipeline tier with only the per-fact checker swapped. It is not a second ablation and it does not license family-versus-family comparisons. The completeness-scored single-sample design carries three replicates here; the faithfulness-only yes/no design carries one, and so does each pipeline tier. The replicates establish that each family's own figure is stable from run to run, not that the difference between the families is real: no test of that difference was run, and none of the deltas below is a tested effect. The eight-sample completeness-scored design was never run on this family, so two designs of the eight are the whole monolithic picture.

\subsection{What was run}

The second-family judge is \texttt{google/\hspace{0pt}gemini-3.1-pro-preview}, served through OpenRouter on the \texttt{google-vertex} route, at temperature 1.0, \texttt{max\_\hspace{0pt}tokens} 1024, and \texttt{reasoning\_\hspace{0pt}effort} \texttt{minimal}. The endpoint refuses to run with reasoning switched off, which is the first of this appendix's two confounds (Section G.4). The primary family is the pinned GPT-family judge of Section 4, at the settings it ran everywhere else. Appendix D carries the model pins, and names the release homes of the prompt hashes, run manifests and compute ledger for both runs here.

The monolithic replication used the judge prompts byte for byte as the eight designs ran them, on the same set of pairs: 495 evaluation pairs and 112 verified-clean twins, giving 293 omission records, 202 commission records (additions and alterations) and 112 clean twins per design per replicate. Replicate seeds are the same three the eight designs used (11, 22, 33). Every primary-family row printed beside a second-family row below is recomputed from the original judgement store by the same scoring functions and checked against the published figures before anything here was written, agreeing within 0.0015 on all four paired values.

\subsection{The asymmetry and the restatement-trace collapse both reproduce}

Paired discrimination is the share of pairs in which the judge scores the flawed note strictly below its own verified-clean twin, plus half of any ties, so 0.500 is a coin flip. Both error columns are the same measure on the same notes; only the class of error differs.

{
{\small\begin{longtable}[]{@{}
  >{\raggedright\arraybackslash}p{(\linewidth - 10\tabcolsep) * \real{0.1667}}
  >{\raggedright\arraybackslash}p{(\linewidth - 10\tabcolsep) * \real{0.1667}}
  >{\raggedright\arraybackslash}p{(\linewidth - 10\tabcolsep) * \real{0.1667}}
  >{\raggedright\arraybackslash}p{(\linewidth - 10\tabcolsep) * \real{0.1667}}
  >{\raggedright\arraybackslash}p{(\linewidth - 10\tabcolsep) * \real{0.1667}}
  >{\raggedright\arraybackslash}p{(\linewidth - 10\tabcolsep) * \real{0.1667}}@{}}
\toprule\noalign{}
\begin{minipage}[b]{\linewidth}\raggedright
design
\end{minipage} & \begin{minipage}[b]{\linewidth}\raggedright
judge family
\end{minipage} & \begin{minipage}[b]{\linewidth}\raggedright
replicates
\end{minipage} & \begin{minipage}[b]{\linewidth}\raggedright
paired commissions
\end{minipage} & \begin{minipage}[b]{\linewidth}\raggedright
paired omissions
\end{minipage} & \begin{minipage}[b]{\linewidth}\raggedright
commissions minus omissions
\end{minipage} \\
\midrule\noalign{}
\endhead
\bottomrule\noalign{}
\endlastfoot
faithfulness only, yes/no, asked once (F-bin-k1) & primary & 3 & 0.875 & 0.500 & +0.375 \\
faithfulness only, yes/no, asked once (F-bin-k1) & \textbf{second} & 1 & \textbf{0.951} & \textbf{0.551} & \textbf{+0.399} \\
faithfulness and completeness, 0-to-10 score, asked once (FC-score-k1) & primary & 3 & 0.895 & 0.585 & +0.310 \\
faithfulness and completeness, 0-to-10 score, asked once (FC-score-k1) & \textbf{second} & 3 & \textbf{0.961} & \textbf{0.683} & \textbf{+0.278} \\
\end{longtable}\addtocounter{table}{-1}}
}

Multi-replicate rows are the mean of the per-replicate values. The direction, the rough size and the ordering of the two designs all reproduce: on this family, content that was added or altered is caught at 0.951 and 0.961 while missing content sits at 0.551 and 0.683, a gap of 0.40 and 0.28 on a measure whose chance level is 0.500.

Every paired and detection figure in the table below sits above its primary-family counterpart, at a lower false-alarm rate, and the per-replicate rows show that the completeness-scored design's own figure is stable rather than a single lucky draw. The exception among the monolithic rows is the restatement-trace case, G.2's residual table; G.3's pipeline transfer has several figures that move the other way. Each row is a complete independent pass over the full set of pairs; nothing is pooled inside a row. Detection is the flag rate on the 293 omission-errored notes under the score threshold fixed before the run (a score of 7 or below), and the false-alarm rate is the same design's flag rate on the 112 clean twins.

{
{\small\begin{longtable}[]{@{}
  >{\raggedright\arraybackslash}p{(\linewidth - 12\tabcolsep) * \real{0.1429}}
  >{\raggedright\arraybackslash}p{(\linewidth - 12\tabcolsep) * \real{0.1429}}
  >{\raggedright\arraybackslash}p{(\linewidth - 12\tabcolsep) * \real{0.1429}}
  >{\raggedright\arraybackslash}p{(\linewidth - 12\tabcolsep) * \real{0.1429}}
  >{\raggedright\arraybackslash}p{(\linewidth - 12\tabcolsep) * \real{0.1429}}
  >{\raggedright\arraybackslash}p{(\linewidth - 12\tabcolsep) * \real{0.1429}}
  >{\raggedright\arraybackslash}p{(\linewidth - 12\tabcolsep) * \real{0.1429}}@{}}
\toprule\noalign{}
\begin{minipage}[b]{\linewidth}\raggedright
replicate
\end{minipage} & \begin{minipage}[b]{\linewidth}\raggedright
seed
\end{minipage} & \begin{minipage}[b]{\linewidth}\raggedright
paired commissions
\end{minipage} & \begin{minipage}[b]{\linewidth}\raggedright
paired omissions
\end{minipage} & \begin{minipage}[b]{\linewidth}\raggedright
omission detection
\end{minipage} & \begin{minipage}[b]{\linewidth}\raggedright
false alarms (clean twins)
\end{minipage} & \begin{minipage}[b]{\linewidth}\raggedright
detection-vs-false-alarm z
\end{minipage} \\
\midrule\noalign{}
\endhead
\bottomrule\noalign{}
\endlastfoot
r1 & 11 & 0.960 & 0.700 & 20.1\% (59/293) & 1.8\% (2/112) & 4.62 \\
r2 & 22 & 0.965 & 0.675 & 19.1\% (56/293) & 0.9\% (1/112) & 4.72 \\
r3 & 33 & 0.958 & 0.674 & 20.5\% (60/293) & 3.6\% (4/112) & 4.17 \\
\textbf{mean / pooled} & & \textbf{0.961} & \textbf{0.683} & \textbf{19.9\% (175/879)} & \textbf{2.1\% (7/336)} & \textbf{7.79} \\
\end{longtable}\addtocounter{table}{-1}}
}

Across the three replicates: paired omissions 0.674 to 0.700, paired commissions 0.958 to 0.965, detection 19.1 to 20.5\% against a false-alarm rate of 0.9 to 3.6\%, and a detection-versus-false-alarm separation that is significant in every one, z 4.17 to 4.72. On this family the completeness-scored single-sample judge clears its own noise on single notes in all three replicates, which no design in the primary family did.

Conditioning on how much of the fact survives reproduces as well, including the part that matters clinically. A restatement trace means the fact's main statement is gone while an explicit or closely paraphrased mention survives elsewhere in the note.

{
{\small\begin{longtable}[]{@{}llll@{}}
\toprule\noalign{}
replicate & complete removal & fragment trace & restatement trace \\
\midrule\noalign{}
\endhead
\bottomrule\noalign{}
\endlastfoot
r1 & 0.792 & 0.657 & 0.518 \\
r2 & 0.750 & 0.651 & 0.509 \\
r3 & 0.770 & 0.622 & 0.500 \\
\textbf{mean} & \textbf{0.771} & \textbf{0.643} & \textbf{0.509} \\
\end{longtable}\addtocounter{table}{-1}}
}

The restatement-trace case sits at chance in every replicate, 0.500 to 0.518, exactly as it does on the primary family, where the best of the eight designs reads 0.526. The severity breakdown runs in the same order on this family: critical 0.754, supporting 0.614, peripheral 0.590 for the completeness-scored design over three replicates, against the primary family's 0.683, 0.586 and 0.568 on its best design. The faithfulness-only design, on its single replicate, reads 0.567, 0.558 and 0.500 by residual and 0.570, 0.539 and 0.513 by severity.

\subsubsection{Absolute detection, and how far the pooled separation can be trusted}

Detection means nothing except beside the same design's false-alarm rate at the same threshold, so both are printed together, with the flag rule that produced them.

{
{\small\begin{longtable}[]{@{}
  >{\raggedright\arraybackslash}p{(\linewidth - 12\tabcolsep) * \real{0.1429}}
  >{\raggedright\arraybackslash}p{(\linewidth - 12\tabcolsep) * \real{0.1429}}
  >{\raggedright\arraybackslash}p{(\linewidth - 12\tabcolsep) * \real{0.1429}}
  >{\raggedright\arraybackslash}p{(\linewidth - 12\tabcolsep) * \real{0.1429}}
  >{\raggedright\arraybackslash}p{(\linewidth - 12\tabcolsep) * \real{0.1429}}
  >{\raggedright\arraybackslash}p{(\linewidth - 12\tabcolsep) * \real{0.1429}}
  >{\raggedright\arraybackslash}p{(\linewidth - 12\tabcolsep) * \real{0.1429}}@{}}
\toprule\noalign{}
\begin{minipage}[b]{\linewidth}\raggedright
design
\end{minipage} & \begin{minipage}[b]{\linewidth}\raggedright
judge family
\end{minipage} & \begin{minipage}[b]{\linewidth}\raggedright
flag rule
\end{minipage} & \begin{minipage}[b]{\linewidth}\raggedright
omission detection
\end{minipage} & \begin{minipage}[b]{\linewidth}\raggedright
commission detection
\end{minipage} & \begin{minipage}[b]{\linewidth}\raggedright
false alarms (clean)
\end{minipage} & \begin{minipage}[b]{\linewidth}\raggedright
detection-vs-false-alarm z
\end{minipage} \\
\midrule\noalign{}
\endhead
\bottomrule\noalign{}
\endlastfoot
F-bin-k1 & primary, 3 reps pooled & FAIL verdict & 9.6\% (84/879) & 84.0\% (509/606) & 8.9\% (30/336) & 0.34 \\
F-bin-k1 & \textbf{second, 1 rep} & FAIL verdict & \textbf{13.0\% (38/293)} & 92.6\% (187/202) & \textbf{2.7\% (3/112)} & \textbf{3.07} \\
FC-score-k1 & primary, 3 reps pooled & score \textless= 7 & 4.7\% (41/879) & 58.7\% (356/606) & 5.7\% (19/336) & -0.71 \\
FC-score-k1 & \textbf{second, 3 reps pooled} & score \textless= 7 & \textbf{19.9\% (175/879)} & 90.9\% (551/606) & \textbf{2.1\% (7/336)} & \textbf{7.79} \\
\end{longtable}\addtocounter{table}{-1}}
}

The pooled z values in that table, on both families, count three passes over the same notes as three independent observations, which inflates them. The per-replicate z values above (4.17 to 4.72) are the honest figures for the second family, and the same caveat applies wherever a pooled row is quoted for the primary family. Nothing in the paper rests on a pooled z.

One further operating point needs its mechanism stated with it. Swept to the most sensitive threshold whose false-alarm rate stays within 10\%, the second family's completeness-scored design reads 44.9\% detection (394/878) at 9.6\% false alarms (32/335), against 4.7\% (41/879) at 5.7\% (19/336) for the same design on the primary family. That 44.9\% is bought at a threshold of ``anything short of a perfect 10 is a flag'', and it works only because this judge scores clean notes at exactly 10. It is an operating point on one corpus with no evidence that it survives off-distribution, and it is why the paper's cross-family claim is about the gap between commission and omission detection rather than about absolute levels. The swept denominators are 878 and 335 rather than 879 and 336 because the two records that failed to parse (below) drop out of the swept curve, one omission record and one clean twin.

\subsubsection{Parse failures}

Two records failed to parse in 2,428, both on the same consultation, \texttt{aci\textbar{}D2N086}: none on the faithfulness-only design, two on the completeness-scored one. Twenty-two samples hit the single retry fixed before the run and twenty of those recovered. Unparseable output is recorded as a parse failure and never halts a run.

In replicate 1 the failure was the clean twin, so every pair of that consultation loses its reference and drops out of the paired measure for that replicate. In replicate 2 it was an errored note, so that one pair drops. Absolute rates are unaffected, because a parse failure counts as not-flagged against the full denominator. Paired denominators after those drops: the faithfulness-only design 293 omission and 202 commission records; the completeness-scored design 290 and 200 in replicate 1, 292 and 202 in replicate 2, and 293 and 202 in replicate 3, against 293 and 202 with nothing dropped.

\subsection{The pipeline transfer: only the check stage changed}

The second run swapped the per-fact checker of Section 7's pipeline to the same second-family judge and changed nothing else. Extraction and the critic audit were read from the cache of the original run for all 47 consultations, so the fact lists and the severity grades are identical character for character across the two checkers; the quote-verified second look also stayed on the auditor model. Extraction runs on a third model family and the audit on the primary family, so in both configurations the enumerated facts are not the checking judge's own. The check stage ran at \texttt{reasoning\_\hspace{0pt}effort} \texttt{minimal}, temperature 1.0, one replicate, seed 11.

The set of pairs is Section 7's confirmation set exactly: 151 held-out pairs over 47 consultations (10 additions, 10 alterations, 62 complete omissions, 69 partial omissions) plus 47 clean twins, 198 notes per tier. The primary-family columns are recomputed from the original confirmation store by the same functions, and are replicate one throughout, so that both checkers are read at equal treatment; the 0.786 and 0.762 of Sections 7.2 and 7.5, and the paired commission figures Table 8 prints beside them, are means over three primary-family runs. There were zero parse failures on either tier.

{
{\small\begin{longtable}[]{@{}
  >{\raggedright\arraybackslash}p{(\linewidth - 8\tabcolsep) * \real{0.2000}}
  >{\raggedright\arraybackslash}p{(\linewidth - 8\tabcolsep) * \real{0.2000}}
  >{\raggedright\arraybackslash}p{(\linewidth - 8\tabcolsep) * \real{0.2000}}
  >{\raggedright\arraybackslash}p{(\linewidth - 8\tabcolsep) * \real{0.2000}}
  >{\raggedright\arraybackslash}p{(\linewidth - 8\tabcolsep) * \real{0.2000}}@{}}
\toprule\noalign{}
\begin{minipage}[b]{\linewidth}\raggedright
tier
\end{minipage} & \begin{minipage}[b]{\linewidth}\raggedright
checker
\end{minipage} & \begin{minipage}[b]{\linewidth}\raggedright
paired omissions
\end{minipage} & \begin{minipage}[b]{\linewidth}\raggedright
paired commissions
\end{minipage} & \begin{minipage}[b]{\linewidth}\raggedright
swept detection at false alarms \textless= 10\%
\end{minipage} \\
\midrule\noalign{}
\endhead
\bottomrule\noalign{}
\endlastfoot
two-stage pipeline (B2) & primary & 0.801 & 0.650 & 16.0\% (21/131) at 8.5\% (4/47) \\
two-stage pipeline (B2) & \textbf{second} & \textbf{0.798} & \textbf{0.575} & \textbf{19.1\% (25/131) at 8.5\% (4/47)} \\
three-stage pipeline with the severity audit (B3) & primary & 0.752 & 0.500 & 18.3\% (24/131) at 8.5\% (4/47) \\
three-stage pipeline with the severity audit (B3) & \textbf{second} & \textbf{0.824} & \textbf{0.700} & \textbf{14.5\% (19/131) at 4.3\% (2/47)} \\
\end{longtable}\addtocounter{table}{-1}}
}

Conditioning by how much of the fact survives, paired, omissions only:

{
{\small\begin{longtable}[]{@{}
  >{\raggedright\arraybackslash}p{(\linewidth - 8\tabcolsep) * \real{0.2000}}
  >{\raggedright\arraybackslash}p{(\linewidth - 8\tabcolsep) * \real{0.2000}}
  >{\raggedright\arraybackslash}p{(\linewidth - 8\tabcolsep) * \real{0.2000}}
  >{\raggedright\arraybackslash}p{(\linewidth - 8\tabcolsep) * \real{0.2000}}
  >{\raggedright\arraybackslash}p{(\linewidth - 8\tabcolsep) * \real{0.2000}}@{}}
\toprule\noalign{}
\begin{minipage}[b]{\linewidth}\raggedright
tier
\end{minipage} & \begin{minipage}[b]{\linewidth}\raggedright
checker
\end{minipage} & \begin{minipage}[b]{\linewidth}\raggedright
complete removal
\end{minipage} & \begin{minipage}[b]{\linewidth}\raggedright
fragment trace
\end{minipage} & \begin{minipage}[b]{\linewidth}\raggedright
restatement trace
\end{minipage} \\
\midrule\noalign{}
\endhead
\bottomrule\noalign{}
\endlastfoot
two-stage (B2) & primary & 0.935 & 0.792 & 0.561 \\
two-stage (B2) & \textbf{second} & 0.919 & 0.778 & \textbf{0.591} \\
three-stage (B3) & primary & 0.863 & 0.819 & 0.470 \\
three-stage (B3) & \textbf{second} & 0.911 & 0.847 & \textbf{0.636} \\
\end{longtable}\addtocounter{table}{-1}}
}

The per-fact rule of Section 7 flags a note if any fact the audit graded critical is judged absent. Because the severity grades are the cached ones, the rule's trigger class is identical for the two checkers by construction, and what changes is only which facts each checker judges absent.

{
{\small\begin{longtable}[]{@{}
  >{\raggedright\arraybackslash}p{(\linewidth - 4\tabcolsep) * \real{0.3333}}
  >{\raggedright\arraybackslash}p{(\linewidth - 4\tabcolsep) * \real{0.3333}}
  >{\raggedright\arraybackslash}p{(\linewidth - 4\tabcolsep) * \real{0.3333}}@{}}
\toprule\noalign{}
\begin{minipage}[b]{\linewidth}\raggedright
metric, on 151 held-out pairs and 47 clean twins
\end{minipage} & \begin{minipage}[b]{\linewidth}\raggedright
primary checker
\end{minipage} & \begin{minipage}[b]{\linewidth}\raggedright
second-family checker
\end{minipage} \\
\midrule\noalign{}
\endhead
\bottomrule\noalign{}
\endlastfoot
detection, all omissions & 20.6\% (27/131) & \textbf{22.1\% (29/131)} \\
false alarms, clean twins & 2.1\% (1/47) & \textbf{2.1\% (1/47)} \\
detection-vs-false-alarm z & 2.99 & \textbf{3.14} \\
detection on critical-severity pairs & 32.8\% (19/58) & \textbf{36.2\% (21/58)} \\
complete removals & 33.9\% (21/62) & 35.5\% (22/62) \\
fragment trace & 16.7\% (6/36) & 19.4\% (7/36) \\
restatement trace & 0.0\% (0/33) & \textbf{0.0\% (0/33)} \\
commissions flagged & 15.0\% (3/20) & 15.0\% (3/20) \\
\end{longtable}\addtocounter{table}{-1}}
}

The comparator that makes this a like-for-like test is the same family's own monolithic judge, restricted to exactly these 151 pairs and 47 clean twins from the records of Section G.2. Those records were already bought, so this row cost nothing and is measured on identical notes.

{
{\small\begin{longtable}[]{@{}
  >{\raggedright\arraybackslash}p{(\linewidth - 8\tabcolsep) * \real{0.2000}}
  >{\raggedright\arraybackslash}p{(\linewidth - 8\tabcolsep) * \real{0.2000}}
  >{\raggedright\arraybackslash}p{(\linewidth - 8\tabcolsep) * \real{0.2000}}
  >{\raggedright\arraybackslash}p{(\linewidth - 8\tabcolsep) * \real{0.2000}}
  >{\raggedright\arraybackslash}p{(\linewidth - 8\tabcolsep) * \real{0.2000}}@{}}
\toprule\noalign{}
\begin{minipage}[b]{\linewidth}\raggedright
configuration, all judged by the second family
\end{minipage} & \begin{minipage}[b]{\linewidth}\raggedright
paired omissions
\end{minipage} & \begin{minipage}[b]{\linewidth}\raggedright
paired commissions
\end{minipage} & \begin{minipage}[b]{\linewidth}\raggedright
detection
\end{minipage} & \begin{minipage}[b]{\linewidth}\raggedright
false alarms
\end{minipage} \\
\midrule\noalign{}
\endhead
\bottomrule\noalign{}
\endlastfoot
monolithic completeness-scored, asked once, 3 reps (score \textless= 7) & 0.668 & 1.000 & 16.5\% (65/393) & 2.1\% (3/141) \\
two-stage pipeline (B2), swept at false alarms \textless= 10\% & 0.798 & 0.575 & 19.1\% (25/131) & 8.5\% (4/47) \\
three-stage pipeline (B3), swept at false alarms \textless= 10\% & 0.824 & 0.700 & 14.5\% (19/131) & 4.3\% (2/47) \\
\textbf{three-stage pipeline under the per-fact rule} & (as B3) & & \textbf{22.1\% (29/131)} & \textbf{2.1\% (1/47)} \\
\end{longtable}\addtocounter{table}{-1}}
}

The monolithic comparator's three replicates on this subset read 0.688, 0.658 and 0.660 paired on omissions, 1.000 on commissions in each, and 16.0\% (21/131), 16.8\% (22/131) and 16.8\% (22/131) detection at 2.1\% (1/47), 0.0\% (0/47) and 4.3\% (2/47) false alarms. The pipeline rows are one replicate each, so the monolithic mean is the like-for-like figure.

Five readings follow, and the first is the one Section 7 rests on. \textbf{Restructuring the task keeps its lead on the stronger judge.} On identical notes, the pipeline reads 0.798 and 0.824 paired against 0.668 for that family's own monolithic judge, worth +0.129 and +0.156.

\textbf{The per-fact rule is the closest thing in this study to an operating point that is not judge-specific.} It is a rule over the individual fact verdicts and was tuned on nothing, and across the change of checker family it moves from 20.6\% (27/131) detection at one false alarm in 47 to 22.1\% (29/131) at the same one false alarm in 47, which is two further true detections in 131 notes. On the 58 pairs carrying a critical-severity omission it goes from 32.8\% (19/58) to 36.2\% (21/58). At that matched 2.1\% false-alarm rate this family's monolithic judge reads 16.5\% (65/393), so the rule's edge here is +5.6 points of detection, alongside the fact that its flag names the missing fact and its severity where a monolithic flag is a number.

\textbf{The ranking between the two tiers is judge-dependent even though the ranking of pipeline over monolithic is not.} The three-stage tier overtakes the two-stage one on this checker (0.824 against 0.798), reversing the primary family's ordering, and the reason is visible in the verdict mix: this checker uses the trichotomy's middle label, which the primary checker at no reasoning effort did not. The same behaviour shows up downstream, where the quote-verified second look ran over 319 flagged absences on this checker against 299 on the primary family. Tier comparisons therefore have to travel with their checker.

\textbf{The restatement-trace problem splits in two, and this part is new rather than a replication.} On the paired measure the second-family three-stage pipeline discriminates restatement traces at 0.636, well clear of the 0.47 to 0.56 band every method in Section 7 sits in; under the per-fact rule its detection there is 0 of 33, identical to the primary checker's 0 of 33. The information is present at the fact level and the decision layer cannot yet act on it, so the open problem of Section 7 is at least partly a decision-rule problem rather than purely a discrimination one.

\textbf{The cost on commissions persists, and it is wider on this family.} The monolithic judge scores 1.000 on the subset's 20 commission pairs while the pipeline tiers score 0.575 and 0.700 and the rule flags 3 of them: a judge that enumerates what should be present wins on absence and loses on invention here too. Twenty commission pairs make that size indicative rather than precise.

\subsection{The two confounds, and what these runs cannot settle}

\textbf{The reasoning setting moves with the family.} The second family's endpoint refuses to disable reasoning, so it ran at the minimal setting where the eight designs ran with none. Judge family and reasoning setting cannot be separated by these runs, and every second-family figure above carries that.

\textbf{Checker effort is not matched tier for tier.} In the primary family the two-stage pipeline's check ran at no reasoning effort and the three-stage pipeline's at medium, as Appendix C.3 records; the second-family checker ran at minimal on both tiers. Family is therefore confounded with deliberation in both directions across the tier rows, and the tier reversal above is one place that matters.

Three further limits belong with the numbers. The pipeline rows are one replicate against a three-replicate monolithic comparator, and the faithfulness-only monolithic row is a single draw whose deltas are one run against a three-run mean. With 151 pairs the residual subgroups are small (62 complete removals, 36 fragment traces, 33 restatement traces) and their intervals are correspondingly wide; the commission side holds 20 pairs. And no significance test was run between the families, so the honest summary of this appendix is that the asymmetry, its conditioning, the restatement-trace collapse and the pipeline's lead over monolithic judging all reproduce on a second family, that single-note detection levels move with the family, and that a full cross-family comparison of the eight designs remains future work.

\section{The deployed faithfulness judge, verbatim}

Sections 4 to 6 use a faithfulness judge deployed in our own production evaluation as a reference judge, in three wordings that differ only in how they treat omissions. This appendix prints the as-shipped prompt in full, so the wording experiment of Section 6 can be read against the exact text it edited, and records how the other two wordings differ from it. All three files ship in the release with SHA-256 hashes (\texttt{w2\_\hspace{0pt}prompts/\hspace{0pt}PROMPTS.sha256}), and the placeholders \texttt{\{transcript\}} and \texttt{\{summary\}} are filled per note at call time. The print below normalises typographic dashes and quotes to ASCII for the monospace font. The released files keep their original bytes, and the hashes refer to those.

\subsection{What the three wordings are}

The as-shipped prompt below carries an explicit exclusion: an ``Omissions'' bullet stating that information present in the transcript but absent from the note is not an error and must not be flagged, plus one mention of omission in a list of things not to flag. The \textbf{exclusion-deleted} wording (\texttt{v14\_\hspace{0pt}noexcl.txt}) removes exactly those two passages and changes nothing else. The \textbf{affirmative} wording (\texttt{v14\_\hspace{0pt}incl.txt}) starts from the exclusion-deleted text and adds one bullet to the list of flaggable error types: a clinically significant omission - a finding, diagnosis, medication, allergy, safety-netting instruction, or red flag that is present in the transcript and material to the patient's care, but absent from the note. Section 6 reports what each change does: deleting the exclusion does not help and if anything hurts, and the whole effect belongs to the affirmative instruction.

\subsection{The as-shipped prompt}

\begin{lstlisting}
### Task
You are a clinical safety reviewer. You are given a doctor-patient
conversation transcript and an AI-generated clinical note (a SOAP-style
summary) produced from that transcript. Decide whether the note
contains at least one factual error that a clinician reviewing this
note should be alerted to.

Output a single binary decision: YES (flag for review) or NO.

### The fundamental question

For each thing in the note that's worded differently from the
transcript, ask: "would a clinician form a meaningfully different
understanding of the patient from this note than the transcript
actually supports?"

If yes for any single item in the note → flag YES.
If no for every item → flag NO.

### What COUNTS as flag-worthy

Flag the note if it contains at least one of:

1. **Fabricated content** -- a fact, finding, value, name, dose, date,
   medication, or symptom in the note that is not in the transcript at
   all and cannot be reasonably inferred from context.

2. **Material misrepresentation** -- information from the transcript
   that the note got materially wrong: direction flipped (patient
   denied X, note asserts X), symptom substituted (patient said
   dizziness, note says nausea), attributed to the wrong subject
   (uncle's history attributed to father), or wrong value/quantity.

3. **Substantive over-reach** -- a clinical conclusion or specification
   in the note that goes beyond what the transcript supports in a way
   that changes clinical meaning. Examples: pre-supposing a diagnosis
   the doctor was actually evaluating; specifying a duration or
   quantity the patient did not give; adding a qualifier that narrows
   or shifts clinical interpretation.

### What does NOT count as flag-worthy

The following are NORMAL parts of clinical documentation. Do NOT flag
on these grounds, even though the wording differs from the transcript:

- **Paraphrasing into clinical language.** Patient says "my back hurts
  when I bend over" → note: "back pain worse with flexion". Normal.

- **Standard medical abbreviations.** BP, HR, ROM, PMH, SOB, NKDA,
  WNL, etc. -- these are not errors when applied to transcript content.

- **Standard normal-exam phrasing.** If the doctor said "heart sounds
  clear" or "lungs clear", the note may expand to "no murmurs, gallops,
  or rubs" / "no wheezes, rales, or rhonchi" / "clear to auscultation
  bilaterally". This is conventional clinical write-up, not over-reach.

- **Standard imaging/lab summary statements.** Doctor describes
  specific normal findings on a study; note writes "unremarkable" or
  "within normal limits" -- conventional summary, not error.

- **Section headings and boilerplate.** "Chief Complaint:",
  "Assessment:", "Plan:", "Review of Systems:". The standard ROS
  template phrasing ("denies fevers, chills, weight changes...") when
  the patient was asked about general systems is also boilerplate.

- **Omissions.** Information present in the transcript that the note
  didn't include. Omissions are NOT errors. Flagging is for things
  that are wrong, not things that are missing.

- **Reasonable contextual inference.** Writing "today's visit" when
  the visit context makes the date obvious; writing "COVID vaccine"
  when the patient mentions getting "the vaccine" in a context where
  that's the relevant one; using the standard clinical name for an
  exam the doctor described informally.

- **Light synthesis.** Drawing a conclusion that follows directly from
  facts already stated in the transcript, where a clinician reading the
  note would arrive at the same conclusion.

Important calibration: most clinical notes contain MANY paraphrases,
abbreviations, summary statements, and template phrases. These are
normal. A flag-worthy note has at least one item that goes beyond
normal documentation -- not just "the wording differs from the
transcript". If you can't point to a specific item where a clinician
would say "wait, that's not what the transcript supports", the answer
is NO.

### Examples (synthetic -- illustrative)

Each example shows a transcript snippet and a corresponding note.
The verdict is the doc-level YES/NO decision and the reasoning.

--- Example 1 (verdict: NO) ---

Transcript:
[doctor] How have you been since last year, Mr. Park?
[patient] Pretty good. The blood pressure medication is working -- at
home it's been around 130 over 80 most days.
[doctor] Any chest pain, shortness of breath, or palpitations?
[patient] No, none of those.
[doctor] How's your sleep?
[patient] Sleep is fine.
[doctor] Let me take a listen. ... Heart sounds clear, lungs are
clear. You're up about two pounds from last visit -- that's within
normal variation.
[doctor] Let's continue the same medication and see you in a year.

Note:
Mr. Park presents for annual follow-up of hypertension. He reports
good control on his current regimen with home BP averaging 130/80.
He denies chest pain, dyspnea, or palpitations. Sleep is adequate.

Cardiovascular: No murmurs, gallops, or rubs.
Pulmonary: Lungs clear to auscultation bilaterally.
Weight: Up 2 lbs from prior visit, within normal variation.

Plan: Continue current antihypertensive regimen. Annual follow-up.

Verdict: NO
Reasoning: Every claim has a transcript anchor. "Dyspnea" is the
clinical term for "shortness of breath" -- paraphrase, not error. "No
murmurs, gallops, or rubs" is the standard write-up of "heart sounds
clear" -- light synthesis, not over-reach. BP and lbs are standard
abbreviations. No item where a clinician would form a different
understanding from the note than the transcript supports.

--- Example 2 (verdict: NO) ---

Transcript:
[doctor] What brings you in today?
[patient] My right knee has been bothering me for a couple of weeks,
mostly when I go up stairs.
[doctor] Any injury you remember?
[patient] No, it just started.
[doctor] Any swelling or redness?
[patient] No.
[doctor] Let me take a look. ... Tender on the inside of the knee, no
swelling, range of motion is full but with some discomfort.
[doctor] Could be a strain or early arthritis. Let's try ibuprofen
and physical therapy. If it's not better in a month, we'll image it.

Note:
HPI: Patient presents with two weeks of right knee pain, worse with
stair climbing. No history of trauma. Denies swelling or erythema.

Exam: Right knee tender to palpation medially. No effusion. Full ROM
with mild discomfort on flexion.

Assessment: Right knee pain, likely musculoskeletal strain vs early
osteoarthritis.

Plan: Ibuprofen for pain. Physical therapy referral. Imaging if no
improvement in 4 weeks.

Verdict: NO
Reasoning: Paraphrasing into clinical language ("erythema" for
"redness", "ROM" for "range of motion", "medially" for "the inside of
the knee", "no effusion" for "no swelling"). Assessment matches the
doctor's verbal differential. No substantive issue.

--- Example 3 (verdict: YES) ---

Transcript:
[doctor] Mrs. Allen, how have you been since starting the new
medication?
[patient] Pretty good, the headaches are mostly gone.
[doctor] Any side effects?
[patient] None that I've noticed.
[doctor] Let's continue the same plan. See you in three months.

Note:
Mrs. Allen reports significant improvement in headaches since starting
amitriptyline 50mg nightly. She denies side effects.

Plan: Continue amitriptyline 50mg HS. Return in three months.

Verdict: YES
Reasoning: Fabricated content. The transcript refers only to "the new
medication" -- no medication name, no dose, no schedule is mentioned.
"Amitriptyline 50mg nightly" has no transcript anchor. A clinician
reading this note would believe the patient is on a specific drug
that is in fact unspecified by the transcript.

--- Example 4 (verdict: YES) ---

Transcript:
[doctor] Any nausea, vomiting, or weight changes?
[patient] No nausea or vomiting. My weight has actually been stable.
[doctor] Let me check your abdomen. ... Soft, non-tender, no masses.

Note:
ROS: Patient reports recent 5-pound weight loss. Denies nausea or
vomiting.

Abdomen: Soft, non-tender, no masses appreciated.

Verdict: YES
Reasoning: Material misrepresentation. The patient said weight is
stable; the note asserts a 5-pound weight loss with a specific
number. Direction flipped and quantity fabricated. A clinician
reading this note would believe the patient is losing weight.

--- Example 5 (verdict: YES) ---

Transcript:
[doctor] How long have you had the chest pain?
[patient] A while now -- hard to say. Maybe a year?
[doctor] Any cardiac workup before?
[patient] No, never.
[doctor] Given the history, I want an EKG today and probably a stress
test.

Note:
Patient reports a 12-month history of stable angina. Cardiac workup
is being initiated; EKG and stress test ordered.

Assessment: Stable angina, awaiting workup.

Verdict: YES
Reasoning: Substantive over-reach. The doctor is initiating workup to
EVALUATE the chest pain; the diagnosis "stable angina" was not given
and is exactly what the workup is meant to establish. A clinician
reading this note would believe the patient has a confirmed diagnosis
of angina -- fundamentally different clinical understanding.

### Transcript
{transcript}

### Clinical note
{summary}

### Instructions
1. Read the note. For each item that's worded differently from the
   transcript, ask the fundamental question: would a clinician form a
   meaningfully different understanding from this note than the
   transcript supports?
2. If you find at least one item where the answer is YES, flag the
   doc YES.
3. If every item is paraphrase / abbreviation / standard write-up /
   reasonable contextual inference / light synthesis / omission, flag
   the doc NO.
4. End your response with exactly one line:
   `Verdict: YES` or `Verdict: NO`
\end{lstlisting}

\section{The comparator literature in full}

Section 2 names four literatures and states what each contributes; this appendix carries the passages that section no longer prints at full resolution, and collects the published human-detection figures that Section 8's human-baseline limitation summarises in a sentence. Nothing here is a new claim: every figure was verified against its primary source between 17 and 23 August 2026 and is quoted at the resolution the primary supports.

\subsection{The concurrent and adjacent studies, in full}

\textbf{The reviewing role, on non-clinical answer keys.} Concurrent work locates the same asymmetry in the reviewing role itself. Models asked to critique a key with planted defects catch inserted items six to seven times more often than omitted ones in that study's single-reviewer condition, and the authors argue the reason is structural - a reviewer can interrogate only candidates it can name, so certifying that nothing is missing is itself an enumeration task \citep{chen2026judging}. Their acceptable sets are finite and constructed and their omissions binary. Ours is the setting they leave untested: free clinical text with no enumerable answer set, omissions graded by severity and by surviving trace, and a source document on hand to supply the candidates their reviewer lacks.

\textbf{Three further angles on judge reliability, including the one that points the other way.} A perturbation study \citep{dahlberg2026measuring} shows a clinical judge responding to deletions at roughly half its response to modifications, across three languages, without remarking on it. One mental-health study \citep{hussain2026blending} reports the opposite direction on prevalence-sensitive F1, a task and metric difference rather than a contradiction. Two concurrent vendor-authored studies pose the reliability question from adjacent angles. In one, ten external clinicians agreed with each other only fairly on whether an automated judge's flag was genuine (Gwet's AC1 0.24, an agreement coefficient), with the judge inside that envelope \citep{bergman2026judges}. In the other, re-running one hallucination judge under two criteria on the same 100 SOAP notes moved the measured rate from 35.2\% to 9.1\% \citep{vachhani2026beyond}, a criterion effect on the commission side of the kind Section 6 measures on the omission side.

\textbf{The instrument's own share of a published rate.} A concurrent five-country paired study makes the instrument's share measurable from inside one vendor: on the same AI-generated notes, clinician adjudication found errors in 6.2\% where a calibrated automated reviewer found 24.4\%, and the authors publish both figures rather than choosing one \citep{bergman2026quality}.

\subsection{Published human detection of omissions}

We did not run a human baseline, and the published record supplies the comparators. In a planted-error study, reviewing physicians caught omissions at roughly the same rate as objective errors, about a quarter to a third of each \citep{biro2025opportunities} - no human asymmetry, at levels in the same range as our best configurations' 24.6 to 36.9\%, though on a different task and corpus. MEDEC's two physician baselines reach 0.81 and 0.69 error-flag accuracy on commissions on its MS-test subset (document-level flags, not a per-error catch rate). Clinicians reading with the source document alongside recall 50.4\% of omissions in one review study \citep{bedi2026care}, an optimistic upper bound. And radiologists catch word-level, internally detectable omissions at rates from 42\% to 92\% (mean 61.8\% across their twelve readers) \citep{shen2026error}, an easier task than semantic absence. No published figure covers human detection of omissions whose restatement survives elsewhere. If humans also miss those, that case is a document-design problem as much as a judge problem.

\section{What to ask of any scribe evaluation, including this one}

\subsection{Four questions, and our own answers}

Four questions separate an evaluation that means something from one that does not, and we answer them against our own record.

\textbf{Which error class is being measured?} Existing clinical judge benchmarks centre on commissions, where judges are strong - MEDEC's error types are all substitutions of incorrect content, with no omission class among them, so a judge validated there tells you nothing about the class that dominates production. \emph{Ours}: both classes, separately, never pooled.

\textbf{Against what ground truth, and how was absence established?} Every clinician reference note we audited was materially discrepant with its own transcript by our audit instrument's standard (100\% of 98 notes, 7 to 11 discrepancies each), so a scribe evaluated against reference notes is scored against a lossy answer key. And certain absence is harder to build than it sounds: 44\% of our complete-removal attempts failed verification. \emph{Ours}: constructed absence, verified by a cross-family panel, graded by surviving trace and severity, failures published rather than patched.

\textbf{At what false-alarm rate?} A detection number with no false-alarm rate beside it is uninterpretable, and this is a common defect in coverage-style evaluation: the natural flag-everything rule of the RAGAS-style recipe of Section 4 fires on 98.7\% of clean notes while its score carries the best omission signal we measured. \emph{Ours}: every operating point is quoted with the false-alarm rate at that same threshold.

\textbf{With which instrument, and is it published?} A failure count is a joint property of the scribe and the instrument that counted it. The companion census shows how large the instrument's share can be. An early version of its verification panel verified 37.3\% of the candidates put to it and the final cross-family panel 10.48\%, a factor of three and a half from the same three products. That is an uncontrolled contrast from the census's own records, since more than the panel changed between the runs. Any vendor rate you are shown is an instrument reading. \emph{Ours}: the instrument, its prompts, its model pins and its corrections ship with the rate, and its own weaknesses are Section 8's subject.

\section{The three judges on real vendor notes}

Everything in Sections 5 to 7 is measured on constructed absence: pairs in which exactly one fact was removed by design. The companion census supplies the other kind of evidence - real notes written by three deployed scribes, with omissions found and verified by an independent panel - and this appendix reports the one run in the study that points the paper's judges at them. Nothing was tuned on this corpus: each judge runs once, at the rule the benchmark sections publish for it. Section 8's injected-errors limitation summarises the result; this is the full record.

\subsection{Design}

The set is 261 of the census's 565 notes: all 87 that carry a panel-verified omission, and 174 of the 388 on which twelve discovery passes and a two-family verification panel found nothing at all, drawn stratified by product and corpus stratum (proportional, largest remainder, seed 20260825, worst marginal gap against the 388-note pool 0.32 percentage points). The 90 notes whose only verified findings are non-omission classes are excluded - they are neither clean nor omission-positive - and three of the 90, whose findings match no published class at all, fall under the same rule rather than being moved quietly into the clean class. A note's class is recomputed from its findings through the study's taxonomy code rather than read off a stored field: 155 of the census's 618 verified findings carry no stored top-tier class, and the stored field alone would miss 14 of the 87 omission notes.

Three judges run at their own published rules: the best monolithic judge (the winsorised eight-sample mean, flag below 8.0), the evolved prompt (route two in the released data) at its deployed flag-below-10 rule, and the enumerate-then-check pipeline's three-stage tier (route one in the released data) under the per-fact rule. The pipeline must be the three-stage tier here. The two-stage tier judges the unaudited fact list, so its verdicts carry no severity grades and the per-fact rule cannot be stated on its output at all (the only rule it can express, any fact absent, fires on 72.3\% of the benchmark's own clean twins). The census's 142 consultations are a strict superset of the evaluation set's 112, so extraction and audit artefacts are reused wherever they exist: of the 131 consultations these notes span, 103 had both cached and 28 were extracted and audited fresh. The run produced 783 judgements with zero parse failures. Its records judge the withheld vendor notes and are handled under the same terms as the notes themselves (Section 9).

\subsection{Every operating point moves, and none dominates}

{
{\small\begin{longtable}[]{@{}
  >{\raggedright\arraybackslash}p{(\linewidth - 8\tabcolsep) * \real{0.2000}}
  >{\raggedright\arraybackslash}p{(\linewidth - 8\tabcolsep) * \real{0.2000}}
  >{\raggedright\arraybackslash}p{(\linewidth - 8\tabcolsep) * \real{0.2000}}
  >{\raggedright\arraybackslash}p{(\linewidth - 8\tabcolsep) * \real{0.2000}}
  >{\raggedright\arraybackslash}p{(\linewidth - 8\tabcolsep) * \real{0.2000}}@{}}
\toprule\noalign{}
\begin{minipage}[b]{\linewidth}\raggedright
judge
\end{minipage} & \begin{minipage}[b]{\linewidth}\raggedright
rule as published
\end{minipage} & \begin{minipage}[b]{\linewidth}\raggedright
detection, 87 verified-omission notes
\end{minipage} & \begin{minipage}[b]{\linewidth}\raggedright
false alarms, 174 no-finding notes
\end{minipage} & \begin{minipage}[b]{\linewidth}\raggedright
the same rule on the benchmark
\end{minipage} \\
\midrule\noalign{}
\endhead
\bottomrule\noalign{}
\endlastfoot
the best monolithic judge (FC / score / k=8) & winsorised mean below 8.0 & 29.9\% (26/87) {[}21.3, 40.2{]} & 9.8\% (17/174) {[}6.2, 15.1{]} & 8.3\% at 6.5\%, swept (Section 5) \\
the pipeline under the per-fact rule (route one) & any critical fact absent & 51.7\% (45/87) {[}41.4, 61.9{]} & 22.4\% (39/174) {[}16.9, 29.2{]} & 24.6\% at 2.7\% (Section 7) \\
the evolved prompt (route two) & score below 10 & 85.1\% (74/87) {[}76.1, 91.1{]} & 44.2\% (77/174) {[}37.1, 51.7{]} & 36.9\% at 6.2\% (Section 7) \\
\end{longtable}\addtocounter{table}{-1}}
}

Detection roughly doubles for both methods and more than triples for the monolithic judge, and false-alarm rates rise 8.3-fold on the pipeline, 7.1-fold on the evolved prompt and by half on the monolithic judge, so the three sit in a strict ordering in which whichever detects more also flags more clean notes. A threshold established on constructed clean twins does not carry to a production corpus - a caution about deployment rather than about the judges, and the reason the benchmark's operating points are quoted nowhere in this appendix as expectations for real traffic.

The mechanism is measurable from the stored per-fact verdicts. The pipeline checks the same number of audited facts per note on both corpora - 37.6 on the benchmark's held-out notes against 37.3 here - but it judges 2.0 facts absent per real note against 1.1 on the benchmark, and its second look is called over 795 flagged facts against 299. A constructed pair has exactly one fact removed, and a real note is simply less complete. That is also why the pipeline's check-and-second-look stage costs \$0.0667 a real note against the benchmark's \$0.0440. The per-consultation extraction and audit stages are unchanged, and the benchmark's full-pipeline price is \$0.45. The ``false alarm'' column above should be read with that in mind: the 174 notes are clean only as far as the census's discovery reached, and the same verdicts say they are not fact-complete.

\subsection{The evolved prompt's calibration, measured failing and re-established}

Section 7 states that the evolved prompt's threshold is a calibration, not a design guarantee: it works because clean notes score exactly 10, which 96 to 98\% of held-out twins and 90 to 94\% of evaluation-set twins do. On real no-finding notes 97 of 174 score exactly 10 - 55.8\% - with 54 more at 9, so flag-below-10 fires on 44.2\% of them. That is the predicted failure, measured. The signal survives re-calibration. Swept on the same records, the evolved prompt reads 25.3\% detection at 5.2\% false alarms at flag-below-7 (32.2\% at 5.2\% at flag-below-8), while the monolithic judge has no comparable threshold - 9.2\% at 1.1\% at its own below-7, and 49.4\% false alarms by below-9. So the ordering between those two survives the move to real errors even though the thresholds do not. The two are not read at the same false-alarm rate here, since the monolithic judge's threshold is the quieter of the pair, and no monolithic threshold reaches the evolved prompt's detection inside the 5\% false-alarm band. (The per-fact rule carries no swept threshold and was not re-established.) Calibration also varies by product more than anything else measured here - the evolved prompt's false-alarm rate spans 30 to 71\% across the three scribes - so a per-deployment clean set is not optional for this method, and per-product rates measured this way are properties of the instrument as much as of any product.

\subsection{The pipeline's flags name the right fact}

A per-fact flag can be checked against the census's independently verified finding: two verbalisations of the same absence, one by the judge from the transcript, one by the panel. On the 87 verified-omission notes the rule fired on 45, naming 60 facts, and 34 of the 45 flagged notes (75.6\% {[}61.3, 85.8{]}) name a fact that matches the panel's verified omission under a lexical-containment screen at a 0.3 threshold (68.9\% at 0.4, 57.8\% at 0.5). The pairs are legible rather than marginal - ``ulnar styloid fracture was stated to be present'' against ``documented ulnar styloid fracture finding omitted (and negated) in the note''. This is an indicative screen, not an adjudication: read it as about three flags in four naming the fact the panel independently verified, not as a precision estimate with two significant figures. No injected benchmark can produce this number, because agreement there is true by construction.

\subsection{What these rates are, and are not}

\begin{itemize}
\tightlist
\item
  \textbf{The false-alarm rates are upper bounds.} Discovery recall on the census is unmeasured, panel-cleared notes are not fact-complete (K.2), and the census's own standards experiment moves the share of notes carrying something reportable from 27.8\% to 96.5\% depending on the review instruction. A flag on a panel-cleared note is therefore not automatically wrong, and the three rates above are ceilings on the three judges' error, not measurements of it.
\item
  \textbf{Detection is conservative.} Ground truth is the panel's strict standard, so the 87-note positive class is a floor on the notes that carry an omission.
\item
  \textbf{Two of the three flags are note-level.} The 87 notes carry 284 verified findings across all classes, so a flag from the monolithic judge or from the evolved prompt cannot be attributed to the omission rather than to a co-occurring failure. Only the pipeline's flag names a fact, which is why K.4 exists for it alone.
\item
  \textbf{The evolved prompt runs once here, against a deployed rule that is a majority of three.} A single draw detects and false-alarms more than the majority does (Section 7.7's estimator-matched readings), so its figures here sit at a noisier point than its deployed rule would.
\item
  \textbf{The two ``critical'' gradings are different instruments} - the pipeline's audit grades on the benchmark's three-value rubric, the census on a two-value scale - so no severity-conditioned rate here is comparable with the benchmark's without naming the grader.
\item
  \textbf{One run, one judge family, one corpus}, with no replicates on any arm. These rates share no denominator with the benchmark's, so they are reported beside its tables and never inside them.
\end{itemize}

\end{document}